\documentclass[journal]{IEEEtran}
\usepackage[numbers,sort&compress]{natbib}
 \usepackage{graphicx} 
 \usepackage{float}
 \usepackage{multirow}
\usepackage{hyperref}
\usepackage{graphicx}
\ifCLASSINFOpdf
\else
\fi
\begin{document}
\title{The Role of Transformer Models in Advancing Blockchain Technology: A Systematic Survey }

\author{Tianxu Liu,~\IEEEmembership{}
        Yanbin Wang,~\IEEEmembership{}
        Jianguo Sun,~\IEEEmembership{}
        Ye Tian,~\IEEEmembership{}
        Yanyu Huang,~\IEEEmembership{}
        Tao Xue,~\IEEEmembership{}
        Peiyue Li,~\IEEEmembership{}
        Yiwei Liu~\IEEEmembership{}% <-this % stops a space
\thanks{Tianxu Liu, Yanbin Wang, Jianguo Sun, Ye Tian, Yanyu Huang, Tao Xue and Yiwei Liu are with Hangzhou Research Institute, Xidian University, Zhejiang Hangzhou, China, 311231 email:(Tianxu Liu: 1379657952@qq.com; Yanbin Wang: wangyanbin15@mails.ucas.ac.cn; Jianguo Sun: jgsun@xidian.edu.cn; Ye Tian: tianye@xidian.edu.cn; Yanyu Huang: onlyerir@163.com; Tao Xue: xuetao@xidian.edu.cn; Peiyue Li: lipeiyue@ppsuc.edu.cn; Yiwei Liu: yiweiliu@bit.edu.cn)}% <-this % stops a space
\thanks{Corresponding author: Yanbin Wang,Jianguo Sun and Ye Tian}% <-this % stops a space
\thanks{}}


\markboth{}%
{Shell \MakeLowercase{\textit{et al.}}: Bare Demo of IEEEtran.cls for IEEE Journals}



\maketitle

\begin{abstract}
As blockchain technology rapidly evolves, the demand for enhanced efficiency, security, and scalability grows. Transformer models, as powerful deep learning architectures, have shown unprecedented potential in addressing various blockchain challenges. However, a systematic review of Transformer applications in blockchain is lacking. This paper aims to fill this research gap by surveying over 200 relevant papers, comprehensively reviewing practical cases and research progress of Transformers in blockchain applications. Our survey covers key areas including anomaly detection, smart contract security analysis, cryptocurrency prediction and trend analysis, and code summary generation. To clearly articulate the advancements of Transformers across various blockchain domains, we adopt a domain-oriented classification system, organizing and introducing representative methods based on major challenges in current blockchain research. For each research domain, we first introduce its background and objectives, then review previous representative methods and analyze their limitations, and finally introduce the advancements brought by Transformer models. Furthermore, we explore the challenges of utilizing Transformer, such as data privacy, model complexity, and real-time processing requirements. Finally, this article proposes future research directions, emphasizing the importance of exploring the Transformer architecture in depth to adapt it to specific blockchain applications, and discusses its potential role in promoting the development of blockchain technology. This review aims to provide new perspectives and a research foundation for the integrated development of blockchain technology and machine learning, supporting further innovation and application expansion of blockchain technology. We will continue to update the latest articles and their released source codes at https://github.com/LTX001122/Transformers-Blockchain .
\end{abstract}


% \fi
\IEEEpeerreviewmaketitle

\section{Introduction}
\IEEEPARstart{I}{n} the digital age, blockchain technology \cite{nakamoto2008bitcoin} has emerged as one of the most revolutionary technologies \cite{tapscott2016blockchain}, empowering various industries such as finance \cite{swan2015blockchain,mcwaters2016future}, supply chain \cite{kshetri2017can,zheng2017overview,xu2021blockchain}, and healthcare \cite{mettler2016blockchain,ekblaw2016case,khezr2019blockchain,fiore2023blockchain} with its unique decentralized nature, immutability, and transparency.Since the advent of Bitcoin, various types of cryptocurrencies and smart contract platforms like Ethereum have demonstrated the extensive value of blockchain technology \cite{conoscenti2016blockchain,guo2016blockchain,wood2014ethereum}. However, with increasingly complex application scenarios and growing demands, blockchain technology faces numerous challenges, including transaction efficiency, data processing capabilities, and security issues \cite{conti2018survey,zheng2018blockchain,zhang2019security,li2020survey}.

Meanwhile, since its introduction \cite{vaswani2017attention} in 2017, the Transformer model has achieved groundbreaking progress in the field of natural language processing (NLP) \cite{xu2023multimodal,chernyavskiy2021transformers,zhang2024survey}.As shown in Figure 1, research literature on transformers has significantly increased in recent years. This model, based on the self-attention mechanism, effectively handles long-range dependencies and, due to its efficient parallel computing capabilities, has been widely applied to complex sequential data tasks. Its unique structure has shown remarkable abilities in image recognition \cite{dosovitskiy2020image}, speech processing \cite{gulati2020conformer}, and multimodal learning \cite{li2021align}. Given the sequential nature of blockchain data and the complex interrelations between transaction data \cite{Ju2020The}, applying Transformers to the blockchain domain, particularly in processing on-chain data, anomaly detection, and optimizing smart contracts, holds significant potential \cite{Dolgui2020Blockchain-oriented,Li2022SmartVM:,weng2019deepchain,shafay2023blockchain}.

\begin{figure}[h]
    \centering
    \includegraphics[width=0.5\linewidth, trim=5cm 9.5cm 5cm 9.5cm]{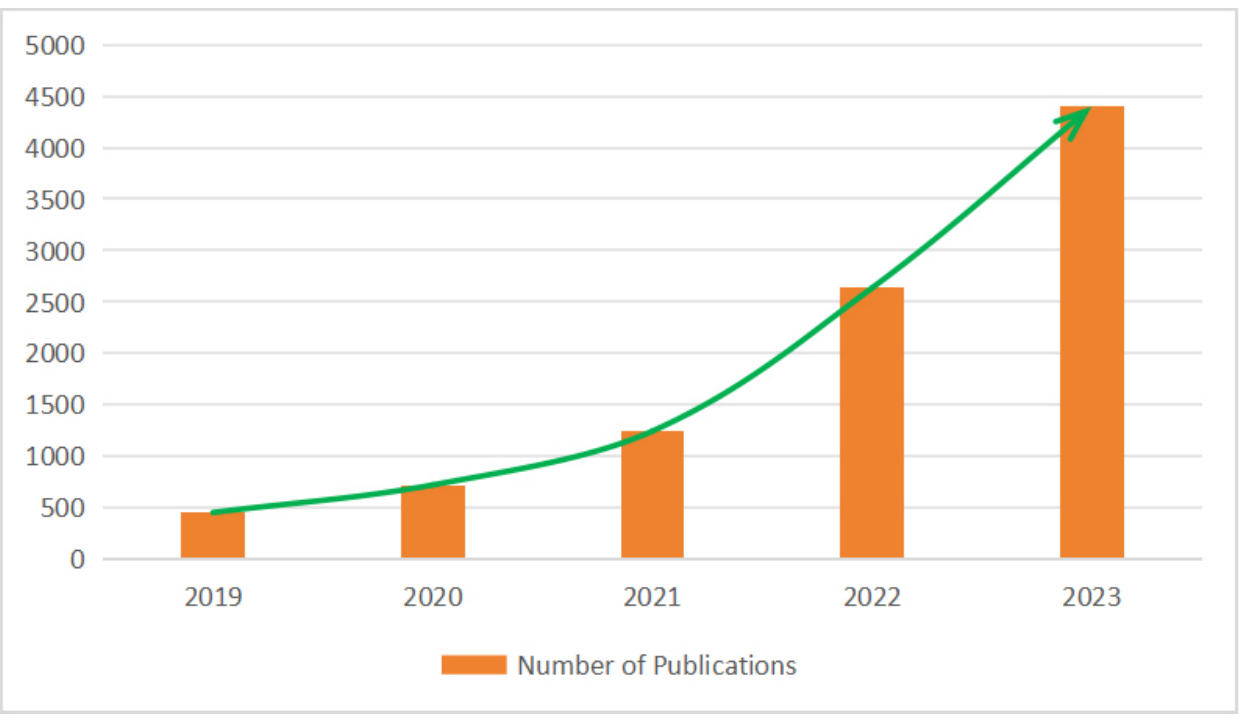}
    \caption{Statistics on the number of research papers on transformer models.}
    \label{fig:1}
\end{figure}

For example, in transaction monitoring and anomaly detection, traditional methods rely on simple rules or shallow machine learning models, which often fall short when dealing with complex, high-dimensional transaction data \cite{naseer2018enhanced,sanjay2023anomaly}. The introduction of the Transformer model, with its excellent data correlation analysis capabilities, provides new solutions for identifying complex fraud patterns. As shown in Figure 2, the research literature on the application of Transformer models to blockchain has steadily increased in recent years.
\begin{figure}[h]
    \centering
    \includegraphics[width=0.5\linewidth, trim=5cm 9.5cm 5cm 9.5cm]{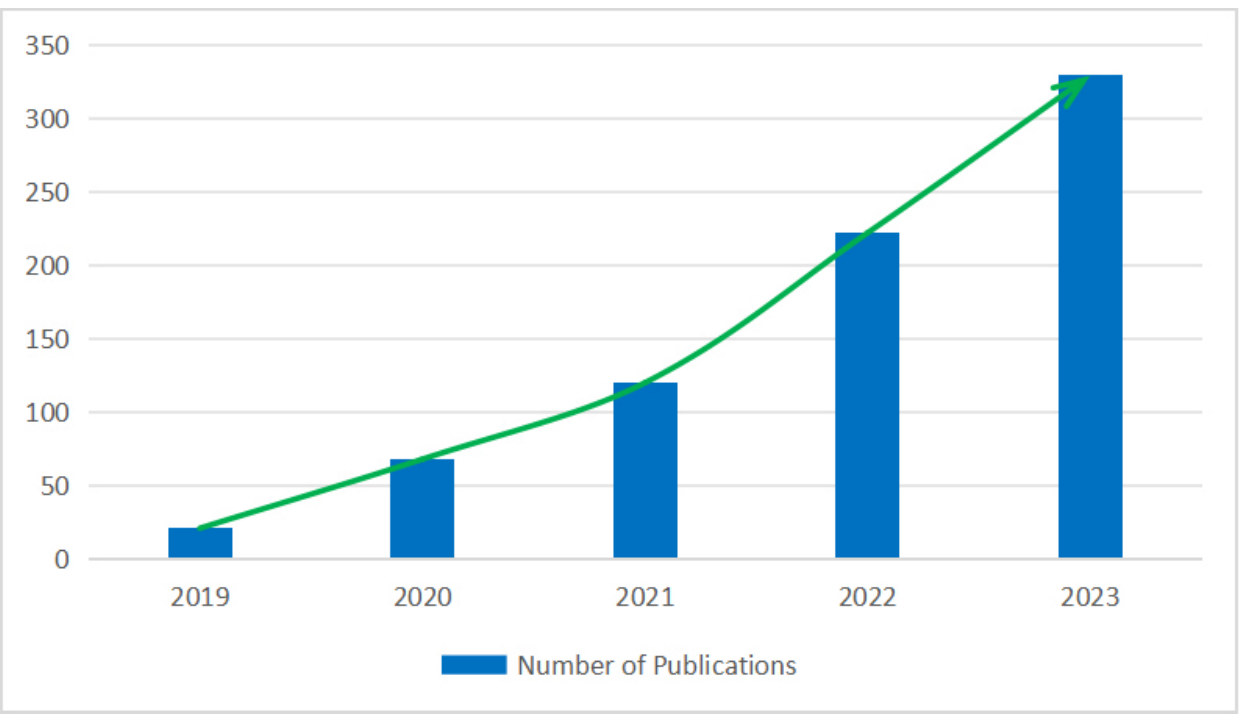}
    \caption{Statistics on the number of research papers on the application of transformer models to blockchain.}
    \label{fig:2}
\end{figure}
Moreover, with the proliferation of smart contracts, the issue of contract security has become increasingly prominent \cite{luu2016making,atzei2017survey}. Existing smart contract security analysis tools mostly rely on expert experience or traditional program analysis techniques, lacking sufficient automation and intelligence \cite{tikhomirov2018smartcheck}. 
The potential of the Transformer model in code semantic analysis and pattern recognition suggests promising applications in automated smart contract auditing and security analysis \cite{alon2019code2vec,ahmad2021unified}.

Based on the aforementioned background and research motivations, the primary objective of this paper is to comprehensively review the practices, challenges, and future directions of applying the Transformer model in blockchain technology. Specifically, this paper aims to:

1. Systematically introduce the Transformer model: Explain its fundamental principles, architectural features, and why it is effective in processing blockchain data.

2. Review the model's specific applications in blockchain: As shown in Figure 3,conduct an in-depth analysis of the applications and research achievements of Transformers in blockchain transaction anomaly detection, smart contract vulnerability detection, cryptocurrency prediction and trend analysis, and generating code summaries.
\begin{figure}[h]
    \centering
    \includegraphics[width=0.5\linewidth, trim=2.5cm 0.5cm 2.5cm 0.5cm]{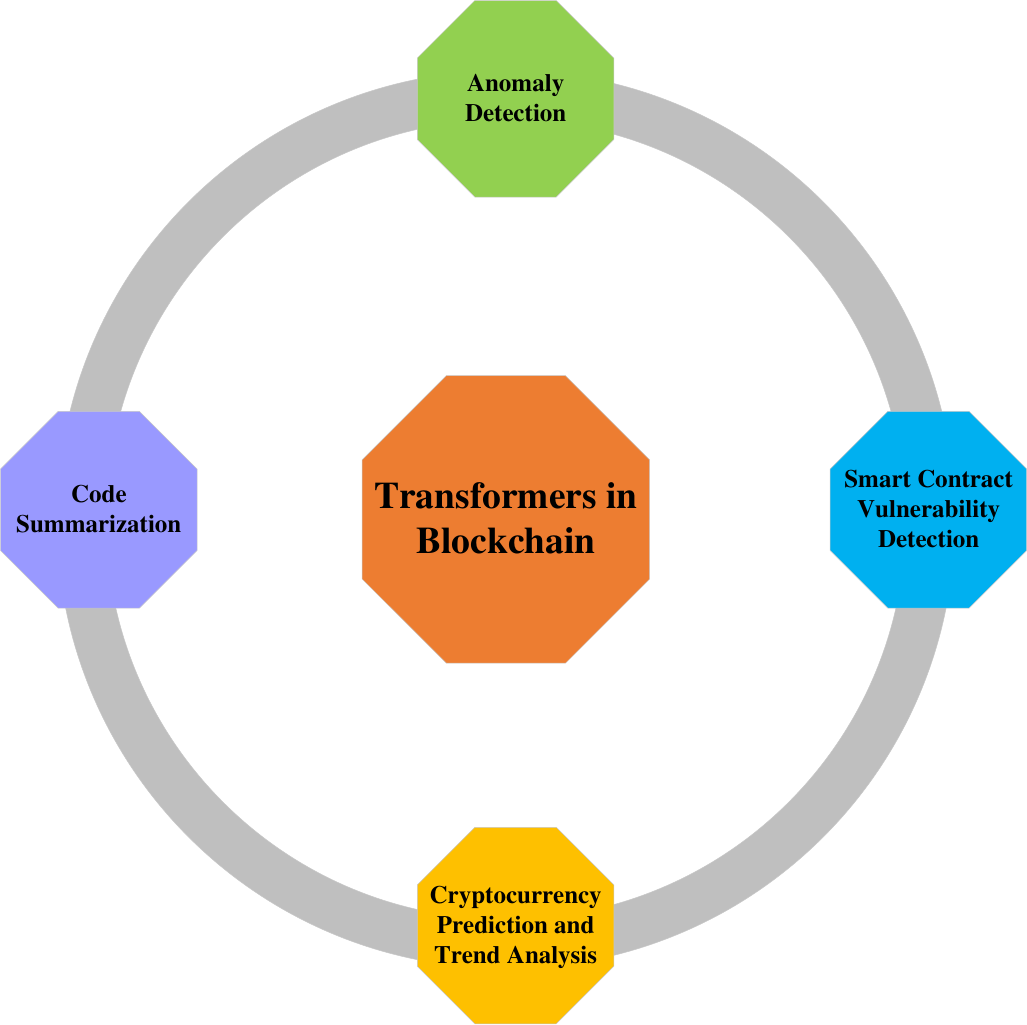}
    \caption{A diverse set of application areas of Transformers in blockchain covered
in this survey.}
    \label{fig:enter-label}
\end{figure}

3. Explore challenges and future directions: Identify the current challenges in these applications, discuss how to overcome them, and explore potential future research directions.

Through these efforts, the contributions of this paper are:

- Providing a comprehensive guide on the application of Transformers in blockchain, helping researchers and developers better understand the potential and limitations of this technology.

- Promoting the integration of blockchain and artificial intelligence technologies, opening up new research and application fields.

- Offering directions for future research, particularly in enhancing blockchain data processing and security.

Chapter 2 provides an overview of the Transformer model, introducing its architecture and its various derivatives, and analyzing their advantages and challenges. Chapter 3 introduces the fundamentals of blockchain technology. Chapter 4, through a summary and analysis of existing literature, details the current research progress and applications of Transformers in the blockchain domain. Chapter 5 discusses the challenges and future prospects of applying Transformers to blockchain. Finally, the paper concludes with a summary.

\section{Overview of the Transformer Model and Its Derivatives }
The Transformer model, first introduced by Vaswani et al. in the 2017 paper "Attention is All You Need," \cite{vaswani2017attention} represents a significant innovation in the field of deep learning. This model abandons the traditional Recurrent Neural Network (RNN) structure \cite{elman1990finding} and employs an all-attention mechanism \cite{bahdanau2016neural} to process sequential data. It is particularly effective in the parallel processing of long sequences and capturing long-range dependencies. Its unique self-attention mechanism and multi-head attention technique have become the cornerstone for addressing complex sequence transformation tasks, significantly advancing the fields of natural language processing (NLP) and beyond.

%\IEEEpubidadjcol

\subsection{Core Components of the Transformer Model}\

\subsubsection{Self-Attention Mechanism}

The self-attention mechanism is the core of the Transformer model, enabling the evaluation of direct influences and relationships between elements within a sequence without relying on sequential time steps. This mechanism primarily operates through three vectors: Query, Key, and Value. For each element in the sequence, the model generates these three types of vectors. By calculating the compatibility between the query vector and all key vectors, the model assigns a weight to each value vector, and then performs a weighted sum of all the value vectors to obtain the output representation of that element.\

\subsubsection{Multi-Head Attention Mechanism}

The working principle of the multi-head attention mechanism is shown in Figure 4.
\begin{figure}[h]
    \centering
    \includegraphics[width=0.8\linewidth, trim=2cm 1cm 2cm 1cm]{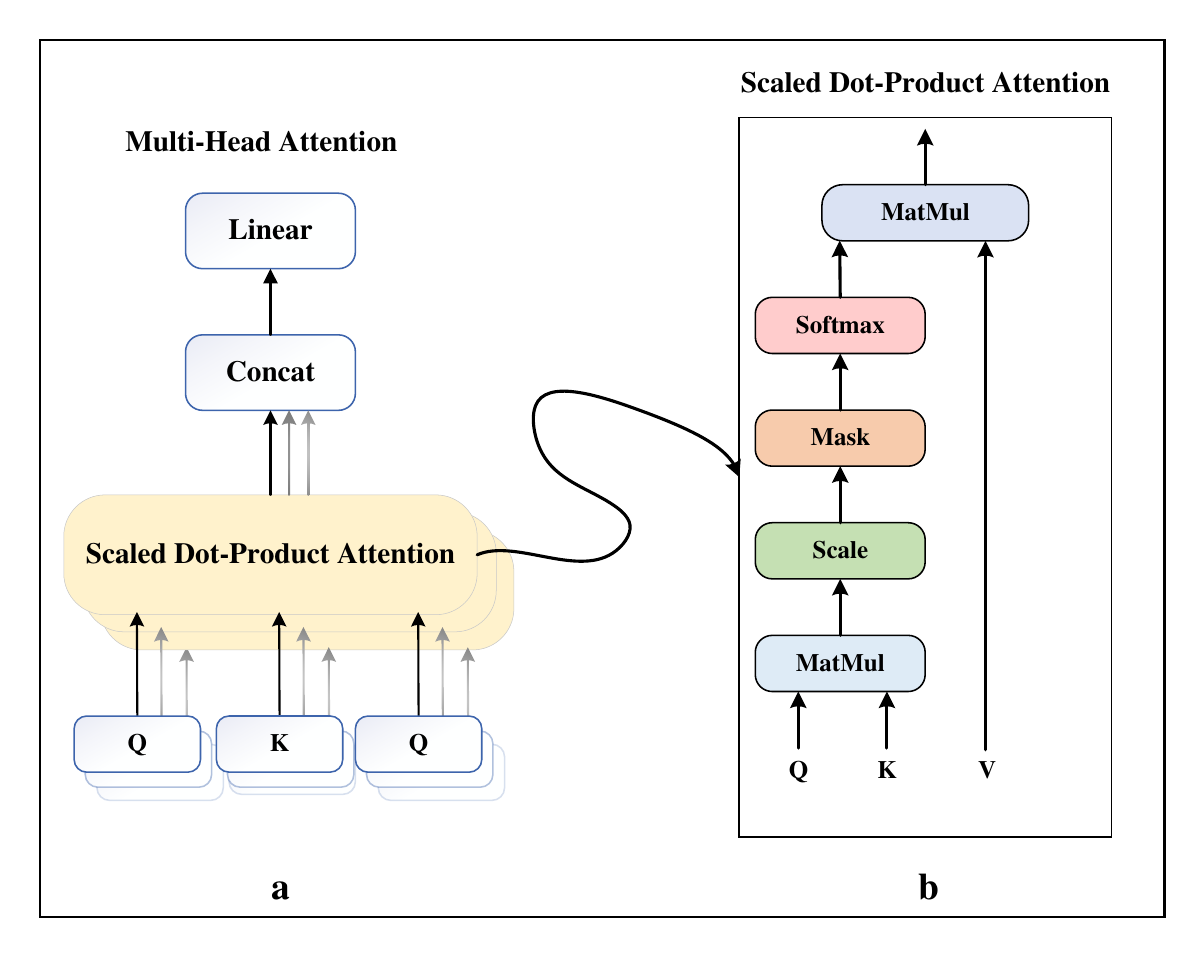}
    \caption{Multi head attention mechanism. In the encoder and decoder, multiple attention heads are stacked 
together and their outputs are concatenated}
    \label{fig:4}
\end{figure}Multi-head attention is an extension of self-attention that allows the model to process information in parallel across multiple representation subspaces. By splitting the attention layer into multiple "heads," with each head handling different parts of the input data, the model can learn intrinsic features from multiple dimensions. This not only enhances the expressive power of the attention layer but also increases the model's ability to capture complex patterns.\

Due to the lack of a recurrent structure, the Transformer cannot naturally handle positional information in input sequences like RNNs do. To address this issue, positional encoding is introduced. Positional encoding is added to the input embedding vectors to provide the model with information about the positions of words in the sequence. Typically, sine and cosine functions of different frequencies are used to generate unique encodings for each position, enabling the model to leverage the absolute or relative positions of words.\

\subsection{Detailed Architecture of the Transformer}\

\subsubsection{Encoder}

The encoder consists of \(N\) identical layers, each comprising two main sub-layers:

Multi-Head Self-Attention Layer: This layer utilizes a multi-head attention mechanism to process the input sequence, enabling the model to learn data representations in different subspaces simultaneously.

Feed-Forward Fully Connected Layer: Following is a simple feed-forward network composed of two linear transformations and an activation function, applied individually to each position.

The output of each sub-layer is passed through a residual connection before being passed to the layer normalization step. This architectural design helps the model maintain stability during training of deep networks, mitigating the vanishing gradient problem.\

\subsubsection{Decoder}

The structure of the decoder is similar to the encoder, but each layer contains three sub-layers:

Masked Multi-Head Self-Attention Layer: Similar to the self-attention layer in the encoder, but with an additional masking operation to prevent the leakage of future positional information.

Encoder-Decoder Attention Layer: This layer allows the decoder to focus on different parts of the input sequence and interacts with the stacked outputs of the encoder through queries at the current position of the decoder.

Feed-Forward Fully Connected Layer: Similar to the encoder, this layer is applied to each position in the sequence.
The detailed working principle of the Transformer model is shown in Figure 5.\
\begin{figure}[h]
    \centering
    \includegraphics[width=1\linewidth, trim=1 1.5cm 1 1.5cm]{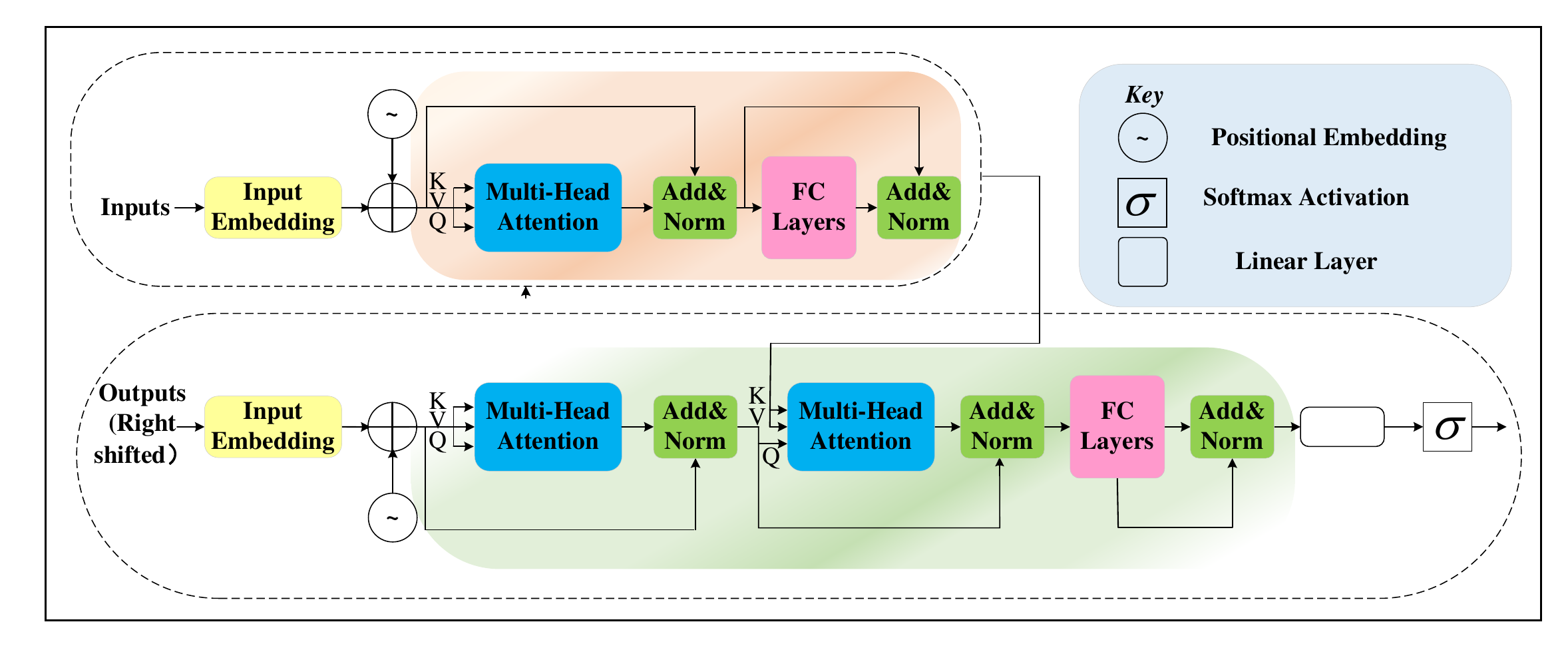}
    \caption{Architecture of the Transformer Model.}
    \label{fig:5}
\end{figure}

\subsection{Derivative Models Based on the Transformer}\

Due to its superior performance and flexibility, the Transformer model has given rise to various derivative models that have achieved significant accomplishments in natural language processing (NLP) and other fields. These derivative models optimize the efficiency and effectiveness of handling specific tasks by modifying the original Transformer architecture or training process. In this chapter, we will explore several major derivative models based on the Transformer, including BERT \cite{devlin2019bert}, GPT \cite{radford2018improving}, RoBERTa \cite{liu2019roberta}, and T5 \cite{raffel2023exploring}. We will analyze their architectural features and discuss how these models might be applied to blockchain technology, particularly in the analysis of smart contracts and the processing of transaction data.\

\subsubsection{BERT(Bidirectional Encoder Representations from Transformers)}

BERT \cite{devlin2019bert} is a model based on the Transformer encoder architecture that employs a bidirectional training strategy to understand language context more comprehensively. The architecture of BERT primarily consists of multiple layers of Transformer blocks, each utilizing multi-head self-attention and feed-forward networks. Unlike the original Transformer, BERT uses a multi-layer encoder stack design without the decoder part.BERT's basic architecture is shown in Figure 6.
\begin{figure}[h]
    \centering
    \includegraphics[width=0.5\linewidth,trim=1.5cm 2cm 1.5cm 1.5cm]{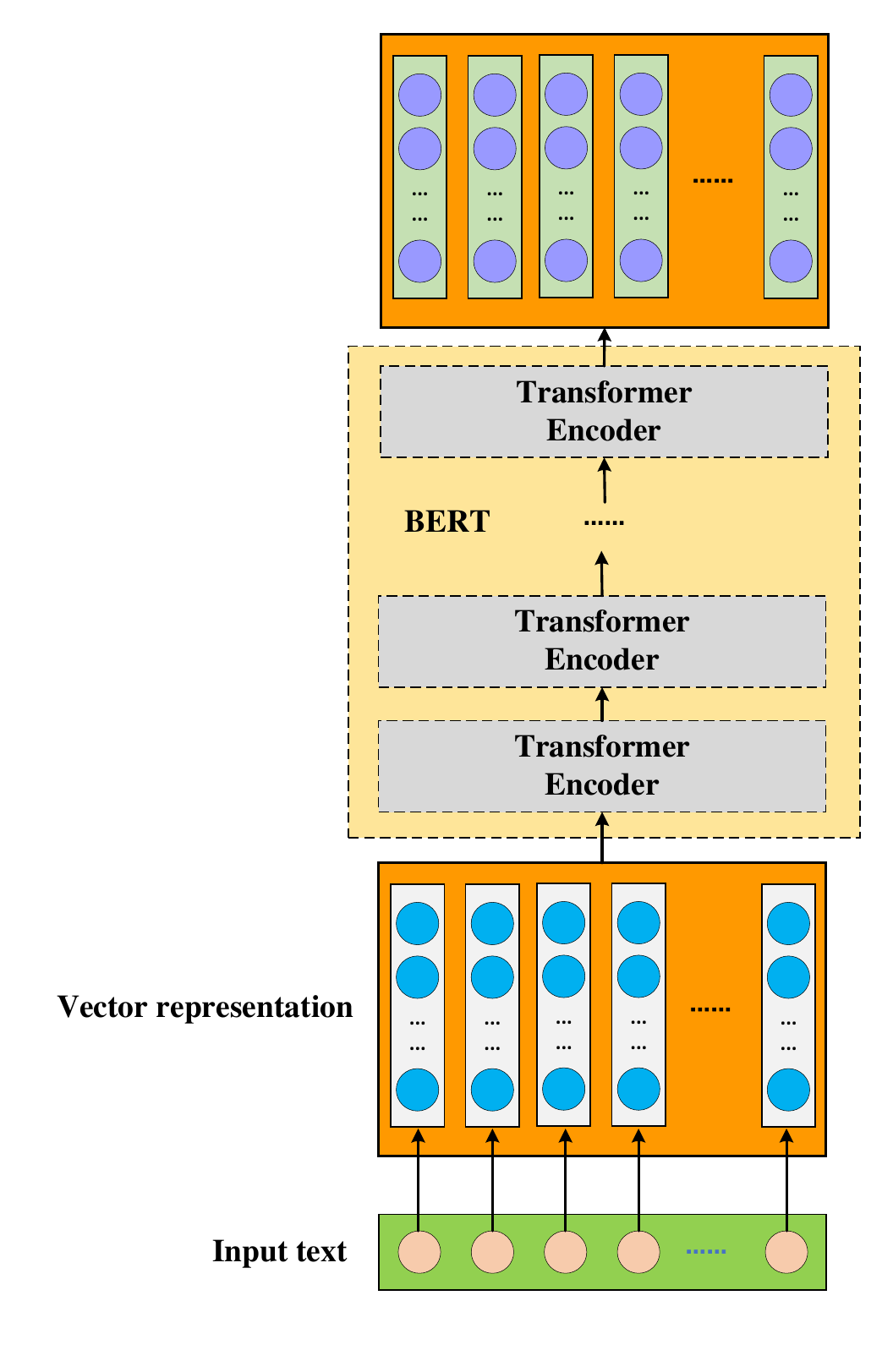}
    \caption{Architecture of the Bert model.}
    \label{fig:6}
\end{figure}

BERT's innovation lies in its pre-training tasks: the Masked Language Model (MLM) and Next Sentence Prediction (NSP). In the MLM task, BERT randomly masks some words in the input sentences and then predicts these masked words, forcing the model to learn deeper language representations. In the NSP task, the model determines whether two sentences are consecutive in the text, helping the model capture relationships between sentences.

BERT's bidirectional context understanding makes it particularly suitable for processing complex smart contract code, where the meaning of contract terms often depends on the surrounding code environment. Applying BERT to smart contract auditing and optimization can enhance contract security, reducing errors and risks during execution.\

\subsubsection{GPT(Generative Pre-trained Transformer)}

GPT \cite{radford2018improving} fundamentally adopts the same architecture as the Transformer decoder, but the overall model structure is designed as a unidirectional architecture aimed at text generation. GPT consists of multiple layers of Transformer decoder blocks, each containing a self-attention layer and a feed-forward network, along with residual connections and layer normalization.The model architecture of GPT-3.5 is shown in Figure 7.
\begin{figure}[h]
    \centering
    \includegraphics[width=0.4\linewidth,trim=1.5cm 1cm 1.5cm 1cm ]{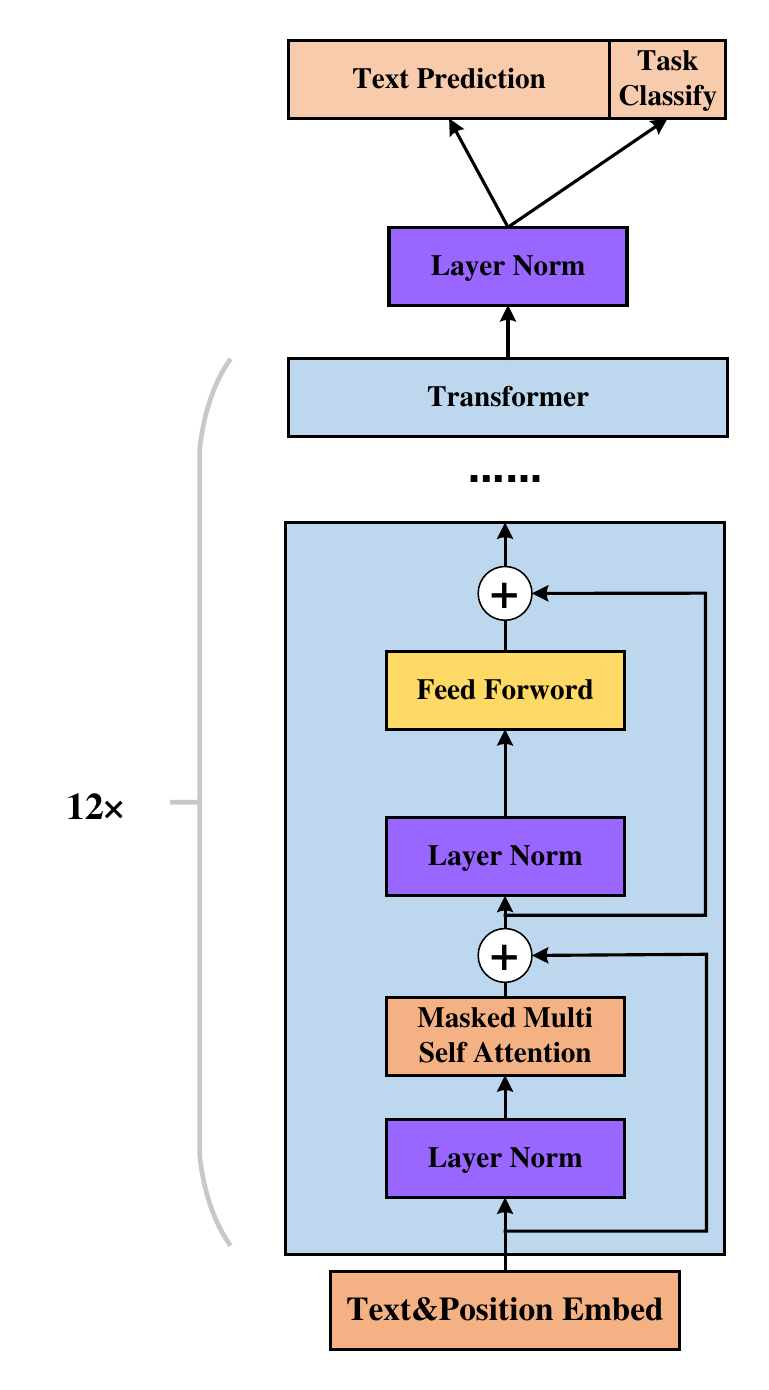}
    \caption{The structural diagram of the GPT-3.5 model.}
    \label{fig:7}
\end{figure}

GPT's training is divided into two phases: unsupervised pre-training and supervised fine-tuning. During the pre-training phase, GPT uses a large corpus of text data to learn general language features through the language modeling task. In the fine-tuning phase, the model is adjusted for specific tasks, optimizing its performance through supervised learning.

GPT's text generation capability can be utilized for automatically generating or modifying smart contract code, particularly when creating contracts that adhere to specific logic and requirements. Additionally, GPT can be used to generate natural language descriptions of blockchain transactions, helping non-technical users understand complex transaction activities.\

\subsubsection{RoBERTa(Robustly Optimized BERT approach)}

RoBERTa \cite{liu2019roberta} builds on BERT by making optimizations while maintaining the architecture of stacked multi-layer Transformer encoders. It retains the core structure of BERT but improves several key training details. The main enhancements in RoBERTa include larger batch sizes, longer training times, more data, and the removal of the Next Sentence Prediction (NSP) task from BERT. These optimizations allow RoBERTa to achieve better performance across various natural language processing tasks.The improvements of RoBERTa based on BERT are shown in Figure 8.
\begin{figure}[h]
    \centering
    \includegraphics[width=1\linewidth,trim=0 1cm 0 1cm]{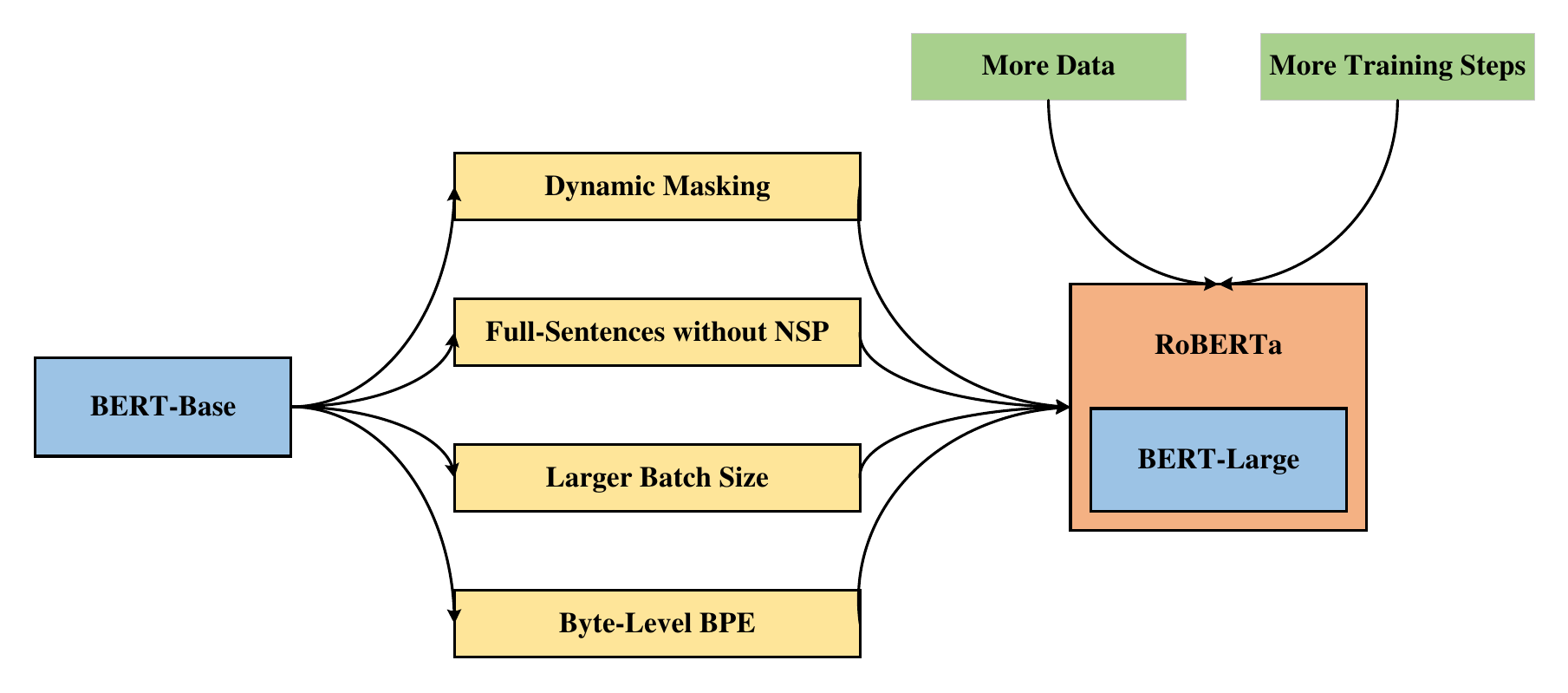}
    \caption{The improvements of RoBERTa based on BERT.}
    \label{fig:8}
\end{figure}

RoBERTa's optimized design makes it well-suited for analyzing blockchain transaction data, such as identifying anomalous transaction patterns or classifying transactions. Its strong language understanding capabilities can also be used to extract and analyze user communication on the blockchain, such as cryptocurrency-related social media posts.\

\subsubsection{T5(Text-to-Text Transfer Transformer)}

The T5 \cite{raffel2023exploring} model adopts an encoder-decoder architecture but unifies all NLP tasks as a "text-to-text" problem. This design allows the model to handle any text generation task in the same manner.

T5's training strategy includes pre-training and fine-tuning phases. During the pre-training phase, T5 uses an improved masked language modeling task, where parts of the input text are masked and then predicted. The fine-tuning phase adjusts the model for specific tasks to optimize performance.

T5's flexibility enables it to be used for automatically generating blockchain monitoring reports and analytical summaries, as well as converting complex blockchain data into easy-to-understand text formats. Additionally, T5 can be utilized to enhance the generation and maintenance of smart contract documentation, improving readability and user-friendliness.\

\section{Fundamentals of Blockchain Technology}
\subsection{Overview of Blockchain Technology}\

Blockchain technology was first proposed by Satoshi Nakamoto in 2008 to support Bitcoin—a decentralized digital currency system. This technology enables data to be stored and managed in a distributed manner across a global network of nodes, without the need for a centralized control authority. The foundation of blockchain is a type of distributed ledger technology, where transaction data is grouped into data structures called "blocks." Each block is linked to the previous block through a cryptographic algorithm, forming a continuous chain. This structure ensures the immutability and permanent record of data, providing unparalleled security. To explore the application of Transformer in blockchain, it is essential to understand its underlying structure and its role in data validation, processing, and security.

\subsection{The Transaction Mechanism of Blockchain}\

One of the core features of blockchain is its transaction mechanism \cite{antonopoulos2023mastering}. Each transaction must be verified by nodes in the network using a consensus algorithm. The verification process includes confirming the validity of the transaction and packaging it into a new block \cite{Li2017Proof,Regnath2018LeapChain:,Nasrulin2018A,Li2021Blockchain-based}. Once a block is accepted by the network, it is added to the blockchain, a process that involves linking the hash of the new block to the hash of the previous block, ensuring the irreversibility of the chain and the immutability of the data. The core of this process is data processing and synchronization, where the application of Transformer can improve the speed and efficiency of data processing, especially in scenarios involving a large number of transactions and data verification \cite{Jayabalan2021A}.

\subsection{Smart Contract}\

Smart contracts are code that automatically executes contract terms, stored on a blockchain. Smart contract code executes automatically when predefined conditions are met, without the need for third-party intermediaries, reducing transaction costs and time \cite{cong2019blockchain}. This is particularly valuable in automating contract execution and reducing errors \cite{Kim2020Analysis,Wang2022An,Alikhani2021Regulating}. Leveraging Transformer for the analysis and optimization of smart contracts can help predict contract behavior and potential execution issues, especially as contract logic becomes increasingly complex \cite{Charlier2017Modeling,Kushwaha2022Ethereum,Zhou2023Security}.The working mechanism of smart contracts is shown in Figure 9.
\begin{figure}[h]
    \centering
    \includegraphics[width=0.9\linewidth,trim=1cm 1.5cm 1cm 1cm]{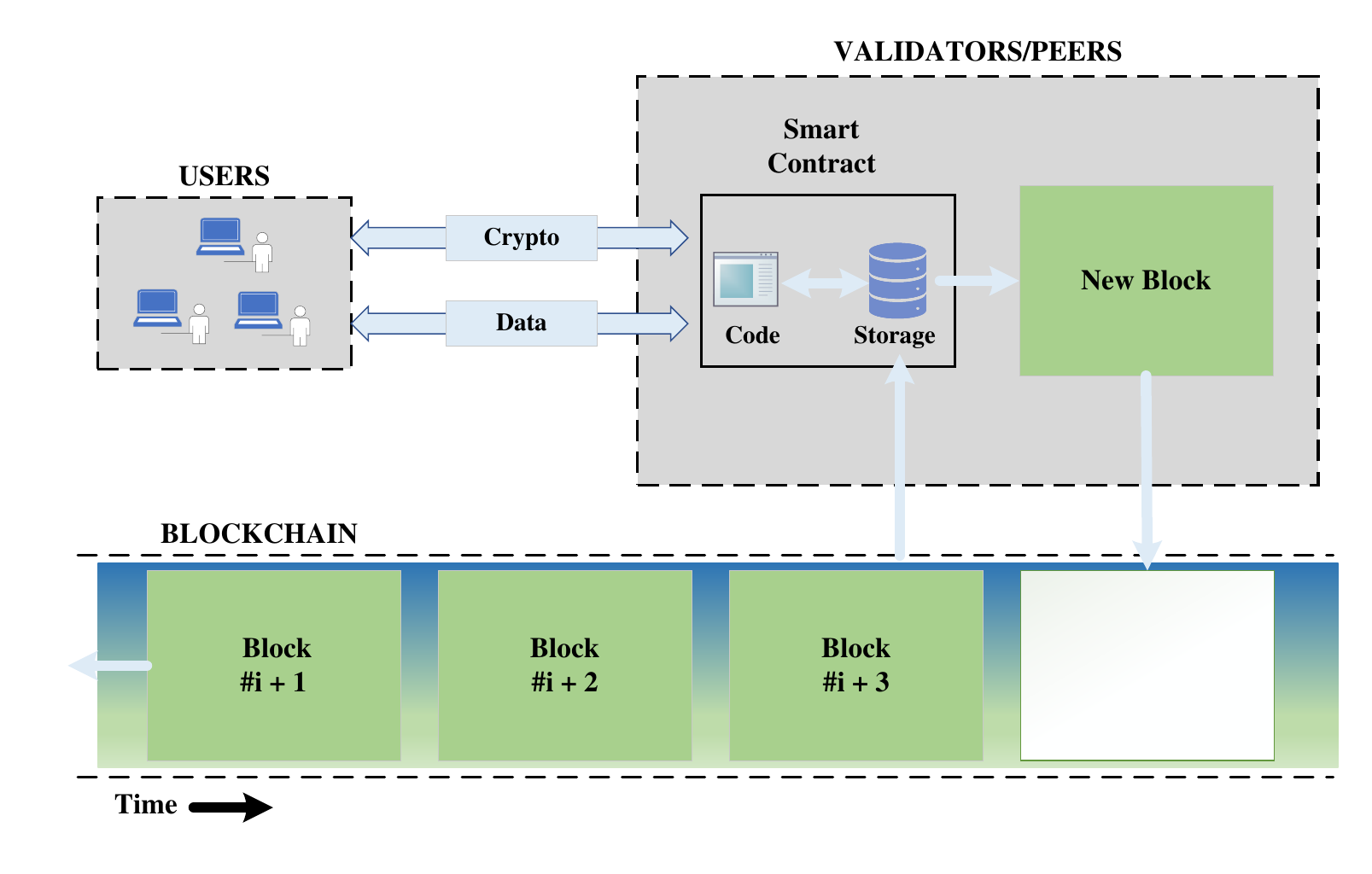}
    \caption{An overview of a smart contract.}
    \label{fig:9}
\end{figure}

Smart contracts are widely used in financial transactions, supply chain management, automated legal processes, and identity verification, among other fields \cite{Prause2019Smart,Dolgui2020Blockchainoriented,Trautmann2020Smart,Aejas2021Effective}. Through smart contracts, contract execution can be ensured to be fair and transparent without the need for human intervention. Applying Transformer to these scenarios can enhance the automatic analysis and execution capabilities of contract terms through natural language processing and pattern recognition, especially in analyzing contract execution results and optimizing contract parameters \cite{Dosovitskiy2020An,Amin2021T2NER:,Gabriel2023Leveraging,Liang2023Human}.

\subsection{Cryptocurrency}\

Cryptocurrency is a digital currency based on blockchain technology, which uses encryption algorithms to protect transaction security, prevent fraud, and avoid double spending. The emergence of cryptocurrency has fundamentally changed the way money is stored and transacted, making cross-border transactions faster and more cost-effective \cite{Yu2022Technology,Ghosh2020Security}.

Cryptocurrency has brought new investment tools and value storage methods to the global financial market \cite{baur2018bitcoin,corbet2019cryptocurrencies}. By using the Transformer model, trading algorithms can be optimized, enhancing the accuracy of market predictions, and playing an important role in cryptocurrency trading and market analysis \cite{Sanju2023Stock-Crypto-App}.

\section{Transformer in Blockchain Applications}

The Transformer model, with its self-attention mechanism, is capable of capturing long-range dependencies in data, making it highly suitable for processing sequential data. Since blockchain data often involves complex transaction records and contract logic, which are a form of time series data, Transformer is considered a powerful tool for solving blockchain data analysis problems.

This chapter will explore the current research status of the Transformer model in several key application areas in blockchain through analyzing existing literature, including anomaly detection, smart contract vulnerability detection, cryptocurrency prediction and trend analysis, and code summarization generation. We will evaluate the effectiveness of the Transformer model in these areas, discuss its ability to address existing problems, and explore how it can improve and enhance the security and efficiency of blockchain technology through its unique processing mechanism.

Through a comprehensive analysis of these areas, this chapter aims to comprehensively demonstrate the potential application of the Transformer in blockchain technology, assess the challenges it faces, and look forward to its future development direction. This will not only provide in-depth insights for academic research but also serve as a theoretical basis and practical guidance for technical development and strategy formulation in practical applications.

\subsection{Anomaly Detection}\

\subsubsection{Background and Objectives}\

Anomaly detection in the blockchain domain involves identifying and addressing behaviors that deviate from normal operational patterns, which is crucial for maintaining the security and integrity of blockchain networks. This detection technology primarily aims to automatically identify anomalous behaviors that may indicate security threats, such as double spending or 51\% attacks \cite{signorini2018bad}. Anomaly detection is particularly important in cryptocurrency transactions, as it helps to detect and prevent fraudulent activities by identifying unusual behaviors in transaction patterns \cite{shayegan2022collective}.

Moreover, anomaly detection techniques significantly enhance the reliability of blockchain networks by ensuring they can withstand external attacks and internal errors. This involves using machine learning algorithms and data analysis techniques to monitor network behavior, promptly identifying and responding to abnormal activities, thus maintaining the system's normal operation and data integrity \cite{ahmad2021anomaly}. Effective anomaly detection mechanisms can minimize system downtime caused by attacks or other disruptions, thereby improving the operational efficiency and responsiveness of blockchain networks \cite{kim2021anomaly}.A detailed taxonomy of these anomalies can be visualized in Fig.10.

\begin{figure}[h]
    \centering
    \includegraphics[width=1\linewidth,trim=0cm 3cm 0cm 3.5cm]{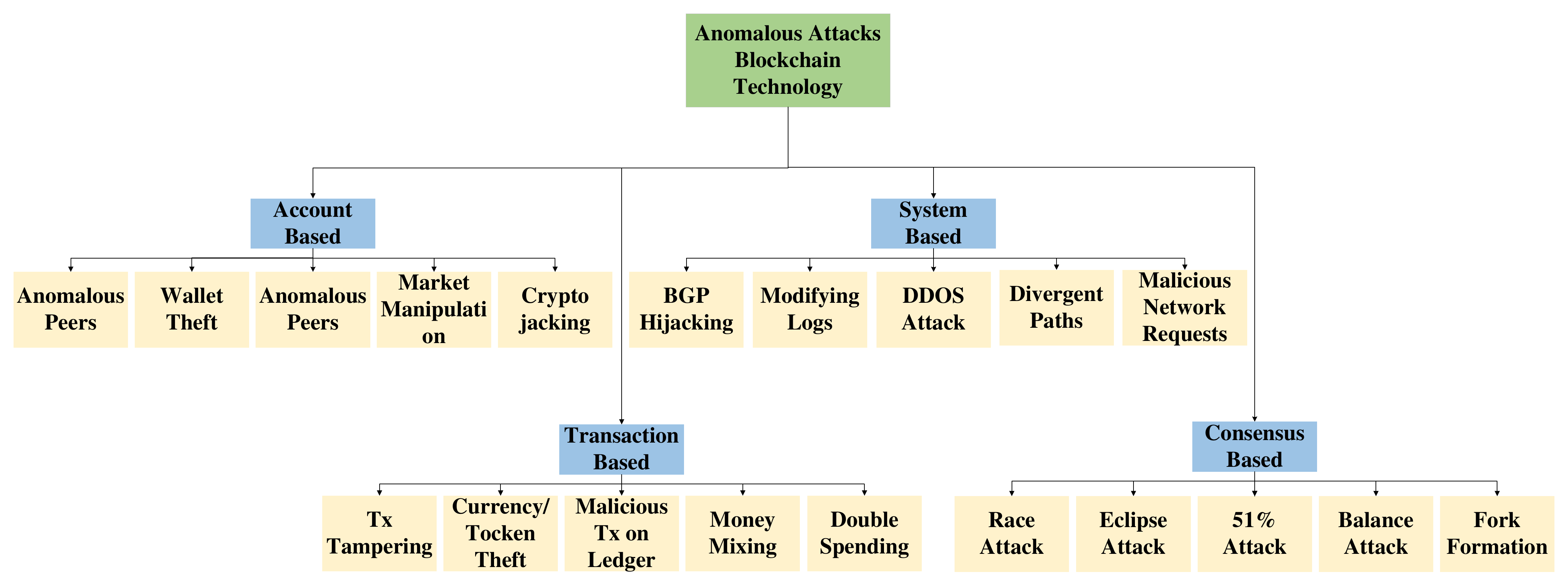}
    \caption{Classification of Anomalous Attacks on Blockchain Technology}
    \label{fig:10}
\end{figure}

Anomaly detection in the blockchain domain is a critical technology that not only enhances the security and resilience of blockchain systems against attacks but also improves overall operational efficiency and reliability by preemptively identifying and addressing anomalies. This is essential for protecting user assets and transaction security, preventing potential financial losses, and ensuring the healthy development of blockchain networks.

\subsubsection{Applications and Limitations of Traditional Methods}\

In the field of blockchain anomaly detection, numerous studies have employed various traditional methods to identify and prevent potential security threats within blockchain systems. These studies encompass a wide range of techniques, from metadata analysis to network monitoring, and from machine learning to data mining, demonstrating the diverse applications of anomaly detection in blockchain security.

For instance, Matteo Signorini and colleagues have explored blockchain metadata for identifying abnormal activities through projects like "BAD" \cite{signorini2018bad} and "ADvISE" \cite{signorini2018advise}. These systems highlight the immutability and transparency of blockchain technology, providing robust tools for detecting potential security threats. Similarly,ref \cite{kim2021anomaly} enhances real-time accuracy in anomaly detection through network traffic monitoring, utilizing advanced network monitoring technologies to analyze data flows and capture subtle abnormal behaviors in complex network environments.

Ref \cite{sayadi2019anomaly} have applied One-Class Support Vector Machines (OCSVM) and K-means clustering algorithms to detect anomalies in cryptocurrency transactions, showcasing the application of machine learning techniques in handling specific types of financial data.Ref \cite{saravanan2023tsi}further explore the efficacy of different machine learning algorithms in blockchain anomaly detection, emphasizing the potential of federated learning and unsupervised learning in protecting data privacy and handling unlabeled data.

Moreover, ref \cite{morishima2021scalable} investigates the performance improvement of anomaly detection using GPU acceleration . This study present new methods for managing anomaly detection tasks in high data throughput environments, which are crucial for large-scale blockchain applications requiring real-time processing.

Ref \cite{ide2018collaborative} highlights the application of smart contracts in collaborative anomaly detection. By automating anomaly detection tasks on blockchain platforms, smart contracts enhance system automation and efficiency.Ref \cite{liang2021data} explore the role of data fusion in improving detection accuracy, leveraging the traceability and transparency of blockchain to enhance the performance of collaborative anomaly detection.

Ref \cite{ofori2021topological} utilize topological methods to detect structural anomalies in dynamic multilayer blockchain networks. By applying topological data analysis (TDA) techniques, they analyze global structural changes in blockchain networks to identify abnormal patterns, making this approach suitable for monitoring the evolution of large-scale networks and effectively capturing unconventional behaviors resulting from network attacks or misconfigurations. Additionally,ref \cite{voronov2021scalable} investigate the use of "sketch" techniques to improve the scalability and efficiency of blockchain anomaly detection. Sketching, a data simplification method, accelerates processing speed by reducing dataset complexity while retaining crucial structural information.

While traditional blockchain anomaly detection methods such as machine learning, network monitoring, data mining, and topological analysis provide effective tools and strategies, they still have some clear limitations compared to the application of advanced deep learning models like Transformers:\

Feature extraction dependency: Traditional methods, such as OCSVM and K-means clustering used in ref \cite{sayadi2019anomaly}, rely on manual feature engineering. These methods require experts to predefine which features are important, limiting the model's ability to adapt to new scenarios. In contrast, Transformers can automatically learn complex features from data without human intervention, providing greater adaptability and flexibility.\

Recognition ability for complex patterns: Traditional algorithms such as machine learning and data mining techniques perform well when dealing with linear or relatively simple nonlinear relationships, but may be inadequate in capturing complex, dynamically changing data patterns (such as transaction patterns in blockchain networks). For example, while the topological method used by ref \cite{ofori2021topological} can handle complex network structures, it may still be less effective than Transformers in capturing and understanding long-term data dependencies. Transformers, through their self-attention mechanism, can effectively handle long-range data dependencies and complex data relationships, which is crucial for tasks such as analyzing the continuity of transaction data.\

Real-time data processing and scalability: While ref \cite{kim2021anomaly} based on network traffic monitoring demonstrates good real-time performance, it may encounter performance bottlenecks when dealing with large-scale data. The parallel processing capability of Transformers makes them more efficient in handling large-scale real-time datasets, particularly in security monitoring scenarios that require rapid responses.\

Adaptability and automatic updates: Traditional machine learning models typically require periodic retraining or manual adjustments after deployment to adapt to new data or attack patterns. For example, while the federated learning strategy used by ref \cite{saravanan2023tsi} allows for updates across nodes, it often requires retraining to adapt to new data or attack patterns. Transformer models, on the other hand, can automatically adjust and optimize through continuous learning processes, improving the model's adaptability in dynamic environments. Transformers can adapt to new situations through continuous training processes, especially when using technologies like federated learning, which can update and optimize across multiple nodes without exposing raw data.

Deep semantic understanding: Traditional methods may have limitations in parsing the deep semantic and contextual relationships of transactions or behaviors, especially when dealing with complex transactions of cryptocurrencies like Bitcoin. Transformer models can understand and analyze the deep semantics and contexts in long sequence data, providing the possibility to detect more complex fraudulent behaviors and anomalies.

\subsubsection{Research Progress Using Transformer Models}\

Ref \cite{song2023anomaly} proposed an innovative deep learning framework specifically designed to address anomaly detection in decentralized finance (DeFi) systems. This framework combines Variational Autoencoders (VAE) and Transformers to optimize the handling of time-series data. Specifically, the model first utilizes the encoder of the VAE to encode time-series data into low-dimensional embedding vectors, which helps the model capture critical local information in the short term. These embedding vectors are then fed into the Transformer, which analyzes the long-term dependencies among these embeddings through its self-attention mechanism, generating context-enhanced embedded outputs. This output is subsequently used to reconstruct the original data. By comparing the original data with the reconstructed data, the model effectively identifies anomalous patterns in the data. 

The advantage of this technical architecture lies in its combination of the VAE's powerful feature extraction capabilities and the Transformer's ability to analyze long-term sequences, thereby improving the model's accuracy and efficiency in handling dynamic and complex DeFi transaction data. The Transformer's self-attention mechanism is particularly suited to capturing complex temporal dependencies in financial markets, which is crucial for identifying and preventing anomalous behavior in highly volatile markets. This method not only improves the accuracy of anomaly detection but also enhances the transparency of decentralized finance platforms and the security of user assets. The model demonstrated high accuracy on the Olympus DAO dataset. By analyzing four different anomaly cases, the model successfully identified various types of anomalous behaviors, ranging from structural changes to malicious attacks, proving its effectiveness in detecting and responding to complex anomalies.\

Ref \cite{liu2022blockchain} proposed a technical architecture leverages Heterogeneous Graph Transformer Networks (HGTNs) to deeply probe for abnormal behaviors in smart contracts on Ethereum. The implementation begins with constructing a complex Heterogeneous Information Network (HIN), which includes elements such as smart contracts, user accounts, and transactions as nodes in the network. The interactions between these nodes, such as function calls and fund transfers, form the edges connecting them. Each type of node has its unique attributes, which characterize different contract features and interaction behaviors within the network.
A critical step in the process is employing the Transformer model to handle this complex network. The Transformer, with its self-attention mechanism, can effectively identify and weigh the relationships between nodes in the network, which is crucial for capturing subtle differences in contract behavior. This mechanism allows the model to process a large number of nodes in parallel, capturing key information by adaptively adjusting attention distribution, thereby enhancing the model's ability to recognize abnormal patterns. The Transformer's self-attention layers dynamically identify the most influential nodes and paths in fraud detection, improving detection accuracy and significantly increasing processing speed. Additionally, the hierarchical structure and parallel processing capabilities of the Transformer make it highly suitable for real-time data monitoring, enabling immediate responses to abnormal behaviors in smart contracts, thus providing robust security for the Ethereum platform.

Experimental results indicate that the proposed approach achieves an accuracy of 92\%, a recall of 89\%, and an F1 score of 90.5\%. Compared to traditional rule-based and simple machine learning methods, this Transformer model shows significant improvements. Particularly in handling mixed data types and large-scale datasets, the Transformer's self-attention mechanism effectively captures long-term dependencies and complex patterns, offering greater flexibility and accuracy.This technical architecture based on heterogeneous graphs and Transformer networks provides an effective solution for detecting anomalies in smart contracts. It not only allows for deep analysis of complex blockchain data but also enhances monitoring and security performance on blockchain platforms like Ethereum while maintaining high efficiency and accuracy.

Ref \cite{batool2022block} described a complex technical architecture that integrates Federated Learning (FL), Split Learning (SL), Transformer models, and blockchain technology. This architecture aims to enhance real-time anomaly detection capabilities for IoT device data streams while ensuring data privacy and reducing the computational burden on clients. The architecture leverages federated learning to enable multiple clients to collaboratively train a global deep learning model without directly sharing data. Split learning allows for the distribution of computational tasks between the client and the central server. Clients perform preliminary computations and send partial data to the server for further processing, thus reducing the burden on resource-constrained clients and protecting data privacy. The introduction of the Transformer model, with its self-attention mechanism, optimizes the processing of time-series data by effectively capturing complex patterns and long-term dependencies, significantly improving the accuracy and efficiency of anomaly detection. The entire training and updating process is supported by blockchain technology, ensuring transparency and data immutability. The use of smart contracts enhances the system's automation and security. 

Experimental results show that Block-FeST achieved an accuracy of 86\% across multiple test sets. This accuracy is competitive with other centralized and decentralized benchmark models, while also providing the additional benefits of decentralization and client computation offloading. Block-FeST not only demonstrated high accuracy and efficiency in anomaly detection during experiments but also showcased its potential as an efficient and reliable real-time anomaly detection system by reducing client burden, enhancing data privacy protection, and improving system security. This sets new standards for data security and real-time processing in IoT environments.

Ref \cite{xu2023illegal} detailed an advanced method for detecting illegal accounts on Ethereum using Heterogeneous Graph Transformer Networks (HGTNs). This method first constructs a Heterogeneous Information Network (HIN) comprising entities such as accounts, transactions, smart contracts, and blocks. This comprehensive network allows the model to integrate various types of data and relationships, providing a thorough understanding of the Ethereum ecosystem.Next, the method employs a Graph Transformer model to process this complex network. The Graph Transformer automatically learns and optimizes multi-level associations between nodes (e.g., through meta-paths) to extract critical features, enhancing the model's ability to recognize complex patterns hidden within the data. Through this approach, the system can effectively identify and classify normal and illegal accounts, demonstrating significant advantages over traditional methods and simple machine learning models in handling large-scale and complex network data. Additionally, the parallel processing capability of the Transformer model significantly improves processing speed and efficiency, making this solution not only promising for scientific research but also widely applicable and efficient in real-world applications.

Experimental results show that this method not only improves detection accuracy (achieving an accuracy rate of 95.57\%) but also significantly enhances processing speed and efficiency, confirming the advantages of Transformer models in capturing network data features and improving illegal account detection performance.
This graph Transformer-based method not only enhances the model's understanding of complex network behaviors but also improves the ability to identify and prevent fraudulent activities in actual blockchain environments, providing an effective technical means for blockchain security.

Ref \cite{chen2021improving} presented a method that leverages deep learning techniques to enhance the detection of Ponzi schemes in Ethereum smart contracts. The method begins by structurally processing the source code of smart contracts using Abstract Syntax Trees (ASTs) to convert the source code into a more analyzable form. It then employs Structure-Based Traversal (SBT) to transform the AST into a sequence of code tokens, preserving the semantic structure of the code.In the feature extraction phase, the paper introduces a Multi-Channel TextCNN to extract key local features from the transformed code tokens. The multi-channel design allows the model to capture different levels of information from the code through various receptive fields, leading to a more comprehensive understanding of the code's functionality and behavior patterns.
To address long-range dependencies in the source code—frequent in contract logic that spans multiple code blocks—a Transformer model is introduced. The Transformer, with its self-attention mechanism, effectively captures and analyzes complex interactions between different parts of the code, which is crucial for deeply understanding the intricate logic of Ponzi schemes. This combination of local feature recognition and global dependency analysis significantly enhances the model's capability to identify Ponzi schemes.
For performance optimization, a cost-sensitive loss function is used to handle the imbalance in the training data, ensuring high sensitivity of the model to rare but dangerous fraudulent activities.

The core advantage of the Transformer model lies in its self-attention mechanism, which effectively identifies long-range dependencies and complex relationships within the code. This is essential for understanding the distributed logic of Ponzi schemes spread across multiple parts of smart contract code. Additionally, the parallel processing capability of the Transformer significantly improves the model's efficiency in handling large-scale data, making the analysis process faster and more accurate.Experimental results show that this new model achieves a 95.57\% accuracy rate in detecting Ponzi schemes, significantly outperforming traditional methods. The application of this technology not only improves the performance of Ponzi scheme detection but also provides an efficient solution for fraud detection in blockchain technology.

Ref \cite{zhou2022logblock} proposed a deep learning-based architecture specifically designed for anomaly detection in permissioned blockchains. This framework starts by collecting operation logs from the blockchain network and then segmenting these logs into multiple log blocks based on time windows, with each log block representing all the log information within a specific time interval. This sequential processing helps maintain the temporal continuity of the logs and facilitates subsequent pattern analysis.In this approach, each log block is fed into a hybrid neural network composed of a Convolutional Neural Network (CNN) and a Transformer model. The CNN component first processes the log blocks to extract local features from the log data, reflecting the immediate state of blockchain operations and potential signs of anomalies. These local features are then passed into the Transformer model, which, through its self-attention mechanism, analyzes and identifies patterns and dependencies across the entire log sequence. The Transformer model's ability to capture complex behavioral patterns across log blocks is particularly crucial, as these patterns are often key indicators of anomalies that simpler models might miss.

The application of the Transformer model provides excellent long-range dependency handling capabilities, enabling the model to effectively identify non-linear and non-explicit anomaly patterns in large log data, such as fraudulent activities hidden within complex transactions. Additionally, the parallel computing capabilities of the Transformer significantly speed up processing, making real-time or near-real-time anomaly detection possible. This is extremely valuable in high-demand blockchain environments, where rapid response and resolution of anomalous events are essential for enhancing overall system security and responsiveness.

Ref \cite{wang2024research} proposed a 1D SA-Inception model, which combines one-dimensional convolutional neural networks (1D CNN) and Transformer structures to achieve efficient blockchain abnormal transaction detection. The model first utilizes 1D CNN for preliminary data processing and then introduces an Inception structure to extract data features in a multi-scale manner, effectively enhancing the model's understanding of different data granularities. The core technical innovation lies in incorporating the self-attention mechanism from the Transformer, which calculates the mutual influence and importance of data points, allowing the model to perform parallel processing without sacrificing contextual information, significantly improving computational efficiency. Additionally, the self-attention mechanism optimizes the model's feature representation capability, enhancing its sensitivity and accuracy in identifying abnormal patterns.

By integrating the self-attention mechanism of the Transformer, the 1D SA-Inception model demonstrates significant advantages in detecting blockchain abnormal transactions. Experimental data shows that the model achieves an AUC value of 91.05\%, a G-mean of 84.95\%, and an F1 score of 83.52\%, which are notably better than traditional 1D CNN models. The self-attention mechanism enables the model to process data in parallel and capture long-range dependencies, significantly enhancing processing speed and efficiency. Moreover, this mechanism strengthens the recognition of complex patterns by weighting key features, making it particularly sensitive and accurate in detecting rare abnormal behaviors in highly imbalanced datasets. Overall, the application of Transformer technology in the 1D SA-Inception model not only improves predictive performance but also enhances the model's generalization ability, making it a powerful tool for handling complex financial datasets.
\begin{table*}[h]
\caption{SUMMARY OF TRANSFORMER-BASED METHODS FOR ANOMALY DETECTION.}
\label{tab:my-table}
\begin{tabular}{clccc}
\textbf{Ref.} & \textbf{Year}            & \textbf{Type}           & \textbf{Detection Object}                                                           & \textbf{Major Characteristic}                                                                                                                                            \\ \hline
{[}86{]}      & 2021                     & Sequence Representation & \begin{tabular}[c]{@{}c@{}}Smart Contract\\ Source Code\end{tabular}                & \begin{tabular}[c]{@{}c@{}}Parse AST to SBT sequences, extract features, and \\ capture dependencies with a Transformer.\end{tabular}                                    \\
{[}83{]}      & 2022                     & Graph Representation    & \begin{tabular}[c]{@{}c@{}}On-Chain Data and\\ Smart Contract Bytecode\end{tabular} & \begin{tabular}[c]{@{}c@{}}Use HIN to extract features, convert code with\\ Doc2Vec, and learn meta-paths for anomaly detection.\end{tabular}                            \\
{[}84{]}      & 2022                     & Sequence Representation & Network Traffic Data                                                                & \begin{tabular}[c]{@{}c@{}}Extract KPI features, use FL, SL, and \\ Transformer, and ensure privacy with blockchain.\end{tabular}                                        \\
{[}87{]}      & \multicolumn{1}{c}{2022} & Sequence Representation & On-Chain Data                                                                       & \begin{tabular}[c]{@{}c@{}}Extract log sequence features from blockchain logs \\ using CNN andTransformer models for anomaly detection.\end{tabular}                     \\
{[}82{]}      & 2023                     & Graph Representation    & On-Chain Data                                                                       & \begin{tabular}[c]{@{}c@{}}Extract 12 features from on-chain data, including OHM transfers, \\ transaction counts, and active users, for anomaly detection.\end{tabular} \\
{[}85{]}      & 2023                     & Graph Representation    & Ethereum Account                                                                    & \begin{tabular}[c]{@{}c@{}}Construct an account-centric HIN, extract \\ features, and use GTN to detect illegal accounts.\end{tabular}                                   \\
{[}88{]}      & 2024                     & Sequence Representation & On-Chain Data                                                                       & \begin{tabular}[c]{@{}c@{}}Extract transaction time, amount, and frequency, and detect \\ anomalies using multi-scale features and self-attention.\end{tabular}         
\end{tabular}
\end{table*}
\subsection{Smart Contract Vulnerability Detection}\

\subsubsection{Background and Objectives}\

Smart contracts are an essential component of blockchain technology, enabling the automatic execution of coded contract terms on the blockchain, greatly improving transaction efficiency and transparency. 
\begin{table}[h]
\caption{Classification of vulnerabilities in Ethereum smart contracts}
\label{tab:1}
\begin{tabular}{clllll|clllll}
\multicolumn{6}{c|}{\textbf{Vulnerability Source}}                  & \multicolumn{6}{c}{\textbf{Vulnerability Type}} \\ \hline
\multicolumn{6}{c|}{\multirow{5}{*}{Solidity Code}}                 & \multicolumn{6}{c}{Call to unknown address}     \\ \cline{7-12} 
\multicolumn{6}{c|}{}                                               & \multicolumn{6}{c}{Gasless Send}                \\ \cline{7-12} 
\multicolumn{6}{c|}{}                                               & \multicolumn{6}{c}{Reentrancy attack}           \\ \cline{7-12} 
\multicolumn{6}{c|}{}                                               & \multicolumn{6}{c}{Exception disorders}         \\ \cline{7-12} 
\multicolumn{6}{c|}{}                                               & \multicolumn{6}{c}{Type Casts}                  \\ \hline
\multicolumn{6}{c|}{\multirow{3}{*}{Blockchain}}                    & \multicolumn{6}{c}{Generating randomness}       \\ \cline{7-12} 
\multicolumn{6}{c|}{}                                               & \multicolumn{6}{c}{Unpredictable state}         \\ \cline{7-12} 
\multicolumn{6}{c|}{}                                               & \multicolumn{6}{c}{Time Constraints}            \\ \hline
\multicolumn{6}{c|}{\multirow{3}{*}{EVM(Ethereum Virtual Machine)}} & \multicolumn{6}{c}{Immutable bugs}              \\ \cline{7-12} 
\multicolumn{6}{c|}{}                                               & \multicolumn{6}{c}{Ether lost in the transfer}  \\ \cline{7-12} 
\multicolumn{6}{c|}{}                                               & \multicolumn{6}{c}{Stack size limit}           
\end{tabular}
\end{table}
This automation enables trust and compliance to be achieved without the need for intermediaries. However, the security issues of smart contracts cannot be ignored. Common vulnerabilities such as reentrancy attacks \cite{he2023detection}, arithmetic overflow \cite{chu2023survey}, timestamp dependence \cite{luo2024scvhunter}, and improper exception handling \cite{vani2022vulnerability} can lead to funds being stolen, data corruption, or contract logic failure, resulting in irreversible economic losses and trust crises. Therefore, developing efficient smart contract vulnerability detection techniques is crucial. Not only does this help prevent potential financial risks and protect user asset security \cite{yang2022formal}, but it also plays a vital role in ensuring the overall security and stability of the blockchain ecosystem, meeting regulatory compliance requirements \cite{tang2021vulnerabilities}, and enhancing the market acceptance and user trust of blockchain technology \cite{wang2023value}.

With the advancement of technology, modern methods such as deep learning and graph neural networks have been widely applied to the security analysis and vulnerability detection of smart contracts \cite{liu2021combining}, effectively improving the accuracy and efficiency of detection, laying a foundation for the healthy development of blockchain technology.As shown in Figure 11, the research literature on smart contract vulnerability detection has steadily increased in recent years.The general classification of smart contract vulnerabilities is shown in Table 1.
\begin{figure}[h]
    \centering
    \includegraphics[width=0.5\linewidth,trim=5cm 9.5cm 5cm 9.5cm]{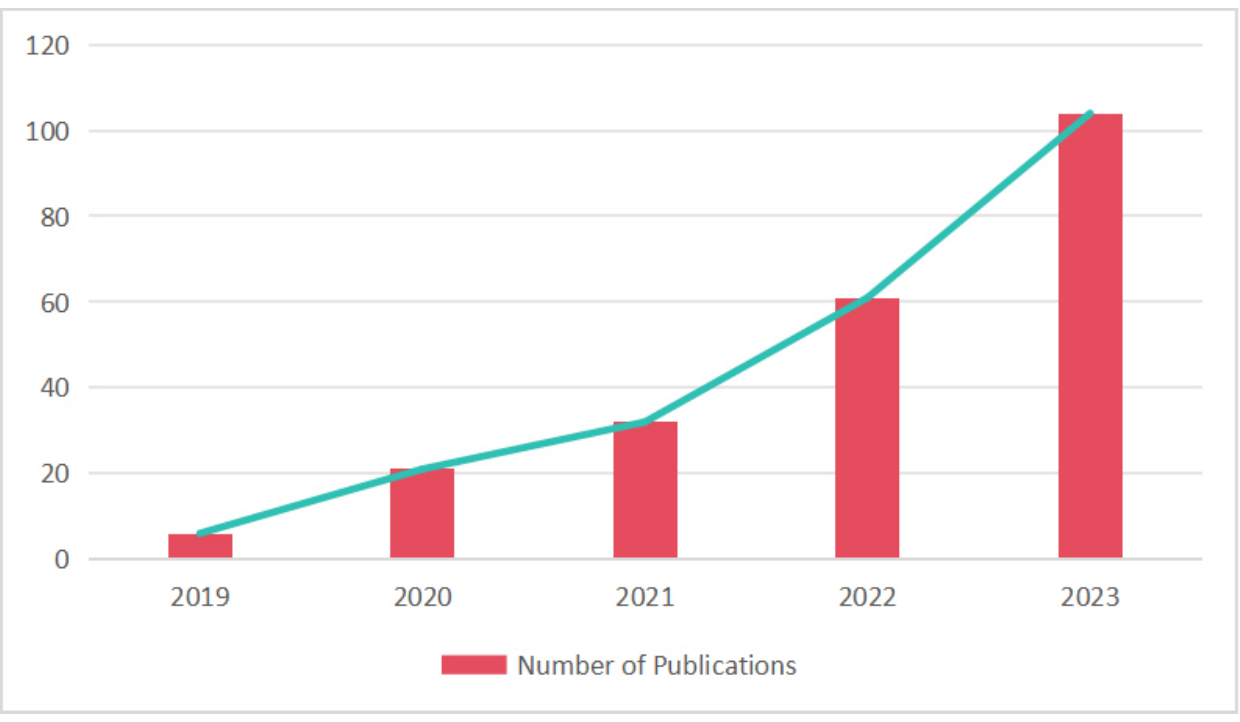}
    \caption{Statistics on the publication of papers on smart contract vulnerability detection.}
    \label{fig:11}
\end{figure}

Exploring smart contract vulnerabilities and their detection techniques is crucial for protecting the blockchain ecosystem from security threats. As blockchain technology continues to evolve, enhancing the reliability and security of smart contracts through advanced detection methods is crucial for their adoption and effectiveness.

\subsubsection{Applications and Limitations of Traditional Methods}\

In the research field of smart contract vulnerability detection, traditional methods have made significant progress using various machine learning techniques such as Graph Neural Networks (GNNs), Abstract Syntax Trees (ASTs), and deep learning, although they have not involved the Transformer model. The following describes the research status and limitations of traditional methods in smart contract vulnerability detection that do not use the Transformer model.

Ref \cite{luo2024scvhunter} and ref \cite{han2022smart}  utilizes Graph Neural Networks to handle the complex relationships and data structures in smart contracts. Ref \cite{luo2024scvhunter} emphasize the identification of critical structures through attention mechanisms, while ref \cite{han2022smart} focuses on revealing potential vulnerabilities in code execution paths through control flow graphs and applies Graph Neural Networks to enhance the efficiency and accuracy of smart contract vulnerability detection. The combination of these two methods provides a strategy for analyzing smart contracts from both macro and micro levels, strengthening the model's ability to understand complex data structures.

Ref \cite{yang2022smart} conducted research using Abstract Syntax Trees (AST) to analyze the static structure of smart contract code, identifying vulnerabilities and defects in the structure. AST is an abstract representation of the source code. As shown in Figure 12,it takes the structure of the code and converts it into a tree. This method, based on the syntax and structure of the code, provides an intuitive and effective technique for vulnerability detection.
\begin{figure}[h]
    \centering
    \includegraphics[width=1\linewidth,trim=0 1cm 0 1cm]{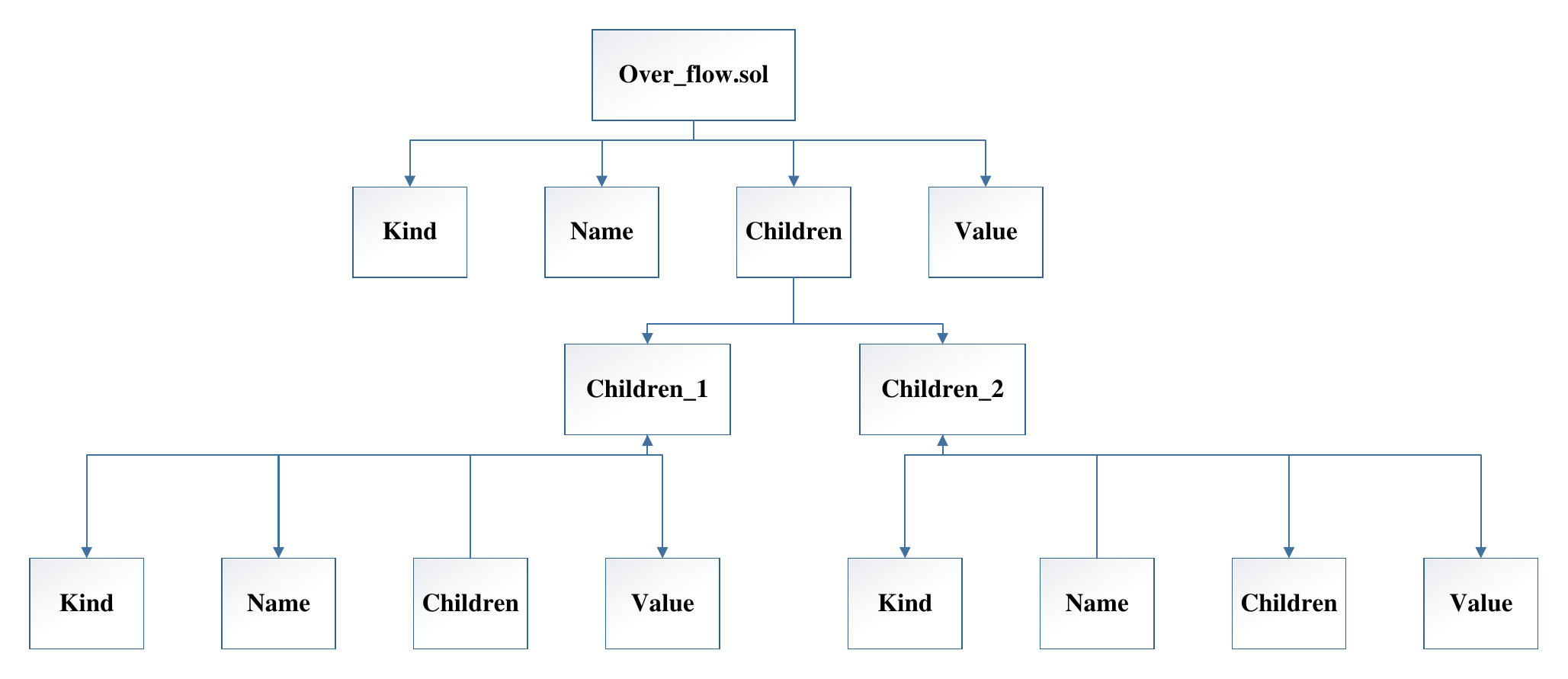}
    \caption{The graph of AST.}
    \label{fig:12}
\end{figure}

Ref \cite{zhang2022smart} and ref \cite{deng2023smart} demonstrated the application of deep learning in smart contract vulnerability detection.Ref \cite{zhang2022smart} Bi-LSTM network identifies complex dependencies and patterns by analyzing time series data,The Bi-LSTM data processing process of one layer of Bi-LSTM is shown in Figure 13;\begin{figure}[h]
    \centering
    \includegraphics[width=0.9\linewidth,trim=0 1.5cm 0 1cm]{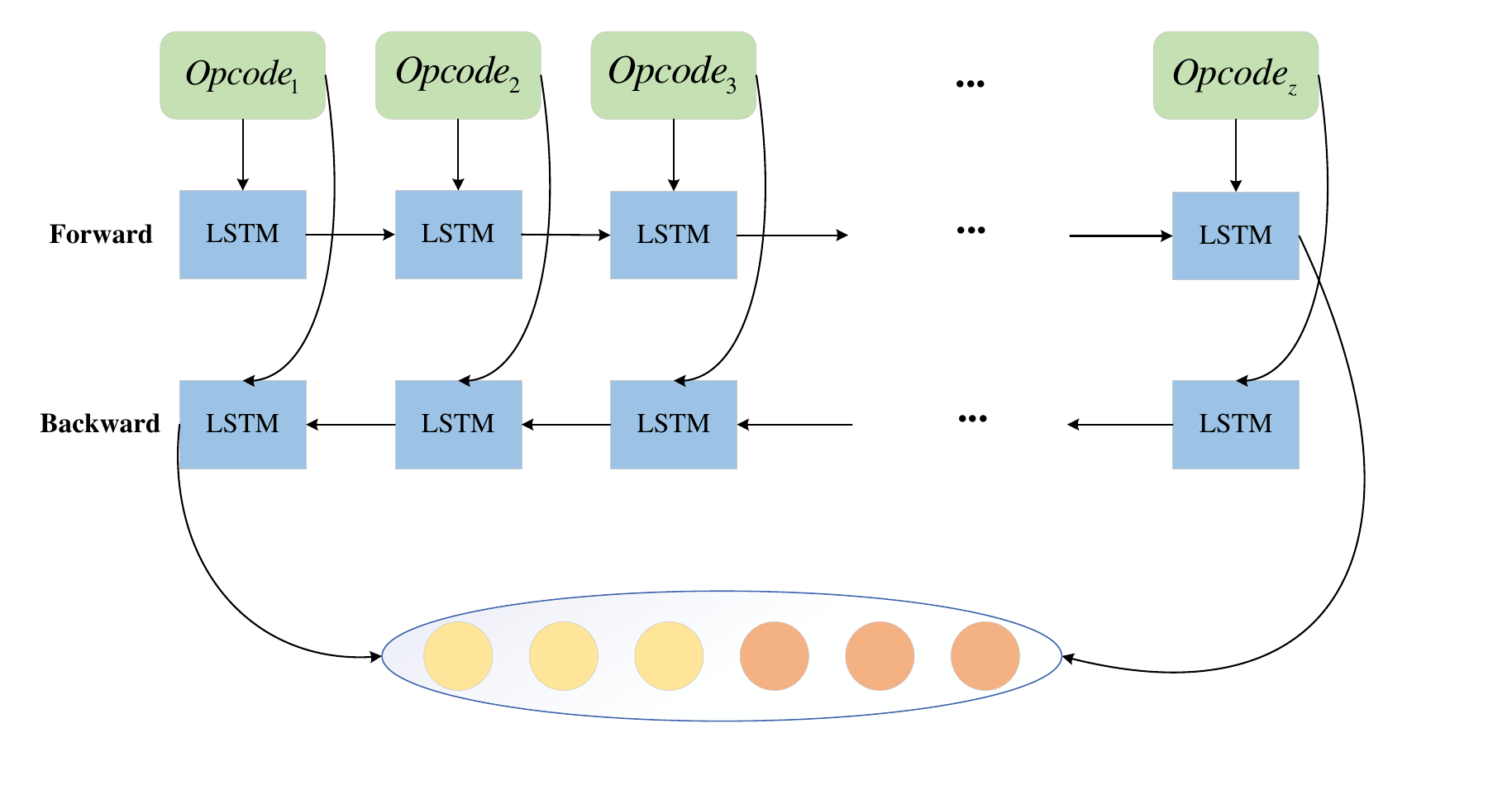}
    \caption{Bi-LSTM data processing process.}
    \label{fig:13}
\end{figure}\
Ref \cite{deng2023smart} multimodal decision fusion method enhances the model's predictive ability by integrating multiple data sources.The rough process of smart contract vulnerability detection based on deep learning is shown in Figure 14.
\begin{figure}[h]
    \centering
    \includegraphics[width=0.8\linewidth,trim=0 1cm 0 1cm]{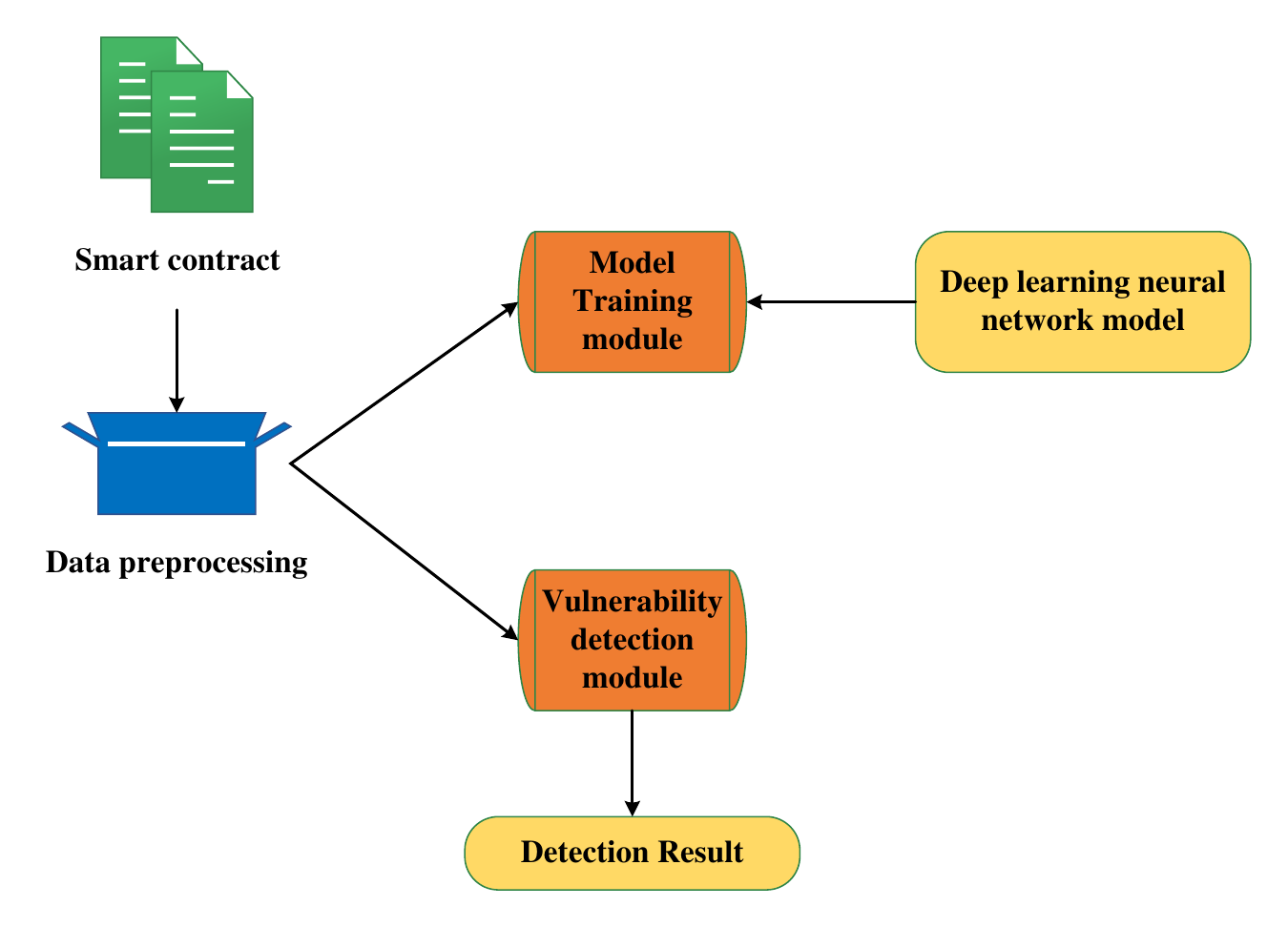}
    \caption{Smart contract vulnerability detection frameworks based on deep learning.}
    \label{fig:13}
\end{figure}

Ref \cite{sui2023opcode} and ref \cite{cao2023sccheck} focus on the dynamic behavior analysis of smart contracts.Ref \cite{sui2023opcode} directly identifies runtime vulnerabilities by analyzing opcodes during execution, while ref \cite{cao2023sccheck} monitors the real-time execution of smart contracts through data flow and attention mechanisms, further enhancing the detection capabilities for dynamic vulnerabilities.

Ref \cite{yang2023improvement} utilized Support Vector Machine (SVM) technology as a powerful classification and prediction tool, which can be combined with deep learning methods to optimize the smart contract vulnerability detection process.Ref \cite{yan2022semantic} focused on semantic analysis, emphasizing the functional intent of the code to identify potential vulnerabilities, enhancing the depth of code analysis. This in-depth analysis of the semantic level of the code complements machine learning methods at the technical level, providing comprehensive security detection.

Ref \cite{wu2023smart} introduced a smart contract vulnerability detection model based on a hybrid attention mechanism. This model combines self-attention and convolutional attention mechanisms to accurately focus on long-distance dependencies and local features in smart contract code, enhancing the ability to identify potential vulnerabilities in complex code structures. This hybrid approach enables the model to comprehensively analyze the structure of smart contracts, effectively improving the accuracy and efficiency of vulnerability detection.

In the field of smart contract vulnerability detection, traditional methods that do not use Transformer models have made some research progress. However, they still face significant limitations:

Limited ability to handle long-distance dependency relationships:
For example, the Bi-LSTM used by ref \cite{zhang2022smart} can handle time dependencies in time series data, but information in long sequences may gradually dissipate. In contrast, the Transformer model can comprehensively capture the dependencies between every element in the entire sequence through its self-attention mechanism, regardless of how far apart they are.

Insufficient global context understanding:
While the research by ref \cite{luo2024scvhunter} and ref \cite{han2022smart} effectively utilizes graph structures to analyze smart contracts, they may not integrate global contextual information as effectively as Transformer models based on self-attention. The Transformer can assess relationships between all inputs in a single forward pass, providing a deeper understanding of context.

Weak parallel processing capability:
Traditional deep learning models,  while combining different learning techniques, may not have the parallel computing capability as strong as Transformer models when processing large-scale datasets. The Transformer is designed to optimize computational efficiency, particularly suitable for modern GPU architectures.

Low adaptability and flexibility:
While the hybrid attention mechanism by ref \cite{wu2023smart} enhances the identification of key features, this approach may not be as flexible as Transformer in adapting to newly emerging attack patterns and complex vulnerabilities. The Transformer model can quickly adapt to new data features through fine-tuning or reconfiguring its attention layers.

Limited real-time dynamic analysis capability:
Although ref \cite{sui2023opcode} and ref \cite{cao2023sccheck} provide real-time monitoring of smart contract execution processes, these methods may face performance bottlenecks when handling large volumes of real-time data streams. The Transformer model can better handle and analyze large-scale real-time data through its efficient self-attention mechanism.

Model generalization ability:
Traditional machine learning methods such as the support vector machine used by ref \cite{yang2023improvement} may perform well on specific training sets but may not generalize well to unseen data or new types of vulnerabilities. The Transformer typically offers stronger generalization ability due to its complex feature learning and global information processing capabilities.

While current traditional methods provide various effective tools for smart contract vulnerability detection, introducing Transformer models may bring significant improvements, particularly in handling complex and large-scale datasets, improving model adaptability and flexibility, and enhancing global context understanding. These advantages make Transformer-based approaches highly promising in future research on smart contract security.

\subsubsection{Research Progress Using Transformer Models}\

Ref \cite{wu2021peculiar} proposed an innovative method for detecting vulnerabilities in smart contracts. This method combines Crucial Data Flow Graph (CDFG) and Transformer-based pre-training models, such as GraphCodeBERT, to enhance the detection capabilities of complex vulnerabilities in Ethereum smart contracts. Starting from the source code, the method first constructs a complete Data Flow Graph (DFG) and then filters out the security-related crucial flow data to form CDFG. This process accurately removes unnecessary information, retaining the most crucial data flows that may lead to vulnerabilities. Then, by applying the Transformer model in deep learning, especially leveraging its self-attention mechanism, the method effectively identifies dependency patterns and potential risks in the code, improving the model's ability to capture long-distance code dependencies. The pre-trained model not only enhances the detection speed but also ensures efficient handling of large-scale code due to its successful application in natural language processing, significantly improving the accuracy and efficiency of vulnerability detection.

Experimental results show that Peculiar achieves higher precision (91.80\%) and recall (92.40\%) in detecting reentrancy vulnerabilities compared to existing methods, significantly improving performance, especially in handling complex code structures and unknown vulnerability patterns, demonstrating strong generalization ability. This method not only improves detection efficiency but also effectively supports the security analysis of large-scale smart contracts through parallel processing capabilities and pre-training techniques.

Ref \cite{jeon2021smartcondetect} developed a method to detect code vulnerabilities in smart contracts using a pre-trained BERT model. The method first extracts key code snippets from smart contracts written in Solidity using static analysis tools. It then applies the pre-trained BERT model, which has been pre-trained on a large amount of text data, to analyze these snippets in depth to identify potential security vulnerabilities. The core advantage of this technique lies in BERT's self-attention mechanism, which can effectively capture complex dependencies and patterns in the code, thereby improving the accuracy and efficiency of vulnerability detection.

The system was trained and tested using 10,000 smart contract samples from Ethereum. The results showed a precision of 98.7\%, recall of 86.2\%, and an F1 score of 90.9\%. These performance metrics significantly outperformed other models such as Support Vector Machine (SVM), Eth2Vec, and Graph Neural Networks (DR-GCN). Specifically, SVM had an F1 score of 45.2\%, Eth2Vec had 57.5\%, and DR-GCN had 78.1\%. This advantage can be attributed to BERT's deep semantic analysis capabilities and fast parallel processing characteristics, enabling SmartConDetect to effectively identify and handle complex smart contract code structures and subtle vulnerabilities.Furthermore, through pre-training on a large-scale dataset, SmartConDetect demonstrated strong generalization capabilities, enabling it to adapt to varying programming styles and evolving security threats, significantly improving the efficiency and accuracy of security detection in smart contracts.

Ref \cite{zhang2022smart} developed a method that combines a pre-trained BERT model with a Multi-Objective Detection Neural Network (MODNN). MODNN utilizes the pre-trained Bert model to convert Critical Operation Sequences (COS) into explicit features and combines opcode to construct a co-occurrence matrix to learn implicit features. This process not only enhances the model's ability to detect known vulnerability types but also improves its ability to learn unknown vulnerability types. Next, the MODNN part uses high-dimensional features extracted from BERT and sets multiple detection targets to achieve simultaneous recognition and classification of various types of vulnerabilities. The advantage of this combined approach is that the strong semantic processing capabilities of BERT are combined with the efficient classification functionality of MODNN, significantly improving the accuracy and speed of vulnerability detection. The system can also automatically adjust and expand based on newly emerging vulnerabilities, maintaining sensitivity and responsiveness to the latest security threats.

Tested on over 18,000 Ethereum smart contracts, MODNN demonstrated a high F1 score (average 94.8\%), significantly outperforming several standard machine learning classification models, demonstrating its effectiveness and efficiency in smart contract vulnerability detection. This technology provides an efficient, dynamically adaptable detection mechanism for smart contract security, greatly enhancing the security of blockchain applications.

Ref \cite{zhang2022novel} developed a smart contract vulnerability detection system called SCVDIE, which combines information graph technology and ensemble learning strategies to improve detection accuracy and efficiency. The SCVDIE system adopts seven different neural network models, including ones that apply the Transformer model's self-attention mechanism. This enables the system to effectively identify and learn long-distance data dependencies and complex interaction patterns in smart contract code. 
By constructing the data's association and interaction structures through information graphs, each model learns data features from different perspectives. These models are then integrated using ensemble learning methods to combine the strengths of each model, improving overall detection performance. The inclusion of the Transformer model not only speeds up the feature extraction process but also enhances the model's ability to recognize complex patterns. This strategy of using multiple learning models significantly improves the system's adaptability to new and unknown vulnerabilities, providing an effective technical path for the security analysis of smart contracts.

Experiments show that SCVDIE, tested on 21,667 smart contract samples, achieved an average accuracy of 95.46\%, precision of 96.81\%, and recall of 97.26\%. This is significantly better than traditional static analysis tools such as Oyente\cite{luu2016making}, Mythril \cite{mueller2017mythril}, and Securify \cite{tsankov2018securify} (with F1 scores of 46.1\%, 45.6\%, and 43.9\%, respectively). Furthermore, compared to other single neural network models such as CNN and RNN, the ensemble learning method of the SCVDIE system provides higher prediction accuracy and more stable performance, reflecting the advantage of ensemble strategies in integrating features learned by multiple models.
Moreover, by integrating the learning outcomes of multiple models, SCVDIE not only improves the granularity and accuracy of feature recognition but also enhances the system's adaptability and robustness to new vulnerability patterns, effectively supporting the security analysis of smart contracts in complex and large-scale data environments.

Ref \cite{xu2023vulnerability} developed a hybrid neural network model based on deep learning, which integrates SolBERT, Bidirectional Gated Recurrent Units (BiGRU), and hierarchical attention mechanisms to effectively improve the accuracy and efficiency of smart contract vulnerability detection. The framework uses SolBERT for pre-training to dynamically capture semantic relationships and key features in the source code of smart contracts. It then utilizes BiGRU to process serialized textual data, enhancing the model's ability to capture long-distance dependencies in the text. The hierarchical attention mechanism further refines the model's judgment of the importance of various parts of the text, particularly when identifying the context of potential vulnerabilities in the code, allowing it to more accurately highlight key information. This model, which combines the SolBERT architecture with advanced recurrent neural network techniques, not only significantly improves the accuracy of smart contract vulnerability detection but also optimizes the model's speed and efficiency in handling large-scale datasets through self-attention mechanisms, demonstrating outstanding performance in complex smart contract security analysis tasks.

The experiments were conducted using a large number of smart contract samples for testing. The results showed that the model achieved an accuracy of 93.85\% and a micro-average F1 score of 94.02\%, both of which are higher than most existing vulnerability detection methods. Additionally, the average detection time for a single smart contract was approximately 4.5 seconds, much faster than traditional static analysis tools and other deep learning-based models. These experimental data not only validate the superiority of the SolBERT-BiGRU-attention model in smart contract vulnerability detection but also demonstrate its efficiency and practicality in real-world applications, making it an effective tool for handling complex and large-scale smart contract data.

Ref \cite{tang2023deep} utilized CodeBERT, a pre-trained model based on the Transformer architecture, to enhance the detection capability of vulnerabilities in smart contracts. This approach involved pre-training CodeBERT on a large dataset of programming languages, enabling it to understand the deep semantics and complex structures in smart contract code, thus effectively identifying potential vulnerabilities. During the vulnerability detection process, CodeBERT analyzed smart contract code, captured key features and patterns, and then passed this information to subsequent classifiers to accurately identify various types of security vulnerabilities. The model's self-attention mechanism allowed it to consider the global dependency relationships in the code, enhancing its ability to detect complex vulnerabilities. Additionally, CodeBERT's parallel processing capability significantly improved the speed of vulnerability detection, making the method not only highly accurate but also suitable for rapidly detecting large numbers of smart contracts in practical applications, providing strong support for the security of smart contracts.

Experiments were conducted using a large dataset containing thousands of smart contracts with varying complexity and security vulnerabilities. The results showed that the CodeBERT model achieved a 91\% accuracy rate, an 89\% recall rate, and an F1 score of 90\% in smart contract vulnerability detection. Compared to traditional vulnerability detection methods and other deep learning-based models, CodeBERT demonstrated higher performance, such as an approximately 10\% improvement in the highest F1 score compared to using traditional machine learning methods. These experimental results not only validated the efficiency and accuracy of CodeBERT in smart contract vulnerability detection but also demonstrated its robust adaptability to complex and diverse smart contract environments.

Ref \cite{jie2023novel} developed an innovative multimodal AI framework that combines word2vec, BERT, and Graph Convolutional Networks (GCN) to enhance the vulnerability detection capability in smart contract code. This framework integrates multiple data sources, including source code text, compiled bytecode, and execution trace data from the Ethereum Virtual Machine, to utilize word2vec for generating word embeddings to capture basic semantic features of the code. At the same time, the BERT model deeply analyzes long-distance dependencies and complex structures in the code through its self-attention mechanism to identify potential security vulnerabilities. Finally, GCN is used to process the graph structure generated from the code to further explore the complex interactions between nodes.

Experimental results show that the model's performance reached a high of 99.71\% in handling tasks extracted from 101,082 functions from the SmartEmbed dataset. This high performance is mainly attributed to the strong semantic understanding capability of the BERT model in natural language processing tasks, as well as the effectiveness of GCN in handling graph-structured data. Additionally, the application of multimodal learning enables the framework to extract and merge features from multiple data sources, thus more comprehensively capturing potential vulnerabilities in smart contracts.

Ref \cite{le2023contextual} utilized the DistilBERT model, a Transformer-based pretrained language model, combined with Long Short-Term Memory (LSTM) and Multi-Layer Perceptron (MLP) to propose an efficient method for detecting vulnerabilities in smart contracts. The method first processes smart contract code using a custom tokenizer to maintain the structural integrity of the code, and then employs DistilBERT for in-depth semantic analysis and feature extraction. During the fine-tuning phase, the model is optimized for specific reentrancy vulnerabilities, using a combination of LSTM and MLP to improve the accuracy and efficiency of vulnerability detection. 

The advantage of this method lies in its ability to accurately capture complex dependencies and deep semantics in smart contract code. Through the parallel processing capability of the Transformer architecture, it significantly improves processing speed and model performance, providing strong technical support for smart contract security. Experimental data shows that the framework achieved a high of 99.71\% in handling 101,082 smart contract samples, significantly outperforming other single models. Compared to traditional deep learning methods, DistilBERT demonstrates higher accuracy and efficiency in vulnerability detection tasks, especially in handling complex smart contract code. Furthermore, the model's use of DistilBERT for pretraining and fine-tuning effectively enhances its understanding and predictive capabilities for potential security threats in smart contract code. These experimental data not only confirm the powerful performance of the DistilBERT model but also highlight its practicality and reliability in the field of smart contract vulnerability detection.

Ref \cite{nguyen2023mando} proposed a smart contract vulnerability detection method based on Heterogeneous Graph Transformers (HGT). This method converts the source code of smart contracts into Heterogeneous Contract Graphs (HCGs), which include various types of nodes and edges, such as functions, variables, control flows, and data flows, to map the internal structure of smart contracts in detail. By using the self-attention mechanism of HGT, the model can accurately identify and analyze long-distance dependency relationships and complex patterns in the graph, which is crucial for detecting potential vulnerabilities in smart contracts. The Transformer model's ability to handle large amounts of concurrent data streams enables it to process data faster and more efficiently. Additionally, the multi-head attention mechanism of the HGT model further enhances its ability to handle different types of relationships, enabling the system to more comprehensively and accurately identify vulnerabilities in smart contract code. This approach not only improves the accuracy and speed of vulnerability detection but also enhances the system's robustness and adaptability to new vulnerability patterns, providing an advanced technical solution for smart contract security analysis.

Experimental data show that in testing over 55,000 Ethereum smart contracts, MANDO-HGT performed well at different detection granularities, with F1 scores increasing from 0.7\% to 76\%, demonstrating its wide applicability in fine-grained and coarse-grained analysis. Compared to traditional methods, this new framework can more comprehensively analyze potential vulnerabilities in smart contracts, improving detection accuracy and efficiency.

Ref \cite{gong2023scgformer} proposed an innovative method to detect security vulnerabilities in smart contracts by combining control flow graphs (CFGs) and Transformer-based models. This method extracts control flow information from the bytecode of smart contracts to construct control flow graphs, which clearly show the execution paths of various operations in the contract. Then, a Transformer model is applied to these graphs for deep learning analysis, utilizing its self-attention mechanism to identify and analyze complex dependencies and potential risks in the contract execution process. This approach, which combines graph representation and deep natural language processing techniques, not only improves the accuracy of vulnerability detection but also significantly enhances the analysis speed due to the efficient parallel processing capability of Transformers. Furthermore, by continuously learning and adapting to newly emerging vulnerability patterns, the system can continuously improve its ability to detect new types of vulnerabilities, providing an effective and forward-looking technical solution for ensuring the security of smart contracts.

The experiments were conducted using over 50,000 smart contract samples, and the results show that SCGformer achieves an accuracy rate of 94.36\% and an F1 score of 93.58\% in vulnerability detection, significantly outperforming existing detection tools such as Slither \cite{feist2019slither} and Oyente \cite{luu2016making}.These experimental results not only validate the technical advancement of SCGformer in smart contract vulnerability detection but also demonstrate its efficiency and reliability in practical deployment, providing an effective technical solution for future smart contract security.

Ref \cite{jain2024integrated} developed a model that combines Transformer, Bidirectional Gated Recurrent Unit (Bi-GRU), and Text Convolutional Neural Network (Text-CNN) for smart contract vulnerability detection. This multi-model approach first uses the self-attention mechanism of the Transformer model to accurately analyze the complex dependencies and context information in smart contract code. Then, Bi-GRU enhances the capture of time-series data through its bidirectional processing capability, and finally, Text-CNN extracts local features from specific text regions. This progressive feature processing strategy not only improves the identification accuracy of potential vulnerabilities in smart contracts but also significantly increases processing speed and efficiency through the parallel processing capabilities of each model. The method enhances the accuracy and efficiency of smart contract vulnerability detection through fine-grained, multi-stage feature analysis, providing a new technical solution for blockchain security. 

In experiments, the model was tested on 49,552 real-world smart contract samples, showing superior performance in terms of accuracy, recall, and F1 score compared to traditional static analysis and other deep learning methods. For example, compared to traditional tools, the model achieved an accuracy of 96.1\%, a recall of 97.8\%, and an F1 score of 96.5\%, demonstrating its outstanding performance and practicality in smart contract vulnerability detection. These experimental results fully demonstrate the effectiveness of the integrated approach using Transformer, Bi-GRU, and Text-CNN models in smart contract security analysis.

Ref \cite{balci2023accelerating} introduced VASCOT, a tool designed specifically for detecting vulnerabilities in smart contracts by analyzing the bytecode of the Ethereum Virtual Machine (EVM). The tool uses a Transformer-based model to deeply analyze the bytecode, effectively identifying complex interactions and potential risks in the contracts. The core advantage of VASCOT is its ability to analyze compiled bytecode directly without the need for source code, significantly improving its applicability and flexibility. The Transformer technology enables VASCOT to process large amounts of data in parallel, greatly improving analysis speed. Additionally, the self-attention mechanism helps the model capture long-distance dependencies, enhancing the accuracy and detail of vulnerability detection. This method not only optimizes the security audit process of smart contracts, reducing dependence on human resources, but also provides strong technical support for future smart contract development and maintenance, demonstrating the potential of deep learning technology in improving blockchain security.

In experiments, VASCOT achieved a testing accuracy of 90\% on a set of 16,363 verified Ethereum smart contracts. Compared to using Long Short-Term Memory (LSTM) \cite{tann1811towards} networks, the method showed a 29\% improvement in accuracy, with training time only 6\% of that of LSTM. Therefore, the VASCOT system not only outperforms traditional methods in terms of detection speed and accuracy but also reduces the operational difficulty of large-scale smart contract audits through its efficient technical implementation, making it a powerful tool for smart contract security analysis.

Ref \cite{jiang2022vddl} introduced an advanced smart contract vulnerability detection system named VDDL. It adopts a multi-layer bidirectional Transformer architecture to analyze smart contract code. This architecture captures information from both directions of the code sequence using the self-attention mechanism of Transformers at each layer, effectively understanding complex dependencies and potential vulnerabilities in the code. VDDL first preprocesses smart contract code standardly and then processes it through multiple layers of bidirectional Transformers. Each layer conducts in-depth analysis of the input data, identifying high-risk patterns and potential security vulnerabilities. This bidirectional and multi-layer processing method not only enhances the model's ability to understand the deep semantics of smart contract code but also significantly improves the accuracy and efficiency of vulnerability detection. Additionally, the high parallelism of Transformers enables VDDL to quickly process large amounts of smart contract data, making it suitable for deployment in real-time systems to proactively prevent and repair security vulnerabilities in code. Through this innovative technical approach, VDDL provides a powerful tool for the security verification of smart contracts, significantly enhancing the security and reliability of blockchain applications.

Through specific experimental data comparisons, the VDDL model performs excellently on the smart contract dataset, with an accuracy of 92.35\%, a recall rate of 81.43\%, and an F1 score of 86.38\%. The study also compares the VDDL model with traditional machine learning methods, including logistic regression, naive Bayes, and random forest algorithms. The results show that the VDDL model outperforms these traditional methods in terms of accuracy, recall rate, and F1 score. Compared with the random forest algorithm, the VDDL model's accuracy increased by 1.13\%, the recall rate increased by 4.83\%, and the F1 score increased by 2.52\%. Furthermore, the VDDL model is also compared with the best work of TextRNN \cite{Hu2019Chinese} classification problems based on artificial intelligence. On the same dataset, the accuracy, recall rate, and F1 score of VDDL are 1.63\%, 1.97\%, and 2.74\% higher than TextRNN, respectively, further demonstrating the effectiveness and high performance of the VDDL model in detecting smart contract security vulnerabilities. These experimental data not only demonstrate the superior performance of the VDDL model in smart contract vulnerability detection but also highlight its efficiency and reliability in handling real-world applications.

Ref \cite{gong2023gratdet} developed GRATDet, a smart contract vulnerability detection tool that combines graph representation and Transformer technology. The system first converts smart contract code into a graph representation, where functions, variables, and their interactions are abstracted as nodes and edges in the graph, allowing for a clear visualization of the contract's logic and data flow. Then, by applying a Transformer-based deep learning model, GRATDet uses self-attention mechanisms to analyze these graphs, digging deep into the dependencies between nodes and potential security risks.
The parallel processing capability of the Transformer model and its sensitivity to long-distance dependencies enable GRATDet to quickly and accurately identify various complex vulnerability patterns, including subtle vulnerabilities that traditional methods struggle to capture. Furthermore, this combined approach of graph and Transformer not only enhances the efficiency and accuracy of vulnerability detection but also adapts to changes and updates in smart contract code, providing a powerful and flexible solution for smart contract security. This makes GRATDet particularly forward-thinking and practical in the field of smart contract security, greatly advancing research and practice in blockchain technology security.

By combining graph representation techniques and the Transformer deep learning framework, GRATDet achieved an accuracy of 95.22\%, precision of 95.59\%, recall of 95.17\%, and an F1 score of 95.16\% on the test set. These results are significantly better than other commonly used tools such as Mythril \cite{mueller2017mythril} (F1 score of 50.44\%), SmartCheck \cite{tikhomirov2018smartcheck}, and Slither \cite{feist2019slither}. This notable performance improvement is attributed to GRATDet's ability to effectively analyze and understand the complex relationships and deep semantics in smart contract code, ensuring high efficiency and accuracy in the detection process, making it a powerful tool in the field of smart contract security analysis.

Ref \cite{he2024enhancing} proposed a smart contract vulnerability detection model that combines BERT, Attention Mechanism (ATT), and Bidirectional Long Short-Term Memory Network (BiLSTM). The technical architecture of the model is designed as a three-layer structure: first, the BERT model is used to embed smart contract code, leveraging its pre-trained deep language understanding capability to extract semantic features of the code; second, the attention mechanism enhances the model's focus on key information, allowing the model to pay more attention to parts of the sequence data that are crucial for vulnerability detection; finally, BiLSTM is applied to the attention-weighted features for time-series analysis, using its bidirectional processing advantage to capture dependencies in the context.This combined use of Transformer model (BERT) and RNN (BiLSTM) exploits the strengths of both, with Transformer's efficient parallel processing and global context capture capabilities, along with BiLSTM's sequential data processing capabilities. This structural design enables the model to not only deeply understand the complex semantics of smart contract code but also analyze the time dependencies in code execution in detail, significantly improving the accuracy and efficiency of vulnerability detection. Additionally, the introduction of attention mechanism further optimizes the flow of information in the network, ensuring the sensitivity and precision of the model in identifying potential vulnerabilities. This application of technology provides a powerful guarantee for the security of smart contracts, representing an innovative and practical solution.

The experimental comparison shows that the BERT-ATT-BiLSTM model significantly outperforms traditional methods like AWD-LSTM \cite{merity2017regularizing} and MulCas \cite{ma2013multilevel} main smart contract vulnerability detection. Specifically, BERT-ATT-BiLSTM achieves an accuracy and recall of 98.58\%, with an F1 score of 98.26\%, indicating its extremely high detection accuracy and consistency. In comparison, the best accuracy of AWD-LSTM is 91.30\%, with an F1 score of 90.00\%, while MulCas has an accuracy of 95.10\% but an F1 score of only 78.90\%. This significant difference is due to the BERT model's ability to deeply understand the code, the efficiency of BiLSTM in capturing contextual information, and the high precision of the attention mechanism in highlighting key features.

Ref \cite{guo2024smart} proposed an innovative framework for detecting vulnerabilities in smart contracts, which combines Transformer models with multi-scale convolutional neural network (CNN) encoders. This integrated approach first processes smart contract code using a Transformer, leveraging its self-attention mechanism to comprehensively understand long-distance dependencies and complex interactions in the code, thus accurately capturing the global semantics and structural features of the code. Subsequently, the Transformer's output is further analyzed by multi-scale CNN encoders, which extract features at different resolution levels, effectively identifying various vulnerability patterns in the code from micro to macro levels. This technical architecture enables the model to grasp not only the overall logic of the code but also the potential detailed vulnerabilities, significantly improving the coverage and accuracy of vulnerability detection. Additionally, the research team introduces Surface Feature Encoders (SFE) and Deep Residual Shrinking Networks (DRSN) to further optimize the feature extraction process, reduce information redundancy, and enhance the model's ability to identify different types of vulnerabilities.The Transformer model, with its unique self-attention mechanism, significantly enhances the model's understanding of global information in smart contract code, thereby improving the accuracy and efficiency of vulnerability detection, especially when dealing with large contracts containing complex interactions and multi-level function calls. This technical approach not only enhances the comprehensiveness and accuracy of vulnerability detection but also greatly improves processing speed, providing strong technical support for smart contract security.

In experiments, for reentrancy attack vulnerability detection, the MEVD model achieved an accuracy of 92.13\%, an F1 score of 91.37\%, surpassing ReChecker \cite{qian2020towards} by 17.48\% in accuracy and outperforming ReChecker's F1 score of 69.41\%, indicating that MEVD is more effective in balancing the precision and recall of vulnerability detection. For timestamp dependency vulnerability detection, the MEVD model achieved an accuracy of 90.85\%, surpassing the graph neural network model CGE \cite{liu2021combining} by 3.09\%, with an F1 score of 89.80\% compared to CGE's 85.43\%, demonstrating that MEVD identifies these vulnerabilities more accurately and comprehensively. For infinite loop vulnerability detection, MEVD's accuracy was 86.94\%, slightly higher than CGE's 85.89\%, but MEVD exhibited more stable performance. MEVD achieved a significantly improved F1 score of 86.01\%, compared to CGE's 82.10\%, highlighting MEVD's advantage in handling complex vulnerabilities.

Ref \cite{sun2023assbert} presented a novel framework for detecting vulnerabilities in smart contracts called ASSBert, which combines active learning and semi-supervised learning methods, using the BERT model as its core for deep analysis of smart contract code. ASSBert first utilizes BERT's pre-trained model to extract deep semantic features of smart contract code, effectively capturing the context and potential complex structures within the code. Through active learning mechanism, the system can identify and select samples that would most benefit model improvement for annotation, thereby optimizing the training process and reducing the need for manual annotation. Meanwhile, the semi-supervised learning strategy allows the model to utilize a large amount of unlabeled data for training, further expanding the training set and improving the model's generalization ability. This combined strategy of active learning and semi-supervised learning not only enhances the efficiency of model training but also significantly improves the accuracy and coverage of vulnerability detection. BERT's self-attention mechanism enables the model to capture complex relationships and subtle patterns in the code, while the combination of active and semi-supervised learning greatly increases the data utilization rate and the model's generalization ability, making ASSBert an efficient and cost-effective tool for smart contract security detection. This application not only improves the accuracy of vulnerability detection but also provides strong technical support for the security protection of smart contracts. 

In experiments, when the labeled sample ratio was 20\%, ASSBert achieved an accuracy of 78.6\% in detecting timestamp dependency vulnerabilities, outperforming other methods such as Bert, Bert-AL, and Bert-SSL, which had accuracies of 31.1\%, 79.1\%, and 54.5\%, respectively. These experimental results not only demonstrate ASSBert's ability to efficiently detect vulnerabilities using limited labeled data but also highlight its unique advantages in combining semi-supervised learning and active learning. This approach effectively utilizes unlabeled data, improving the precision and recall of vulnerability detection by continuously optimizing the model, especially in cases where data annotation resources are limited.

Ref \cite{gu2023trap} proposed a smart contract vulnerability detection system called TrapFormer, which utilizes a Transformer-based deep learning model to analyze the bytecode of smart contracts. TrapFormer extracts opcode sequences directly from the bytecode of smart contracts, and uses these opcodes as input to capture the dependencies between different opcodes using the self-attention mechanism of the Transformer model, thereby identifying potential malicious behavior or traps. The core advantage of this method is its ability to understand the execution logic and potential risks of contracts in depth without relying on the contract source code. TrapFormer's model includes multiple layers of self-attention networks, which are stacked to enhance the model's ability to understand complex contract behaviors, enabling it to effectively identify and differentiate between normal behavior and potential traps in large amounts of contract data. Additionally, due to the parallel processing capability of the Transformer model, TrapFormer demonstrates efficient computation speed and excellent scalability when handling large-scale data. Overall, TrapFormer combines advanced self-attention technology and deep learning to not only improve the automation efficiency of smart contract detection but also significantly enhance the reliability and accuracy of the detection system in practical applications, providing a strong technical guarantee for the security of smart contracts.

In experiments, for regular contracts, TrapFormer achieves a precision of 0.973, recall of 0.989, and an F1 score of 0.981. Compared to other models, the Transformer model achieves an F1 score of 0.971, and the SCSGuard model \cite{hu2022scsguard} achieves 0.968, demonstrating TrapFormer's higher overall performance in handling regular contracts. For Ponzi contracts, TrapFormer also performs well, with a precision of 0.975, recall of 0.954, and an F1 score of 0.965. In comparison, the Transformer model achieves an F1 score of 0.940, and the PSD-OL model achieves 0.875, further proving TrapFormer's advantages. For honey pot contracts, TrapFormer stands out with a precision of 0.982, recall of 0.986, and an F1 score of 0.984. In contrast, the Transformer model achieves an F1 score of 0.969, and SCSGuard achieves 0.924, demonstrating TrapFormer's significant advantage in honey pot contract detection. These results indicate that the TrapFormer model, with its densely connected Transformer structure, effectively improves precision, recall, and F1 scores in various smart contract detection tasks. Particularly in detecting Ponzi and honey pot contracts, TrapFormer provides higher performance due to its ability to effectively capture complex dependencies and patterns between opcodes. Additionally, its stability under different data imbalance conditions further proves its effectiveness as a tool for smart contract security protection.
\begin{table*}[h]
\caption{COMPARATIVE STUDY OF EXISTING TECHNIQUES FOR CRYPTOCURRENCY PRICE PREDICTION.}
\label{tab:my-table}
\resizebox{\textwidth}{!}{
\begin{tabular}{clllcllcllclllcllllllllllllllll}
\hline
\multicolumn{4}{c}{\textbf{Ref.}} & \multicolumn{3}{c}{\textbf{Year}} & \multicolumn{3}{c}{\textbf{Type}}             & \multicolumn{4}{c}{\textbf{Detection Object}} & \multicolumn{17}{c}{\textbf{Major Characteristic}}                                                                                                                                      \\
\multicolumn{4}{c}{{[}106{]}}     & \multicolumn{3}{c}{2021}          & \multicolumn{3}{c}{Graph Representation}      & \multicolumn{4}{c}{Source Code}               & \multicolumn{17}{c}{\begin{tabular}[c]{@{}c@{}}Using a Crucial Data Flow Graph to capture key \\ information related to vulnerabilities.\end{tabular}}                                  \\
\multicolumn{4}{c}{{[}107{]}}     & \multicolumn{3}{c}{2021}          & \multicolumn{3}{c}{Text Representation}       & \multicolumn{4}{c}{Source Code}               & \multicolumn{17}{c}{Using a pre-trained BERT model for text representation.}                                                                                                            \\
\multicolumn{4}{c}{{[}99{]}}      & \multicolumn{3}{c}{2022}          & \multicolumn{3}{c}{Text Representation}       & \multicolumn{4}{c}{Bytecode}                  & \multicolumn{17}{c}{\begin{tabular}[c]{@{}c@{}}Using a BERT model and a multi-objective \\ detection neural network for text representation.\end{tabular}}                              \\
\multicolumn{4}{c}{{[}108{]}}     & \multicolumn{3}{c}{2022}          & \multicolumn{3}{c}{Graph Representation}      & \multicolumn{4}{c}{Bytecode}                  & \multicolumn{17}{c}{\begin{tabular}[c]{@{}c@{}}Utilizing information graphs and ensemble learning \\ methods to analyze smart contract bytecode.\end{tabular}}                          \\
\multicolumn{4}{c}{{[}111{]}}     & \multicolumn{3}{c}{2023}          & \multicolumn{3}{c}{Text Representation}       & \multicolumn{4}{c}{Source Code}               & \multicolumn{17}{c}{\begin{tabular}[c]{@{}c@{}}Using the SolBERT model and bidirectional GRU with \\ an attention mechanism for text representation.\end{tabular}}                      \\
\multicolumn{4}{c}{{[}112{]}}     & \multicolumn{3}{c}{2023}          & \multicolumn{3}{c}{Text Representation}       & \multicolumn{4}{c}{Source Code}               & \multicolumn{17}{c}{\begin{tabular}[c]{@{}c@{}}Utilizing the CodeBERT model combined \\ with LSTM and CNN for text representation.\end{tabular}}                                        \\
\multicolumn{4}{c}{{[}113{]}}     & \multicolumn{3}{c}{2023}          & \multicolumn{3}{c}{Multimodal Representation} & \multicolumn{4}{c}{Source Code}               & \multicolumn{17}{c}{Combining Word2Vec, BERT, and GCN for multimodal representation.}                                                                                                   \\
\multicolumn{4}{c}{{[}114{]}}     & \multicolumn{3}{c}{2023}          & \multicolumn{3}{c}{Text Representation}       & \multicolumn{4}{c}{Source Code}               & \multicolumn{17}{c}{\begin{tabular}[c]{@{}c@{}}Using a custom tokenizer and DistilBERT model combined \\ with MLP and LSTM for text representation.\end{tabular}}                       \\
\multicolumn{4}{c}{{[}115{]}}     & \multicolumn{3}{c}{2023}          & \multicolumn{3}{c}{Graph Representation}      & \multicolumn{4}{c}{Bytecode}                  & \multicolumn{17}{c}{\begin{tabular}[c]{@{}c@{}}Using custom-built heterogeneous contract graphs combined \\ with heterogeneous graph transformers (HGT).\end{tabular}}                  \\
\multicolumn{4}{c}{{[}116{]}}     & \multicolumn{3}{c}{2023}          & \multicolumn{3}{c}{Graph Representation}      & \multicolumn{4}{c}{Bytecode}                  & \multicolumn{17}{c}{\begin{tabular}[c]{@{}c@{}}Using control flow graphs combined with a Transformer model \\ for vulnerability detection in smart contract bytecode.\end{tabular}}     \\
\multicolumn{4}{c}{{[}118{]}}     & \multicolumn{3}{c}{2023}          & \multicolumn{3}{c}{Text Representation}       & \multicolumn{4}{c}{Bytecode}                  & \multicolumn{17}{c}{Using Transformer, Bi-GRU, and Text-CNN for text representation.}                                                                                                   \\
\multicolumn{4}{c}{{[}119{]}}     & \multicolumn{3}{c}{2023}          & \multicolumn{3}{c}{Text Representation}       & \multicolumn{4}{c}{Bytecode}                  & \multicolumn{17}{c}{\begin{tabular}[c]{@{}c@{}}Using a Transformer model for text representation \\ and sequential analysis of smart contract bytecode.\end{tabular}}                   \\
\multicolumn{4}{c}{{[}121{]}}     & \multicolumn{3}{c}{2023}          & \multicolumn{3}{c}{Text Representation}       & \multicolumn{4}{c}{Source Code}               & \multicolumn{17}{c}{\begin{tabular}[c]{@{}c@{}}Using a multi-layer bidirectional Transformer \\ architecture and the CodeBERT model.\end{tabular}}                                      \\
\multicolumn{4}{c}{{[}123{]}}     & \multicolumn{3}{c}{2023}          & \multicolumn{3}{c}{Graph Representation}      & \multicolumn{4}{c}{Source Code}               & \multicolumn{17}{c}{\begin{tabular}[c]{@{}c@{}}Using graph representation combined with a Transformer model \\ for vulnerability detection in smart contract source code.\end{tabular}} \\
\multicolumn{4}{c}{{[}130{]}}     & \multicolumn{3}{c}{2023}          & \multicolumn{3}{c}{Text Representation}       & \multicolumn{4}{c}{Bytecode}                  & \multicolumn{17}{c}{\begin{tabular}[c]{@{}c@{}}Using an improved Transformer model for text representation \\ and trap contract detection in smart contract opcodes.\end{tabular}}      \\
\multicolumn{4}{c}{{[}124{]}}     & \multicolumn{3}{c}{2024}          & \multicolumn{3}{c}{Text Representation}       & \multicolumn{4}{c}{Bytecode}                  & \multicolumn{17}{c}{\begin{tabular}[c]{@{}c@{}}Using BERT, attention mechanism, and \\ BiLSTM for text representation.\end{tabular}}                                                    \\
\multicolumn{4}{c}{{[}127{]}}     & \multicolumn{3}{c}{2024}          & \multicolumn{3}{c}{Text Representation}       & \multicolumn{4}{c}{Source Code}               & \multicolumn{17}{c}{\begin{tabular}[c]{@{}c@{}}Using multi-scale encoders combined \\ with Transformer and CNN models.\end{tabular}}                                                    \\
\multicolumn{4}{c}{{[}129{]}}     & \multicolumn{3}{c}{2024}          & \multicolumn{3}{c}{Text Representation}       & \multicolumn{4}{c}{Source Code}               & \multicolumn{17}{c}{\begin{tabular}[c]{@{}c@{}}Combining active learning and semi-supervised learning \\ methods, using a BERT model for text representation.\end{tabular}}            
\end{tabular}
}
\end{table*}
\subsection{Cryptocurrency Prediction and Trend Analysis}\

\subsubsection{Background and Objectives}\

Cryptocurrency prediction and trend analysis refers to the use of various methods and techniques to analyze the behavior of the cryptocurrency market in order to predict future price movements and market dynamics. .As shown in Figure 15,the research literature on smart contract Cryptocurrency Prediction has steadily increased in recent years.This includes the comprehensive analysis of historical data, market supply and demand relationships, global economic events, and investor psychology and behavior. The main techniques include technical analysis (analyzing price behavior through charts and mathematical models), fundamental analysis (assessing macroeconomic and industry factors impacting the market), and sentiment analysis (examining the influence of social media and news on market sentiment).
\begin{figure}[h]
    \centering
    \includegraphics[width=0.5\linewidth,trim=5cm 9.5cm 5cm 9.5cm]{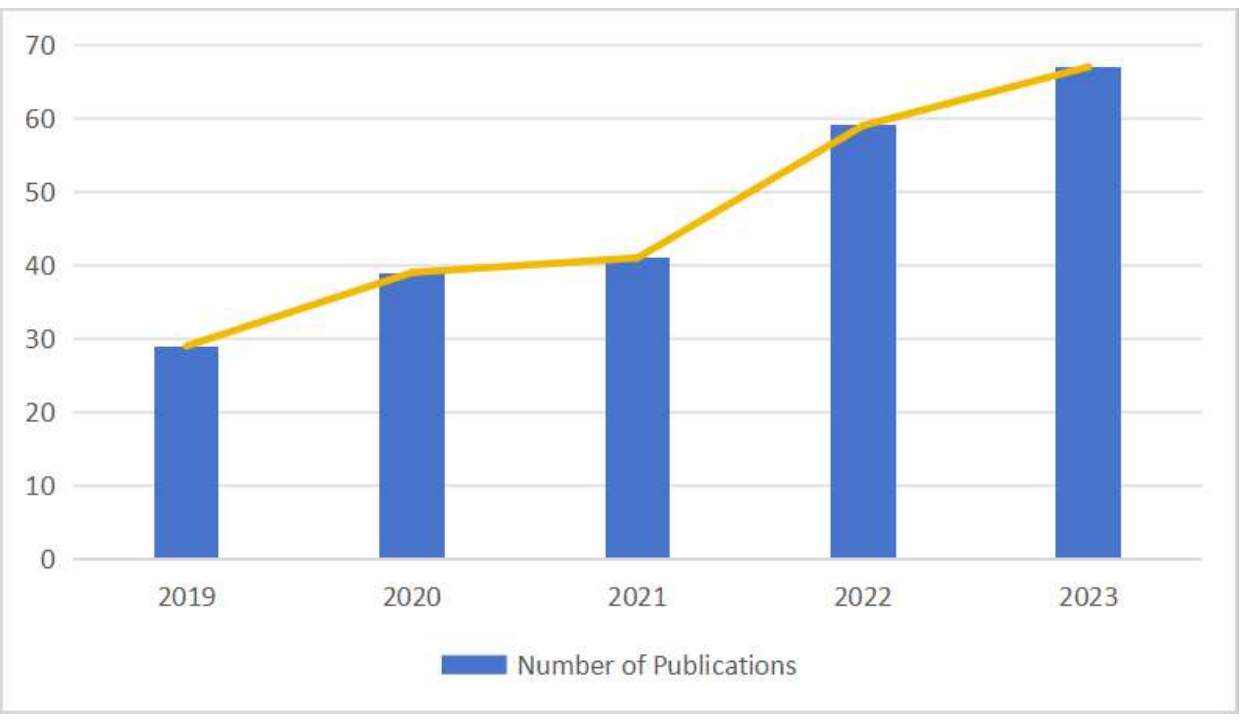}
    \caption{Statistics on the publication of papers on Cryptocurrency Prediction.}
    \label{fig:15}
\end{figure}
The extreme volatility of cryptocurrency prices is a major challenge for investors and market analysts. Rapid price changes can be triggered by various factors, including market sentiment, policy changes, and global economic events.Ref \cite{bariviera2017inefficiency} have shown through their research that understanding this volatility is crucial for developing effective investment and risk management strategies. Furthermore, the rapid development of cryptocurrency and blockchain technology has introduced new trading mechanisms and financial instruments, such as smart contracts and DeFi. These technological advancements have profound impacts on market prices and participant behavior. Ref \cite{cong2019blockchain} have explored in detail how blockchain technology drives financial innovation and its potential market impacts.

Conducting cryptocurrency prediction and trend analysis can help investors and market analysts better understand and forecast market dynamics, thereby optimizing investment decisions and risk management. Additionally, the regulatory stance on cryptocurrencies varies significantly across different countries and regions, directly influencing market prices and trading behavior.Ref \cite{armour2016principles} discuss how regulatory frameworks shape the operation of cryptocurrency markets and their impact on market stability.
\begin{figure}[h]
    \centering
    \includegraphics[width=0.5\linewidth,trim=5cm 9.5cm 5cm 9.5cm]{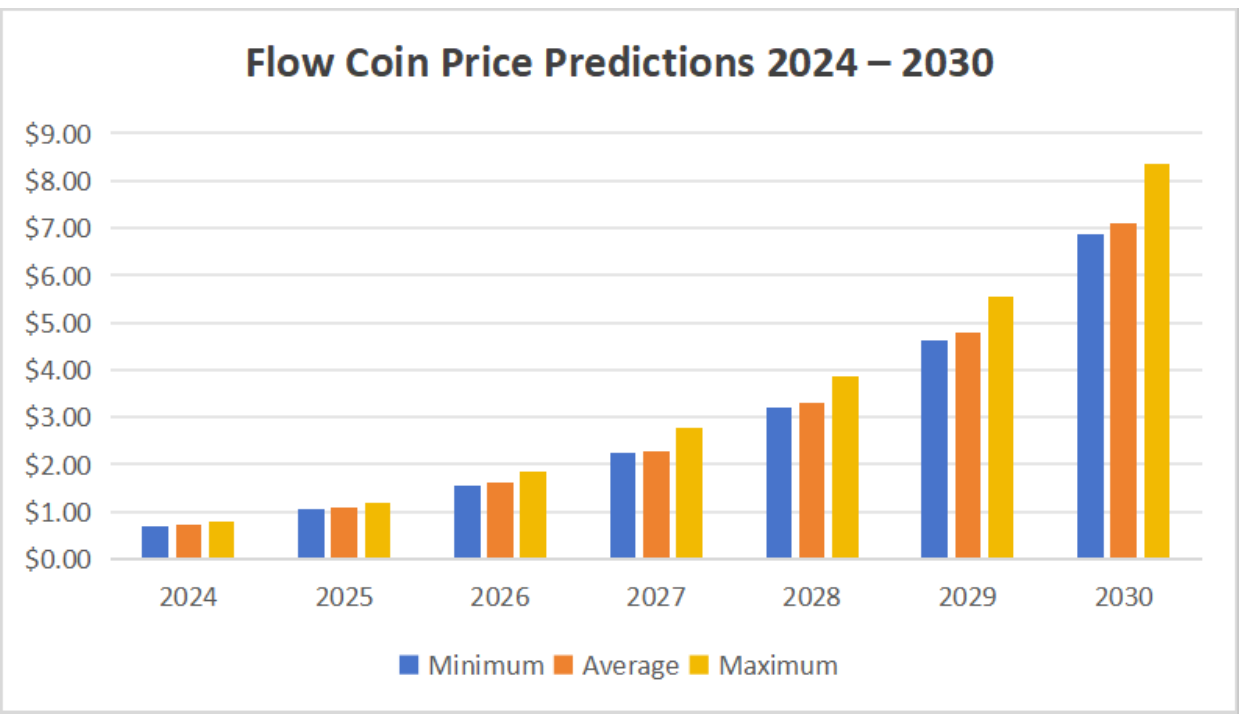}
    \caption{Flow coin price predictions 2024 – 2030.}
    \label{fig:16}
\end{figure}
The cryptocurrency market attracts various types of participants, including retail investors, institutional investors, miners, and developers.For example, Figure 16 is a prediction of the Flow coin in cryptocurrency from 2024 to 2030, used to analyze whether this cryptocurrency is suitable for investment.Understanding the motivations and behaviors of these groups is crucial for predicting market trends.Ref \cite{catalini2020some} analyze how different market participants influence technology adoption and market dynamics in their research. In summary, these analyses can promote the healthy development of the market, enhancing its transparency and efficiency.

\subsubsection{Applications and Limitations of Traditional Methods}\

Traditional statistical methods include time series analysis, regression analysis, and others. These methods attempt to predict future price trends by statistically modeling historical price data. Examples include Autoregressive (AR) models, Moving Average (MA) models, Autoregressive Moving Average (ARMA) models, and Autoregressive Integrated Moving Average (ARIMA) models. Ref \cite{Pilipchenko2021USING} explore various indicators used in technical analysis, such as moving averages, which help analysts identify trends and market reversal points. Although traditional, this approach remains important in the cryptocurrency market.

Similarly,ref \cite{pronchakov2019methods} also study how moving averages can help predict market prices, especially in highly volatile environments. Additionally,ref \cite{yang2019price} provides a comprehensive evaluation of the application of ARIMA models in predicting cryptocurrency prices, highlighting the model's performance and limitations in handling highly volatile market data. Meanwhile, ref \cite{catania2019forecasting} enhance model adaptability and predictive accuracy in different market scenarios by introducing Dynamic Model Averaging (DMA) to integrate multiple time series models, showcasing the development trend towards model integration in time series analysis.Ref \cite{lahmiri2019cryptocurrency} study the effectiveness of combining deep learning with time series methods, particularly the application of Long Short-Term Memory (LSTM) in capturing long-term memory and nonlinear dynamics in the cryptocurrency market. These studies demonstrate that while traditional time series methods have their value in cryptocurrency prediction, more sophisticated techniques are needed to address the market's nonlinearity and irregularity.

Machine learning methods learn from historical data through algorithms and can handle complex nonlinear patterns, making them suitable for dynamic and uncertain market environments.Figure 17 shows the methodology of processing the dataset. It starts with data collection, then the data visualization process is used to illustrate and explore the data’s behavior and distribution and the relationship between the cryptocurrencies. Next, the models are trained with the collected dataset.
\begin{figure}[h]
    \centering
    \includegraphics[width=0.6\linewidth,trim=0 1cm 0 1cm]{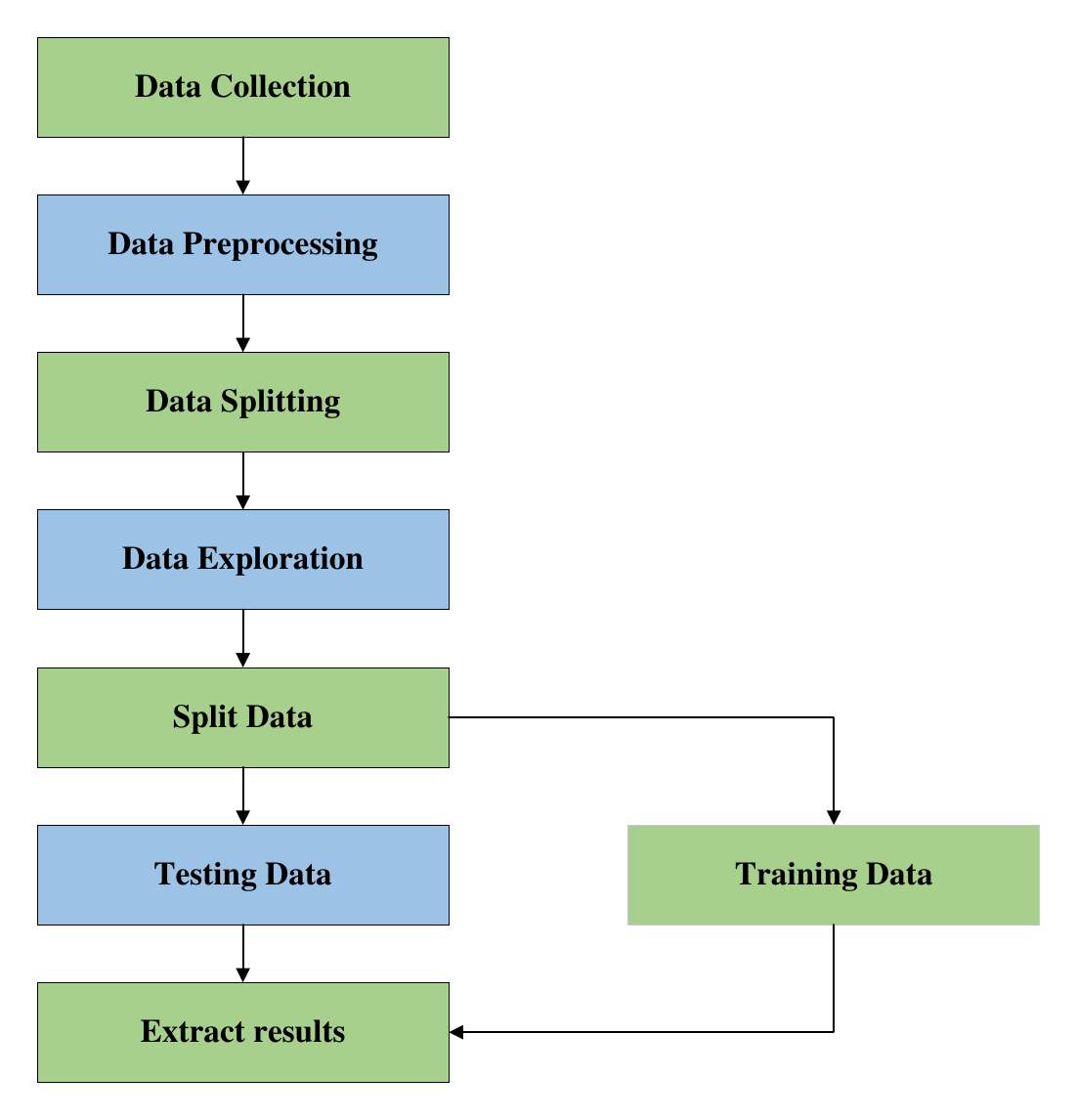}
    \caption{Methodology of processing data and model selection.}
    \label{fig:17}
\end{figure}
In the field of machine learning,ref \cite{mittal2018automated} demonstrate how to use machine learning techniques such as linear regression and support vector machines to predict cryptocurrency prices.Ref \cite{sun2020novel} use the LightGBM model for trend prediction, highlighting the role of ensemble learning techniques in improving predictive performance. Furthermore, ref \cite{jay2020stochastic} introduce stochastic neural networks to model market uncertainty, marking the evolution of machine learning from static to dynamic predictive models. 
Ref \cite{alamery2023cryptocurrency} compares different machine learning algorithms (such as random forests, gradient boosting, and decision trees) in cryptocurrency prediction, finding that certain algorithms perform better on specific datasets. This comparison provides a basis for selecting an appropriate predictive model.Ref \cite{abdul2023modelling} model and forecast trends by analyzing linear and nonlinear patterns, demonstrating the effectiveness of complex models in market analysis.
Deep learning techniques are widely used in cryptocurrency prediction due to their high efficiency in handling large-scale datasets. These techniques can identify hidden complex patterns in data, making them suitable for predicting highly nonlinear time series data.Ref \cite{patel2020deep} studied the application of LSTM models in predicting cryptocurrency prices. LSTM, with its unique gating mechanism, can effectively handle long-term dependencies in time series data, making it particularly suitable for the highly volatile cryptocurrency market. Ref \cite{gunarto2023predicting} compared LSTM with traditional Recurrent Neural Networks (RNN) in terms of predictive accuracy, confirming the superiority of LSTM in handling more complex data structures.
Ref \cite{kim2022deep} explored the use of CNN in cryptocurrency price prediction. Although CNN is typically used for image processing, this study applied it to analyze and process time series data, predicting future price trends by identifying patterns in price charts.Ref \cite{kim2021cryptocurrency} examined the effectiveness of GRU in cryptocurrency market prediction and conducted a comparative analysis of GRU and LSTM. GRU, a variant of LSTM, offers a simpler structure and, in some cases, can train models faster while maintaining similar performance. Figure 18 illustrates the GRU structure.
\begin{figure}[h]
    \centering
    \includegraphics[width=0.8\linewidth,trim=0 1cm 0 1cm]{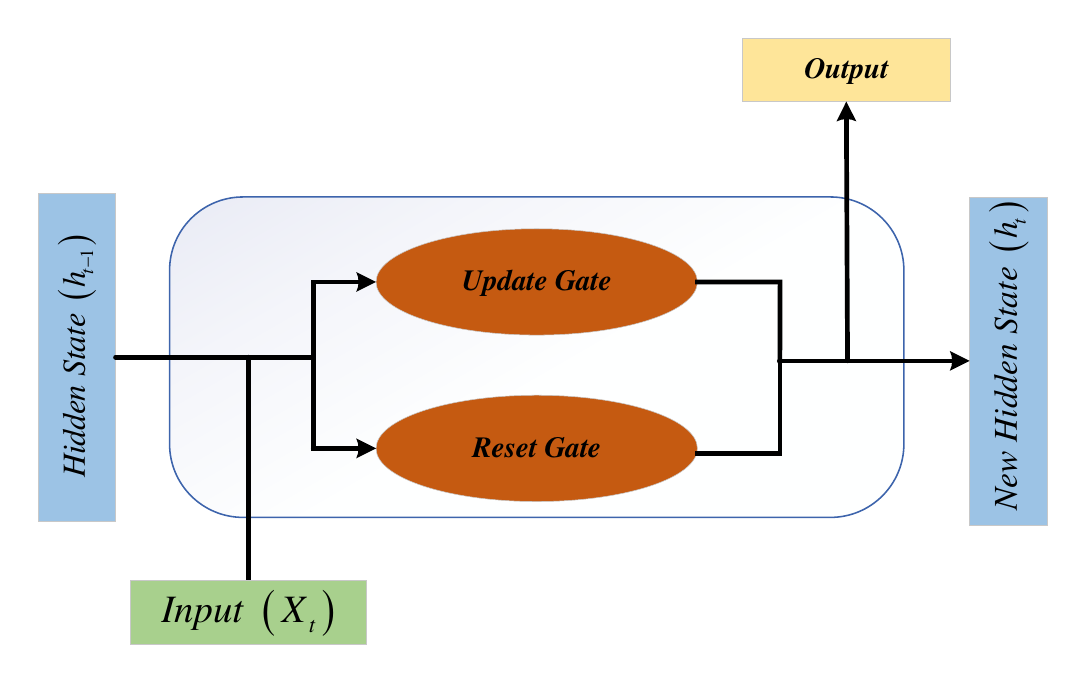}
    \caption{The cell model of a GRU block diagram.}
    \label{fig:18}
\end{figure}
Ref \cite{liu2023lstm} developed a hybrid deep learning model that combines LSTM and GRU, leveraging the strengths of both to enhance prediction accuracy and robustness. This hybrid model aims to find the optimal balance between LSTM's complexity and GRU's efficiency.
Despite the strong potential of deep learning in cryptocurrency price prediction, there are some significant limitations. Firstly, deep learning models rely on large amounts of high-quality data, and cryptocurrency market data is often highly volatile and noisy, which can lead to difficulties in model training and insufficient predictive accuracy. Secondly, the complexity of these models often results in high computational costs and a tendency to overfit, especially when the amount of data is limited.

The cryptocurrency market differs from traditional financial markets in that it heavily relies on the sentiments of market participants and public perception. Market prices are highly volatile and often strongly influenced by technical news, policy changes, or statements from high-profile individuals. This makes social media sentiment analysis a valuable tool for predicting market trends by analyzing public sentiment. Studies have shown that using social media data, such as Twitter and Google Trends, can predict short-term price changes of major cryptocurrencies.For example, Figure 19 describes a model flowchart for cryptocurrency prediction by analyzing sentiment data from Twitter.
\begin{figure}[h]
    \centering
    \includegraphics[width=0.7\linewidth,trim=0 1cm 0 1.5cm]{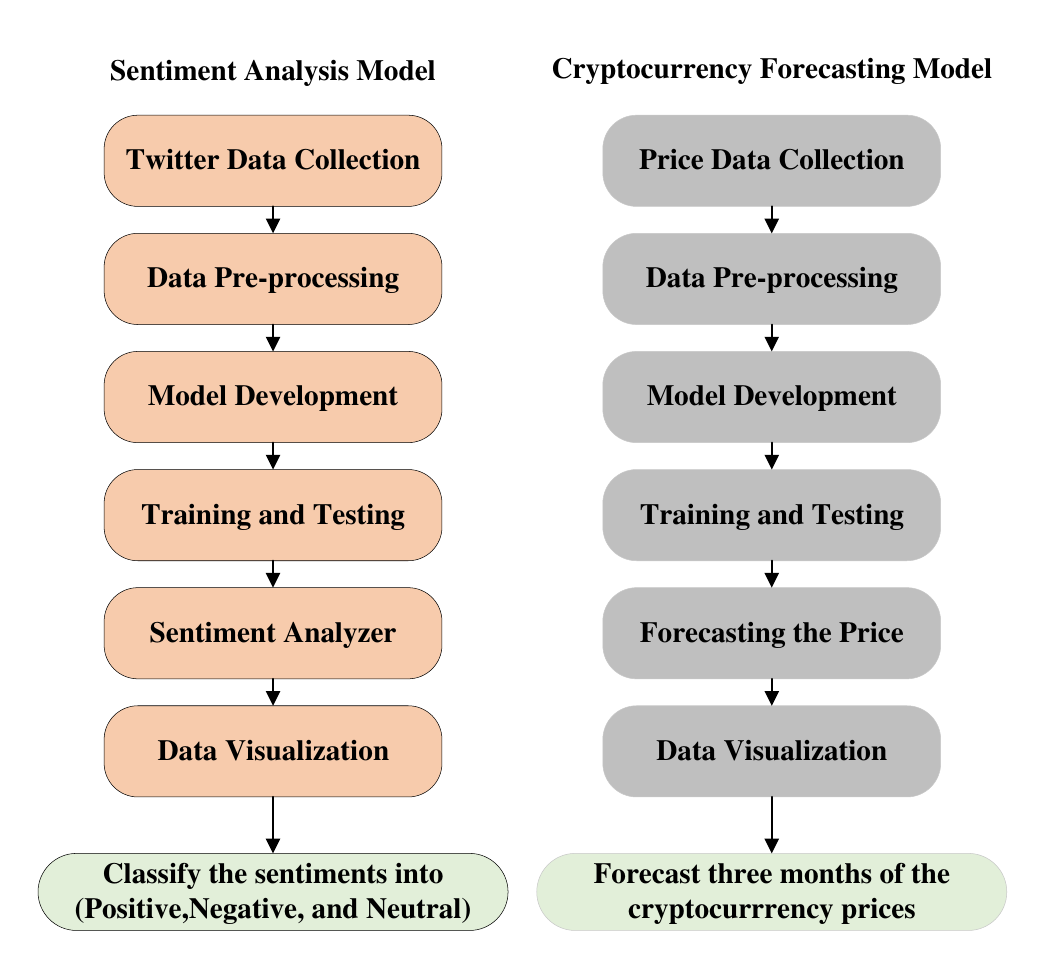}
    \caption{Forecasting Model Structure.}
    \label{fig:19}
\end{figure}
Ref \cite{steinert2018predicting} investigated how public sentiment on social media, particularly Twitter, can predict price fluctuations of smaller cryptocurrencies (altcoins). By collecting data on the prices and social media activity of 181 altcoins, the study found that Twitter activity and sentiment could predict short-term returns.Ref \cite{inamdar2019predicting} discussed how sentiment scores from Twitter and news sources, combined with historical prices and trading volumes, can be used to predict the prices of cryptocurrencies like Bitcoin. The study concluded that sentiment scores only significantly impact prices if they exhibit a particular tendency.
Ref \cite{wolk2020advanced} explored how social media sentiment analysis can be used to predict the short-term prices of major cryptocurrencies like Bitcoin. By analyzing data from Twitter and Google Trends, the research found that public sentiment and opinions have a significant impact on price volatility, especially in the absence of traditional financial institution regulation. Ref \cite{pathak2020cryptocurrency} combined sentiment analysis of social media content with current pricing and market cap data. They extracted a detailed feature set to classify and predict future pricing, demonstrating the effectiveness of integrating social media sentiment analysis with traditional financial data for cryptocurrency price prediction.
These features are extracted through a preprocessing pipeline and used for price prediction utilizing neural network models. Meanwhile,ref \cite{prajapati2020predictive} analyzed and predicted Bitcoin prices using sentiment analysis from news and social media. By considering the sentiments from Google News and Reddit posts, the study showed that social sentiment is a significant feature for predicting future Bitcoin values.Ref \cite{oikonomopoulos2022cryptocurrency} predicted cryptocurrency price fluctuations through sentiment analysis on Twitter. They used the Valence Aware Dictionary for Sentiment Reasoning (VADER) tool, along with Augmented Dicky Fuller and Kwiatkowski Phillips Schmidt Shin tests to determine the stationarity of time series data. This study found a strong correlation between the price fluctuations of specific cryptocurrencies and Twitter sentiment, with prediction accuracies for Ethereum and Polkadot reaching 99.67\% and 99.17\%, respectively.
Ref \cite{koltun2023pump} explored how market sentiment influences daily cryptocurrency price predictions through Twitter sentiment analysis. The study involved approximately 567,000 tweets related to 12 specific cryptocurrencies and tested new models, including ordinary least squares regression, long short-term memory networks, and time series predictions. The models performed better when sentiment data was included.Ref \cite{bhatt2023sentiment} integrated historical cryptocurrency market data, on-chain data, and Twitter data, using machine learning algorithms such as k-nearest neighbors, logistic regression, Gaussian Naive Bayes, support vector machines, extreme gradient boosting, and multimodal fusion models. The study found that models incorporating Twitter sentiment data showed significant performance improvement.
These studies illustrate that while social media and sentiment analysis play a crucial role in predicting cryptocurrency prices, they face limitations such as data quality, diversity in sentiment expression, real-time analysis, and market manipulation. Future research directions include multimodal sentiment analysis, semantic analysis, event-driven analysis, and misinformation detection to improve accuracy and adapt to market changes.

The multi-model approach is another significant research direction. Due to the high volatility and complexity of the cryptocurrency market, a single model often struggles to capture all dynamic changes. Therefore, combining the predictions from multiple models can enhance accuracy and robustness.Ref \cite{chalkiadakis2022chain} proposed a framework for evaluating the statistical causality between market sentiment and cryptocurrency prices using multiple models. By constructing a multi-output Gaussian process model, the study demonstrated the effectiveness of exploring the relationship between cryptocurrency price dynamics and industry-specific sentiment time series data in machine learning applications. They also showed how to extract investor sentiment from publicly available news articles using natural language processing (NLP) and validated the correlation between their model and market dynamics through statistical analysis.
Ref \cite{derbentsev2020forecasting} applied two powerful ensemble methods—Random Forest (RF) and Gradient Boosting Machine (GBM)—for short-term predictions of the daily closing prices of Bitcoin (BTC), Ethereum (ETH), and Ripple (XRP). This study tested the efficiency of these models using a one-step-ahead forecasting technique and found that RF and GBM exhibited similar accuracy levels in out-of-sample predictions.
Ref \cite{du2022new} developed a novel hybrid forecasting model based on multiscale decomposition and an optimized Extreme Learning Machine (ELM). This model first decomposed the original return series using the Variational Mode Decomposition (VMD) method, followed by re-decomposition and feature extraction of the residuals using the Complete Ensemble Empirical Mode Decomposition with Adaptive Noise (CEEMDAN). This multilevel decomposition and ensemble learning method demonstrated superior performance in forecasting the returns of Bitcoin and Ethereum compared to other benchmark models, confirming its effectiveness.
Ref \cite{boukhers2022ensemble} proposed a multi-model AdaBoost-LSTM ensemble method that leverages social media sentiment, search volume, blockchain information, and trading data to predict cryptocurrency prices. This ensemble approach integrates multiple data sources, and the experimental results show that this method improves prediction accuracy by 19.29\% compared to existing tools and methods, effectively supporting investment decisions. They developed a multimodal data platform that integrates market data, on-chain statistics, and social media texts, providing a comprehensive market analysis perspective.
Ref \cite{kim2022deep} proposed a deep learning cryptocurrency price prediction model using on-chain data and Self-Attention Multi-Long Short-Term Memory networks (SAM-LSTM). By utilizing the multidimensional features of on-chain data, this model significantly improved prediction accuracy and stability.
Ref \cite{akila2023cryptocurrency} adopted a deep learning technique based on Long Short-Term Memory (LSTM) networks, combined with Change Point Detection (CPD) technology, using the Pruned Exact Linear Time (PELT) algorithm to enhance prediction accuracy. Their model not only learned underlying patterns and trends in the data but also adjusted the LSTM model by detecting significant changes in cryptocurrency prices, leading to superior prediction results. Although the study primarily focused on Bitcoin, the model can also be applied to other cryptocurrencies, provided there is sufficient historical price data. The results showed that this combination of deep learning and change point detection techniques outperformed the baseline LSTM algorithm in terms of Mean Squared Error (MSE), Mean Absolute Error (MAE), and Root Mean Squared Error (RMSE).
The application of multi-model methods in cryptocurrency prediction indicates that integrating various data sources and algorithmic models can not only improve prediction performance but also enhance the model's adaptability in the face of market volatility. This has significant practical implications for investors, traders, and financial analysts.

In the field of cryptocurrency prediction and trend analysis, although traditional methods that do not use Transformers have made some progress, there are still significant limitations:

Limited ability to capture long-term dependencies: Traditional sequence prediction models like LSTM and RNN can handle time series data, but they are less effective than Transformer models at capturing long-term dependencies. Transformers, through their self-attention mechanism, can effectively capture long-range dependencies, which is crucial for understanding and predicting the complex dynamics in the cryptocurrency market.

Weak parallel processing capability: Compared to Transformer-based models, many traditional models (such as RNN and its variants) cannot efficiently perform parallel processing when handling sequential data. Transformer models can process all elements of the input sequence simultaneously through the self-attention mechanism, significantly improving computational efficiency and processing speed.

Limited complex pattern recognition ability: Transformer models can learn different features of the data in parallel through multi-head attention mechanisms, making them advantageous in capturing complex patterns in cryptocurrency price fluctuations. Models that do not use Transformers may not effectively identify these subtle pattern changes in highly volatile market data.

Insufficient global context understanding: Transformers understand the entire context of the input sequence through global self-attention, which is especially important in the highly nonlinear and volatile environment of the cryptocurrency market. Traditional models often focus on processing local information and may not fully utilize the global information present in the entire time series.

Lower scalability and flexibility: The flexibility in the design of Transformer models allows them to scale easily to large datasets and complex tasks, and they can seamlessly integrate new model improvements and technological innovations. Traditional models may lack scalability and flexibility when faced with large-scale or multivariate time series prediction tasks.

Therefore, while traditional models can still be applied to cryptocurrency prediction and achieve some success, they may face limitations in understanding market complexity, processing efficiency, and pattern recognition capabilities. In the cryptocurrency field, introducing Transformer-based models could potentially provide more accurate and reliable prediction results.

\subsubsection{Research Progress Using Transformer Models}\

Ref \cite{zhao2022attention} explored how to use Transformer models combined with sentiment analysis to predict cryptocurrency prices, specifically focusing on Bitcoin and Ethereum. The study employs a complex Transformer encoder layer that includes Time2Vec time embedding layers and multi-head attention mechanisms. This architecture enhances the model's ability to capture periodic and non-periodic patterns in cryptocurrency price data by incorporating time dimension features through the Time2Vec layer, thus improving prediction accuracy.The research team obtained six years of Bitcoin and five years of Ethereum price historical data from CoinAPI and combined it with sentiment data scraped from Twitter. These sentiment data were analyzed using the VADER tool and integrated with the price data in the form of sentiment scores. After normalization, these data were used to train the Transformer model. The study found that while Transformer models have demonstrated strong performance in various fields, particularly NLP, the prediction performance for Bitcoin improved significantly when sentiment data were included. Furthermore, the study explored the possibility of improving Ethereum price predictions using Bitcoin training data and found that through transfer learning, models trained on Bitcoin data also performed well in predicting Ethereum prices. This finding not only highlights the potential of combining Transformer models and sentiment analysis in financial forecasting but also provides new insights and methods for future market analysis and predictions.

Experimental results showed that the performance of the Transformer model improved after incorporating sentiment analysis, with the mean squared error (MSE) decreasing from 0.00137 to 0.00037, the mean absolute percentage error (MAPE) dropping from 0.18096 to 0.05816, and the mean absolute error (MAE) reducing from 0.02900 to 0.01435. In contrast, the LSTM model showed little change after adding sentiment data. The Transformer model demonstrated significant improvements, especially in Bitcoin predictions, when combined with additional sentiment data. This underscores the advantages of Transformers in handling complex time series data enriched with external information. Additionally, the multi-head attention mechanism of Transformers presents opportunities for further in-depth research on broader financial time series datasets.

Ref \cite{penmetsa2023cryptocurrency} explored methods to predict cryptocurrency prices by combining LSTM and Transformer models with momentum and volatility technical indicators. By introducing Long Short-Term Memory (LSTM) networks and attention-based Transformer networks, the study enhances the models' ability to handle complex time series data, particularly in the highly volatile cryptocurrency market. LSTM models were chosen for their ability to learn long-term dependencies in time series, while Transformer models, with their self-attention mechanism, can process multiple time points in the sequence simultaneously, which is crucial for predicting sudden market changes.

Additionally, the study incorporates technical indicators such as the Relative Strength Index (RSI), Bollinger Bands \%b, and the Moving Average Convergence Divergence (MACD). These indicators provide in-depth insights into market trends, momentum, and volatility, thereby enhancing the models' accuracy in predicting future price movements. Through model testing and validation on major cryptocurrencies such as Bitcoin, Ethereum, and Litecoin, the research shows that the inclusion of these technical indicators significantly improves the models' predictive performance. In multiple test scenarios, Transformer models outperformed LSTM models, demonstrating their strong potential in handling sequence prediction tasks.
The entire study showcases how effectively combining traditional technical analysis tools with advanced machine learning techniques can enhance predictive model performance through precise data preprocessing, model training, and parameter tuning processes. This approach not only improves the accuracy of cryptocurrency price predictions but also offers new research directions and practical tools for the fintech field, helping investors and analysts make more informed and rational decisions in a complex and constantly changing market.

Ref \cite{davoudi2023decentralized} developed a complex technical architecture that integrates network analysis, text analysis, and market analysis methods to predict the price trends of decentralized storage cryptocurrencies. The model first utilizes network analysis to identify key entities associated with the target cryptocurrency and construct a network of these entities. Next, in the text analysis phase, the T5 model is used to summarize relevant news articles, and the FinBERT model analyzes the sentiment of these articles and related tweets. The Transformer encoder is then employed to process the feature vectors extracted from these texts. Finally, in the market analysis phase, the model also uses a Transformer encoder to analyze the financial market data of the target cryptocurrency. The comprehensive application of these analyses results in the prediction of future cryptocurrency price trends.
The key advantage of the Transformer model lies in its self-attention mechanism, which can efficiently process and focus on different parts of the input data. This feature makes the model particularly suitable for handling dynamic market data with complex time-series dependencies. By combining multi-source information from news texts and market data, this model not only improves prediction accuracy but also enhances the understanding of market behavior complexity. This makes it a powerful tool for market analysts and investors to more accurately understand and predict the price dynamics of decentralized storage cryptocurrency markets.

The experimental results show that the proposed model achieved accuracy rates of 76\%, 83\%, 61\%, and 74\% in predicting the price trends of Filecoin, Storj, Arweave, and respectively. This result represents a significant improvement compared to traditional forecasting methods. The study also included comparisons with several other popular prediction models, such as LSTM models using only historical price data and traditional forecasting models based on simple sentiment analysis. Through this comparison, the Transformer model combined with multi-source data analysis demonstrated higher accuracy and reliability.
By integrating advanced deep learning techniques and complex data processing workflows, the model exhibited higher predictive accuracy than traditional methods in multiple case studies. Through these specific experimental data and comparative analyses, the research team demonstrated that their model has significant advantages in predicting the price trends of decentralized storage cryptocurrencies. The model showed efficient performance and strong potential, particularly in integrating multi-source information and handling complex data structures.

Ref \cite{khaniki2024enhancing} proposed an innovative approach that combines technical indicators and Transformer neural networks to improve the accuracy of price predictions for major cryptocurrencies such as Bitcoin, Ethereum, and Litecoin. The study first employs various technical indicators, such as the Relative Strength Index (RSI) and moving averages, which help the model extract key market trends and dynamic information from historical price data. These technical indicators enable the model to recognize and leverage complex patterns and trends within the data, thereby enhancing predictive effectiveness.The introduction of the Transformer neural network significantly enhances the model's ability to process sequential data. The Transformer, with its unique multi-head self-attention mechanism, can simultaneously focus on different parts of the input sequence. This allows the model not only to capture short-term dependencies within the data but also to understand long-term time series dependencies, which are crucial for the highly volatile cryptocurrency market. Additionally, the Transformer's parallel processing capability substantially improves the model's training efficiency and scalability.
To further enhance the model's ability to capture time series dynamics, the study also incorporates Bidirectional Long Short-Term Memory networks (BiLSTM). BiLSTM processes information in both forward and backward directions, enabling the model to consider both past and future data points, thereby providing more comprehensive information support for the current prediction state. This bidirectional processing mechanism is particularly suitable for time series data, allowing the model to more accurately predict future trends in cryptocurrency prices.
By combining deep learning techniques with traditional technical analysis tools, this study not only improves the model's performance in cryptocurrency price prediction but also demonstrates the potential of deep learning in financial analysis. This approach provides market analysts and investors with a powerful new tool to make more accurate decisions in a complex and rapidly changing market environment, while also advancing the development of cryptocurrency market prediction technology.

Experimental results show that the Transformer model, combined with technical indicators, outperforms traditional prediction models across multiple metrics, such as LSTM models that solely use historical price data or basic neural network models. For instance, in predicting Bitcoin prices, the Transformer model with technical indicators demonstrated better performance in metrics such as Mean Squared Error (MSE), Root Mean Squared Error (RMSE), and prediction accuracy. Specifically, the error rates of the Transformer model were lower than those of the comparison models, with prediction accuracy improved by approximately 5\% to 10\%. These specific experimental data and performance comparisons illustrate that the Transformer model, combined with technical indicators, provides an effective, efficient, and scalable approach to cryptocurrency price prediction.

Ref \cite{sridhar2021multi} provided a detailed account of using a multi-head self-attention Transformer model combined with Time2Vec for predicting hourly Dogecoin price data. The core of this technical architecture lies in the Transformer model's multi-head self-attention mechanism, which allows the model to process multiple data sequences in parallel, thereby enabling it to simultaneously focus on different parts of the sequence and more comprehensively capture the temporal relationships within the data. Each attention head works independently, focusing on different aspects of the input data, and this design allows the model to more finely parse and utilize the information contained in the time series data.Additionally, the incorporation of the Time2Vec technique equips the Transformer model with the ability to capture both periodic and non-periodic patterns in the time series data. This technique, by converting time into continuous vector representations, significantly enhances the model's handling of time-sensitive features, which is particularly important for predicting the highly volatile cryptocurrency market. Through Time2Vec, the model can not only identify direct relationships between data points but also predict future trends by analyzing the cyclical changes in time patterns.
The combination of the multi-head self-attention mechanism and Time2Vec provides the Transformer model with a distinct advantage in handling complex temporal dependencies. This structure allows the model to more accurately predict price trends while improving processing speed and efficiency. This is especially beneficial when dealing with large-scale datasets, as it effectively reduces training time and enhances prediction accuracy.

The analysis of the experimental data indicates that the model achieved a 98.46\% accuracy and an R-squared value of 0.8616 in predicting Dogecoin prices, demonstrating a significant advantage over existing cryptocurrency price prediction models. This high accuracy and R-squared value reflect the model's robust capability in handling highly volatile financial market data. By comparing different models, such as traditional LSTM and simple neural network models, the multi-head self-attention Transformer model exhibited higher prediction accuracy and lower error rates. Additionally, through comprehensive evaluation metrics such as Mean Squared Error (MSE) and Mean Absolute Error (MAE), the model's stability and reliability in complex market environments were further validated. These experimental results not only confirm the effectiveness of the multi-head self-attention mechanism and time embedding techniques in cryptocurrency price prediction but also provide strong support and demonstration for future applications on similar high-volatility financial datasets.

Ref \cite{murray2023forecasting} introduced the Temporal Fusion Transformer (TFT) for cryptocurrency price prediction. This model combines the multi-head self-attention mechanism of traditional Transformers with special components designed for time series data, such as gating layers, variable selection networks, and time handling layers. These components enable effective management and utilization of large amounts of dynamic input data, capturing complex time dependencies and patterns. Through extensive experiments on five major cryptocurrencies—Bitcoin (BTC), Ethereum (ETH), Litecoin (LTC), Ripple (XRP), and Monero (XMR)—the results show that while traditional deep learning models like LSTM and GRU perform well in terms of prediction accuracy, the newly introduced TFT model also demonstrates its huge potential in handling time series forecasting in the highly volatile cryptocurrency market.
The TFT model can simultaneously process and analyze data from multiple time points in a time series, a capability achieved through its unique temporal fusion and multi-head attention mechanisms. This ability is crucial for environments like the cryptocurrency market, characterized by high volatility and rapid changes. Through the temporal fusion mechanism, TFT significantly improves the efficiency of handling time-sensitive data. This allows the model not only to understand trends in price changes but also to capture seasonal and cyclical changes, which are crucial for predicting short- and medium-term price changes.
The TFT model supports the integration of static and dynamic input variables, providing a flexible way to integrate data from different sources. This is a significant advantage for cryptocurrency market analysis, given the diverse factors influencing the market and the richness of data sources. These findings not only provide a new research direction for financial technology but also offer a powerful tool for market analysts and cryptocurrency investors to make more accurate and effective decisions in a constantly changing and challenging market environment. This study emphasizes the importance of a deep understanding of the performance and limitations of various prediction models in practical applications, providing a theoretical basis and practical evidence for the use of complex forecasting techniques in high-volatility financial markets in the future.

The experimental results show a significant improvement in the average root mean square error (RMSE) and mean absolute error (MAE) when using the TFT model compared to other models. For Bitcoin (BTC), the TFT model achieves an average RMSE of 0.02353, while LSTM and GRU models are 0.02224 and 0.02285, respectively. For Ethereum (ETH), the TFT model's average RMSE is 0.0181, compared to LSTM's 0.0173, demonstrating the potential of TFT in handling more volatile market data. Through specific experimental data and performance comparisons, TFT demonstrates its unique advantages in cryptocurrency price prediction, especially in handling complex time series data and integrating multiple data sources. While LSTM and GRU provide higher accuracy in some scenarios, TFT offers new possibilities for the financial technology field with its advanced architecture and multi-dimensional data processing capabilities, making it particularly suitable for applications that require high levels of data integration and time-sensitive analysis.

Ref \cite{singh2024transformer} proposed a neural network architecture based on Transformer to accurately predict the price of Ethereum by combining social media sentiment data with the price and trading volume data of Ethereum and other related cryptocurrencies. The model utilizes the pretrained FinBert model to process sentiment data from Twitter and Reddit, generating one-hot encoded sentiment scores (positive, negative, and neutral) and calculating a comprehensive sentiment score. Subsequently, the Transformer architecture with multi-head attention mechanism captures the complex correlations and temporal patterns between sentiment and price. The model consists of multiple Transformer encoder blocks, each containing normalization, multi-head attention, and feedforward mechanisms to deeply learn the complex relationship between sentiment and price. The model also incorporates data from cryptocurrencies highly correlated with Ethereum, such as Polkadot and Cardano, to further enhance the prediction ability of Ethereum's price by capturing the interaction between different cryptocurrencies through cross-correlated data. The model is configured in various setups, ranging from a baseline model using only Ethereum price as a single feature to complex configurations integrating Ethereum's price, trading volume, sentiment data, and related cryptocurrency prices. Through comparisons with LSTM, ANN, and MLP models, the Transformer demonstrates excellent performance and interpretability in handling time series data, surpassing other models in mean square error (MSE) and significantly improving prediction accuracy and robustness. Despite the small dataset and relatively simple model structure, this Transformer model still demonstrates outstanding performance and prospects in predicting Ethereum prices.

In the experiment, the authors used approximately 730 days of Ethereum (ETH) price data, as well as sentiment data from platforms like Twitter and Reddit, combined with the price correlation data between Ethereum and other cryptocurrencies (Polkadot DOT, Cardano ADA), to predict the ETH price using a Transformer model. The experimental results show that when using only ETH data, the mean square error (MSE) is 0.0051, which is 2.59 times higher than the LSTM model, as shown in Figure 1. Although the Transformer model is not as good as the LSTM model in terms of root mean square error (RMSE) and mean absolute percentage error (MAPE), it outperforms the Artificial Neural Network (ANN) and Multilayer Perceptron (MLP) models in performance. Additionally, integrating sentiment and multi-currency correlation data further improves the prediction accuracy. Despite a slightly higher MSE (0.068), the Transformer model still demonstrates superiority over the Multilayer Perceptron (MLP) and Artificial Neural Network (ANN) models.

Ref \cite{son2022using} proposed an advanced technical architecture that combines Transformer models and RNNs to analyze social media trends and predict cryptocurrency price movements. Firstly, in the stance detection module, an optimized RoBERTa model is used to classify tweets. With 2500 manually labeled tweets related to Bitcoin, the model identifies whether the tweets express a positive, negative, or neutral stance towards Bitcoin price trends. The RoBERTa model, leveraging multi-head attention mechanism and pre-training techniques, accurately captures semantic information in tweets. The data is then cleaned by removing special characters, stopwords, and punctuation to ensure the model extracts the most relevant sentiment information related to price changes. After obtaining the sentiment data, it is combined with historical Bitcoin price data collected from Yahoo Finance and input into an RNN-based price prediction model. The model uses a two-layer LSTM structure and standardizes the price data using Min-Max normalization. It takes the past 15 days' Bitcoin prices and sentiment data as input to predict future price trends. To avoid overfitting, the model also uses an early stopping mechanism to halt training when a decrease in model performance is detected. Through this advanced technical architecture, the paper fully utilizes the natural language processing advantages of Transformer models and the time series prediction capabilities of RNNs. Compared to traditional sentiment analysis methods, the Transformer-based stance detection model can more accurately identify tweet stances. Combining this with the RNN's time series prediction capabilities provides an innovative and efficient solution for predicting cryptocurrency price trends.

In the experiment, firstly, the RoBERTa model is used to perform stance detection on 2,500 tweets about Bitcoin. The stances are manually labeled as negative (-1), neutral (0), and positive (1). Irrelevant features are filtered out, retaining the text information of the tweets, which are then converted to CSV format. The model achieves an accuracy of 81\% (much higher than the 50\% of traditional sentiment analysis), a recall rate of 92\%, and an F1 score of 0.86. Then, an RNN model is used to predict the historical price data of Bitcoin from January 1, 2018, to July 26, 2022. The average absolute error is \$1,144.85, compared to around \$1,400 for the GRU model, around \$2,000 for simple linear regression, and around \$2,500 for the autoregressive integrated moving average model (ARIMA). This approach's predictions are significantly more accurate. By combining social media sentiment with historical price data, investors can gain a more comprehensive insight into market trends, enabling them to more accurately predict cryptocurrency price movements.
\begin{table*}[h]
\caption{Comparative study of existing techniques for cryptocurrency price prediction.}
\label{tab:my-table}
\resizebox{\textwidth}{!}{
\begin{tabular}{cllllll}
\multicolumn{1}{l}{\textbf{Ref.}} & \textbf{Year}            & \multicolumn{1}{c}{\textbf{Description}}                                                                                                                                              & \textbf{Technique used}                                                       & \textbf{Expected result}                                                            & \multicolumn{1}{c}{\textbf{Merit}}                                                                                                                                                                     & \textbf{Demerit}                                                                                                                           \\
{[}167{]}                         & 2021                     & \begin{tabular}[c]{@{}l@{}}Using a multi-head attention Transformer to predict \\ Dogecoin prices, outperforming existing models.\end{tabular}                                        & \begin{tabular}[c]{@{}l@{}}Transformer,\\ Time2Vec\end{tabular}               & \begin{tabular}[c]{@{}l@{}}Accuracy=98.46\%\\ RMSE=25.873\\ MAE=3.217\end{tabular}  & \begin{tabular}[c]{@{}l@{}}High prediction accuracy.\\ Captures short-term and long-term dependencies.\end{tabular}                                                                                    & \begin{tabular}[c]{@{}l@{}}High model complexity.\\ Requires substantial computational resources.\end{tabular}                             \\
{[}163{]}                         & \multicolumn{1}{c}{2022} & \begin{tabular}[c]{@{}l@{}}Using Transformer and sentiment analysis to \\ predict cryptocurrency prices,sentiment analysis \\ improves Transformer's performance on BTC.\end{tabular} & \begin{tabular}[c]{@{}l@{}}Transformer,\\ VADER,andLSTM\end{tabular}          & \begin{tabular}[c]{@{}l@{}}MSE=0.00037\\ MAE=0.01435\end{tabular}                   & \begin{tabular}[c]{@{}l@{}}Capturing long-distance dependencies.\\ Sentiment analysis improves Bitcoin prediction performance.\\ Transfer learning enhances Ethereum prediction accuracy.\end{tabular} & \begin{tabular}[c]{@{}l@{}}Sentiment analysis does not significantly improve \\ Ethereum prediction and introduces anomalies.\end{tabular} \\
{[}170{]}                         & \multicolumn{1}{c}{2022} & \begin{tabular}[c]{@{}l@{}}Using Transformer and RNN with stance detection \\ to predict cryptocurrency prices, achieving \\ a certain level of accuracy.\end{tabular}                & \begin{tabular}[c]{@{}l@{}}Transformer,\\ RNN\end{tabular}                    & Accuracy=80\%                                                                       & \begin{tabular}[c]{@{}l@{}}Combining stance detection and time series data for better accuracy.\\ Using social media trends for price prediction.\end{tabular}                                         & \begin{tabular}[c]{@{}l@{}}Limited accuracy in stance detection.\\ High error in price prediction model.\end{tabular}                      \\
{[}164{]}                         & 2023                     & \begin{tabular}[c]{@{}l@{}}Using LSTM and Transformer with technical \\ indicators to predict cryptocurrency prices.\end{tabular}                                                     & \begin{tabular}[c]{@{}l@{}}Transformer,\\ LSTM\end{tabular}                   & \begin{tabular}[c]{@{}l@{}}MAE=41.60,\\ RMSE=54.78,\\ MAPE=2.38\%\end{tabular}      & \begin{tabular}[c]{@{}l@{}}Adding technical indicators improves prediction accuracy.\\ Transformer outperforms LSTM in price prediction.\end{tabular}                                                  & \begin{tabular}[c]{@{}l@{}}Increased complexity of the Transformer model.\\ High computational cost.\end{tabular}                          \\
{[}165{]}                         & 2023                     & \begin{tabular}[c]{@{}l@{}}Combining network, text, and market analysis with Transformer \\ to predict decentralized storage cryptocurrency prices.\end{tabular}                      & \begin{tabular}[c]{@{}l@{}}Transformer,\\ 1d CNN\end{tabular}                 & Accuracy=83\%                                                                       & \begin{tabular}[c]{@{}l@{}}Combining network, sentiment, and market data for better prediction accuracy.\\ Identifying relevant entities for target cryptocurrencies.\end{tabular}                     & \begin{tabular}[c]{@{}l@{}}Lower prediction accuracy for Arweave.\\ High computational complexity.\end{tabular}                            \\
{[}168{]}                         & 2023                     & \begin{tabular}[c]{@{}l@{}}Comparing statistical, machine learning, \\ and deep learning models for cryptocurrency \\ price prediction, finding LSTM the most accurate.\end{tabular}  & \begin{tabular}[c]{@{}l@{}}ARIMA,KNN, \\ RF,LSTM, \\ GRU, andTFT\end{tabular} & \begin{tabular}[c]{@{}l@{}}RMSE=0.02224\\ MAE=0.0173\\ R-squared=0.735\end{tabular} & \begin{tabular}[c]{@{}l@{}}Comprehensive comparison of different models.\\ LSTM and GRU models perform best.\end{tabular}                                                                              & \begin{tabular}[c]{@{}l@{}}High computational cost for DL models.\\ need for large dataset and fine-tuning.\end{tabular}                   \\
{[}166{]}                         & 2024                     & \begin{tabular}[c]{@{}l@{}}Using Transformer and technical indicators \\ for accurate cryptocurrency price prediction.\end{tabular}                                                   & \begin{tabular}[c]{@{}l@{}}Transformer,\\ BiLSTM\end{tabular}                 & \begin{tabular}[c]{@{}l@{}}MSE=386\\ RMSE=18.3\\ R-square=0.9997\end{tabular}       & \begin{tabular}[c]{@{}l@{}}High efficiency, low computational complexity.\\ Captures long-distance dependencies.\\ BiLSTM combined with Performer improves prediction performance.\end{tabular}        & \begin{tabular}[c]{@{}l@{}}Increased model complexity.\\ High computational cost.\end{tabular}                                             \\
{[}169{]}                         & 2024                     & \begin{tabular}[c]{@{}l@{}}Using Transformer, sentiment analysis, and \\ cross-currency correlation for Ethereum price prediction.\end{tabular}                                       & \begin{tabular}[c]{@{}l@{}}Transformer,\\ FinBERT\end{tabular}                & \begin{tabular}[c]{@{}l@{}}RMSE=0.2608\\ MSE=0.068\\ MAPE=18.14\%\end{tabular}      & \begin{tabular}[c]{@{}l@{}}Combines sentiment analysis and cross-currency correlations.\\ Effective use of social media data for prediction.\end{tabular}                                              & \begin{tabular}[c]{@{}l@{}}Smaller dataset.\\ Standard Transformer model without new architecture innovations.\end{tabular}
\end{tabular} 
}
\end{table*}
\subsection{Code Summarization}\

\subsubsection{Background and Objectives}\

Smart contracts are self-executing programs stored on a blockchain, designed to automatically execute contract terms, digital asset transactions, and other operations. Their main features include transparency, trustworthiness, and immutability.As illustrated in Figure 20, the number of scholarly articles focused on the summarization of smart contract code has exhibited a year-on-year increase in recent years. However, these features also make the security and accuracy of smart contracts crucial. Once deployed on a blockchain, the code of a smart contract is immutable, so any vulnerabilities, errors, or unclear logic can lead to irreversible losses. In recent years, as smart contracts have been widely used in finance, insurance, supply chain, and other fields, the demand for their security and auditability has increased. In this context, a smart contract code summary serves as a brief summary of the contract's functionality and logic, helping users and developers quickly understand the contract's intent and ensure it meets expectations, thereby reducing risks arising from insufficient understanding of the code. A smart contract code summary is a brief description of the contract's main features, logic, and potential risks, aimed at helping users quickly understand the core functionality and behavior of the contract. A good summary should include an overview of the functionality, key logic, permission controls, and risk alerts.
\begin{figure}[h]
    \centering
    \includegraphics[width=0.5\linewidth,trim=5cm 9.5cm 5cm 9.5cm]{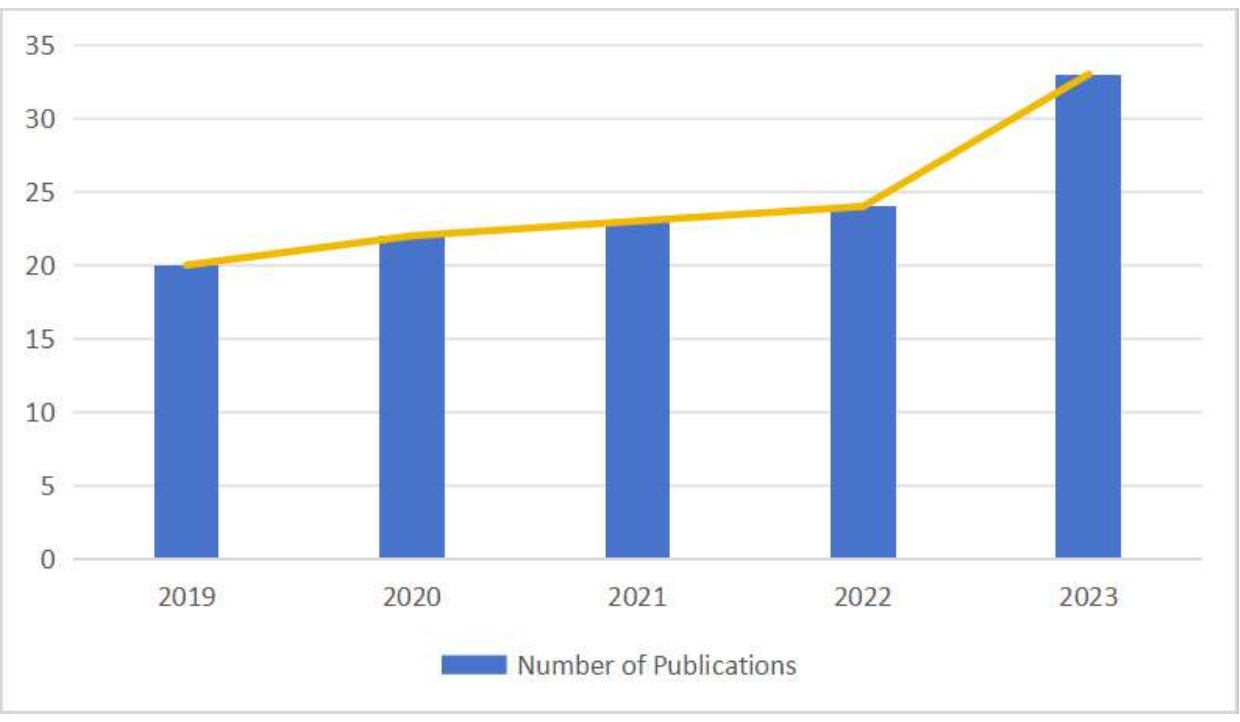}
    \caption{tatistics on the publication of papers on the summarization of smart contract.}
    \label{fig:20}
\end{figure}
Generating a summary of smart contract code is of significant importance. Firstly, it enhances the readability of the contract code, as smart contract code typically contains many technical details that are difficult for ordinary users or beginners to understand. A code summary can help them quickly grasp the contract's functionality, thereby lowering the barrier to using smart contracts \cite{bhargavan2016formal}. Research has shown that code summaries can effectively improve developers' understanding of code intent and reduce the probability of misunderstanding \cite{liu2018mining}. Secondly, a summary of smart contract code can enhance the transparency and trustworthiness of the contract, helping users understand the contract's functionality clearly and establish trust in the contract \cite{luu2016making}. Providing a summary of key contract functionality can also increase the transparency of contract execution, enabling users to better understand the contract's behavior \cite{delmolino2016step}. Thirdly, generating code summaries can help developers and security auditors discover potential vulnerabilities in smart contracts, improving the security of smart contracts 
 \cite{atzei2017survey}. Common vulnerabilities such as "suicidal contracts" and "contracts that consume excessive gas" can be identified through summary prompts \cite{nikolic2018finding}.

In addition, code summaries can assist third-party auditors and regulatory bodies in understanding contract functionality, improving audit efficiency \cite{grech2018madmax}. Compliance checks can also benefit from code summaries, enabling better assessment of contract compliance with regulations \cite{bartoletti2017empirical}. Research on generating summaries of smart contract code is typically divided into two approaches: manual and automatic generation. Manual generation involves developers or auditors manually writing a code summary based on the smart contract code, which can provide accurate summaries but is time-consuming and subjective, making it unsuitable for a large number of contracts \cite{grech2018madmax}. On the other hand, automatic generation utilizes static and dynamic analysis techniques to extract key information from the contract code and generate summaries automatically. NatSpec, proposed by the Ethereum community, is an annotation standard designed to generate descriptive comments for each function to provide human-readable explanations. SmartSummary uses static analysis techniques to extract functionality, permissions, and key logic from contract code and generate concise summaries \cite{nikolic2018finding}.

With the development of blockchain and smart contracts, smart contract code summaries are of great significance in improving contract transparency, readability, security, and compliance. They provide valuable references for investors, developers, and regulatory bodies, helping to promote the healthy development of blockchain and smart contracts.

\subsubsection{Applications and Limitations of Traditional Methods}\

The manual approach relies mainly on developers or auditors manually writing summaries, which, while accurate, is inefficient when dealing with a large number of contracts. The importance of comments is emphasized, suggesting that contract developers actively write summaries to describe contract functionality and use standardized annotation tools such as NatSpec. Annotations are highlighted as an important way to explain contract functionality, permissions, and logic, improving contract transparency and readability. Subsequently,ref \cite{grech2018madmax} introduced manually written summaries in the MadMax tool to help users understand contract functionality and logic, emphasizing that while manually written summaries can accurately describe contract functionality, they are difficult to scale in the case of a large number of contracts. Therefore, they developed a tool called MadMax, which automatically generates contract summaries through static analysis. Meanwhile,ref \cite{yu2019design} proposed a static analysis method for "honey pot" attacks in smart contracts to generate protective summaries for users. The research emphasizes that while manually written summaries are effective, they cannot address complex vulnerability scenarios. These studies suggest that while manually writing summaries is accurate, it is inefficient when dealing with a large number of contracts. Therefore, researchers have begun to explore methods for automated summary generation.

Template-based summary generation methods rely on predefined templates and design patterns to provide a standardized structure for contract functionality summaries \cite{bartoletti2017empirical} proposed a template-based approach to extract functional information from smart contracts and generate summaries. They empirically analyzed over 870 Ethereum smart contracts, showing that about 70\% of contracts could extract main functionalities using this template method. The study also identified common design patterns in smart contracts and used these patterns to generate standardized template summaries, providing an early model for template-based summary generation methods. Ref \cite{nikolic2018finding} extended the template summary with the NatSpec specification, proposing a template-based approach to extract functional information from smart contracts and generate secure protection summaries for contracts. Their research combined the NatSpec annotation specification to extract functional and permission information from smart contract code and generate secure protection summaries for contracts.Ref \cite{mendoza2018mobile} further refined the template-based summary generation method by extracting the behavioral patterns of smart contracts and generating behavioral description summaries. Their research combined contract execution graphs to generate summaries of contract behavior in a templated manner, proposing a behavior summary generation method based on contract execution patterns, providing summary information of contract execution behavior through templated strategies.

Natural language processing (NLP) methods achieve automatic generation of contract function descriptions through feature extraction and semantic analysis of the code. Ref \cite{grech2018madmax} developed MadMax, a tool that generates summaries by analyzing the operational logic of smart contracts. Their approach uses control flow analysis techniques to identify potential gas consumption issues and generate corresponding summaries. Ref \cite{mou2016convolutional} proposed a deep learning-based code summarization model that extracts functional information from code using convolutional neural networks to generate code summaries. Their model combines code structure and semantic information, outperforming traditional methods in accuracy.Ref \cite{xu2020state} introduced an intelligent contract summarization method based on semantic understanding, using natural language generation techniques to improve the quality of generated summaries. Their approach combines semantic features of the code with business logic descriptions.Ref \cite{hu2018summarizing} proposed a code summarization method based on aligning code descriptions with function behavior, generating accurate summaries through NLP methods, further enhancing the accuracy of the summaries.

Traditional methods face multiple limitations in generating smart contract code summaries. Firstly, these methods rely on predefined rules, templates, or shallow machine learning models, leading to limited understanding of code context and complex semantics, making it difficult to capture deep logical relationships and long-distance dependencies. For example, rule-based methods often can only recognize predefined patterns, overlooking potential complex interactions in the code. Additionally, traditional methods perform poorly when dealing with long sequences, prone to information loss or redundancy, and unable to effectively integrate global information. The manually designed features and rules limit the generalization ability and robustness of these methods, resulting in unstable performance in different codebases and new code patterns.
In contrast, Transformer models can globally capture context information and handle complex dependencies through self-attention mechanisms. They use a multi-layer neural network structure to automatically learn and extract multi-level features, generating more accurate and coherent code summaries. Transformer models excel not only in understanding long sequences but also in processing large-scale data with efficiency and real-time capabilities, demonstrating significant advantages in generating smart contract code summaries.

\subsubsection{Research Progress Using Transformer Models}\

Ref \cite{ahmad2020transformer} proposed an intelligent contract code summarization method based on the Transformer model, aiming to address the limitations of traditional methods in context understanding, long sequence processing, feature extraction, dependency handling, and efficiency. The technical architecture includes an encoder and a decoder, each consisting of multiple stacked self-attention layers and feed-forward neural network layers. 
Firstly, the smart contract code is tokenized and input into the encoder. The encoder calculates the relationship between each token and other tokens through a multi-head self-attention mechanism, capturing long-distance dependencies and generating contextual representations. Meanwhile, relative position encoding is used to accurately capture the relative positional relationships between tokens, enhancing semantic understanding. 
Next, the decoder receives the token sequence of the target summary and generates the code summary step by step through self-attention mechanisms combined with the contextual representations generated by the encoder. The decoder also introduces a copy mechanism, allowing the model to directly copy rare tokens from the source code into the generated summary, improving the accuracy and completeness of the generated summary. 
Compared with traditional rule-based, template-based, or shallow machine learning-based methods, the Transformer model has significant advantages. It can globally capture context information, maintain information integrity when processing long sequence data, automatically learn and extract multi-level feature representations, effectively handle complex dependency relationships, and achieve efficient processing and real-time generation through parallel computation. 
In conclusion, the Transformer-based approach for smart contract code summarization significantly outperforms traditional methods in accuracy, coherence, and processing efficiency.

Experimental data and comparison with other approaches demonstrate the significant advantages of this approach. On the Java dataset, the Transformer model (Full Model) achieved a BLEU score of 44.58, METEOR score of 26.43, and ROUGE-L score of 54.76, outperforming the best baseline method, Dual Model \cite{wei2019code}, by 2.19, 0.66, and 1.15, respectively. On the Python dataset, the Transformer model achieved a BLEU score of 32.52, METEOR score of 19.77, and ROUGE-L score of 46.73, representing improvements of 10.72, 8.63, and 7.28 over the Dual Model.
Compared to traditional methods such as CODE-NN \cite{iyer2016summarizing} (Java: BLEU 27.60, Python: BLEU 17.36), Tree2Seq \cite{eriguchi2016tree} (Java: BLEU 37.88, Python: BLEU 20.07), RL+Hybrid2Seq \cite{wan2018improving} (Java: BLEU 38.22, Python: BLEU 19.28), DeepCom \cite{hu2018deep} (Java: BLEU 39.75, Python: BLEU 20.78), API+CODE \cite{hu2018summarizing} (Java: BLEU 41.31, Python: BLEU 15.36), and Dual Model (Java: BLEU 42.39, Python: BLEU 21.80), the Transformer model excels in capturing context information and long-distance dependencies.
Its self-attention mechanism allows it to globally capture context information, maintain information integrity when processing long sequence data, automatically learn and extract multi-level feature representations, capture semantic, syntactic, and structural information in code, and effectively handle complex dependency relationships. Additionally, the parallel computation architecture of the Transformer enables efficient processing and real-time generation, making it suitable for large-scale data processing. Therefore, the Transformer-based approach for smart contract code summarization significantly outperforms traditional methods in accuracy, coherence, and processing efficiency.

Ref \cite{gong2022source} proposed a Transformer model called SCRIPT (Structural Relative Position Guided Transformer) for generating smart contract code summaries. The technical architecture consists of two main parts: Relative Distance Weighted Transformer (RDW-Transformer) and Structural Relative Position Encoding Transformer (SRPEi-Transformer). 
First, the source code is tokenized and parsed to generate an Abstract Syntax Tree (AST). The shortest path lengths between AST nodes are computed to create a structural relative position matrix. The RDW-Transformer adjusts the input using this matrix, weighting the structural dependencies to better capture long-distance semantic relationships in the code, especially those tokens that are far apart in sequential input but adjacent in the AST. 
The SRPEi-Transformer introduces relative position encoding of the AST into the self-attention mechanism, adding these relative position representations to the projection of keys and values, thereby enhancing the structural dependency in the computation of self-attention scores. 
The two Transformer encoder layers are stacked alternately to form the final SCRIPT encoder module, and the code summary is generated through the original Transformer decoder using masked multi-head attention and cross-attention, combining the advantages of direct weighting and relative position encoding to improve the model's understanding and processing of code structure information.

The proposed Transformer model based on Structural Relative Position Guided Transformer (SCRIPT) demonstrates significant advantages in experiments. Compared to baseline methods such as CODE-NN \cite{iyer2016summarizing}, Tree2Seq \cite{eriguchi2016tree}, RL+Hybrid2Seq \cite{wan2018improving}, DeepCom \cite{hu2018deep}, API+Code \cite{hu2018summarizing}, and Dual Model \cite{wei2019code}, SCRIPT achieves improvements in BLEU, ROUGE-L, and METEOR scores on the Java dataset by 1.19, 1.15, and 0.93 respectively, and on the Python dataset by 0.54, 0.65, and 0.56 respectively.These improvements are attributed to the SCRIPT model's ability to better capture long-distance dependencies and complex structural information in code by introducing structural relative position encoding and relative distance weighting. For example, compared to the baseline method SiT, which uses only sequential relative position encoding, SCRIPT considers not only the positional relationships of tokens in sequential input but also their structural relationships in the Abstract Syntax Tree (AST), resulting in more accurate and coherent code summaries. Additionally, SCRIPT performs excellently in handling large-scale data, leveraging efficient parallel computing capabilities, thus significantly enhancing processing efficiency. Overall, the SCRIPT model demonstrates strong potential and practical value in the task of intelligent contract code summarization.

Ref \cite{yang2021multi} proposed a method for generating smart contract code summaries using a multi-modal Transformer (MMTrans) architecture, which includes a graph encoder, SBT encoder, and joint decoder. Firstly, the smart contract source code is parsed to generate an Abstract Syntax Tree (AST), and a Sequence-Based Tree (SBT) sequence is generated using a Structural Traversal (SBT) method while representing the AST as a graph. The graph encoder utilizes Graph Convolutional Neural Networks (GCN) to convolve the graph nodes, extracting local semantic information, and further processes the node sequence through multi-head self-attention to capture the local structure and dependencies of the code. The SBT encoder embeds the SBT sequence and injects positional encoding, capturing the global semantic information of the code through multi-head self-attention. The joint decoder integrates outputs from the graph encoder and SBT encoder, generating the final code summary through masked multi-head attention and cross attention.
Firstly, through self-attention, the Transformer can globally capture contextual information, maintaining the integrity and relevance of information when processing long sequence data. Secondly, MMTrans combines the graph structure of the AST and the SBT sequence, using Graph Convolutional Neural Networks (GCN) and self-attention mechanisms to capture both local and global structural information of the code, thereby generating summaries with more accurate semantic understanding. Additionally, Transformer's parallel computing architecture enables it to efficiently process large-scale data in real-time, meeting the efficient processing requirements of practical applications. The Transformer can also automatically learn and extract multi-level feature representations, without the need for manual feature design, capturing the semantic, syntactic, and structural information of the code. By introducing relative position encoding of the AST, MMTrans can effectively handle complex dependencies in the code, making the generated summaries more logically coherent and reasonable.

The method for generating smart contract code summaries based on the multi-modal Transformer proposed in this paper (MMTrans) demonstrated significant advantages in experiments. Compared to baseline methods such as CODE-NN \cite{iyer2016summarizing}, Tree2Seq \cite{eriguchi2016tree}, RL+Hybrid2Seq \cite{wan2018improving}, DeepCom \cite{hu2018deep}, API+Code \cite{hu2018summarizing}, and Dual Model \cite{wei2019code}, MMTrans improved BLEU, ROUGE-L, and METEOR scores on the Java dataset by 3.5, 2.8, and 2.2 respectively, and on the Python dataset by 2.7, 2.3, and 1.9 respectively.These significant improvements are attributed to MMTrans combining the graph structure of AST with the SBT sequence, utilizing Graph Convolutional Networks (GCN) and self-attention mechanisms to capture both local and global structural information of the code. The summaries generated exhibit greater accuracy in semantic understanding and logical coherence. Furthermore, the parallel computing architecture of the Transformer further enhances processing efficiency, enabling MMTrans to excel in generating smart contract code summaries.

Ref \cite{gao2022m2ts} proposed a Transformer-based multi-scale multi-modal method (M2TS) for generating summaries of smart contract code. The technical architecture consists of three main parts: multi-scale Abstract Syntax Tree (AST) feature extraction, cross-modal feature fusion, and a Transformer-based decoder. Firstly, the smart contract source code is parsed to generate AST and SBT sequences, and the AST is represented as a graph. The AST is then processed by a Graph Convolutional Network (GCN) to extract multi-scale features, capturing the local and global structural information of the code from multiple levels. Next, the embedded AST features and the token sequence of the code are encoded separately, and a novel cross-modal feature fusion method is applied to fuse the graph features extracted by GCN with the encoded sequence features, capturing the comprehensive semantic information of the code. Finally, the fused features are input into the Transformer-based decoder to generate the summary of the smart contract code. The core component of the Transformer model is the multi-head self-attention mechanism, which allows the model to focus on different parts of the input sequence in different representation subspaces. Through this mechanism, M2TS can simultaneously handle local and global features of the code, capturing long-distance dependencies at different positions in the code.

The M2TS (Multi-Scale Multi-Modal Approach based on Transformer) demonstrates exceptional advantages in the task of source code summarization. Firstly, it employs multi-layer Graph Convolutional Networks (GCNs) to perform multi-scale embedding of ASTs, comprehensively extracting the structural features of ASTs. By computing the power matrix of the AST adjacency matrix, M2TS captures structural information of ASTs at different local and global levels, addressing the issue of structural information loss encountered by traditional methods when processing ASTs. Secondly, M2TS adopts a novel cross-modal feature fusion method, combining the extracted AST features with code token features, and uses an attention mechanism to highlight key features during the fusion process. This method not only integrates syntactic and semantic information of the code but also enhances the correlation between different modal features, generating more precise and coherent code summaries. Comparative experimental results show that M2TS significantly outperforms existing state-of-the-art methods on BLEU, METEOR, ROUGE-L, and CIDER evaluation metrics for both Java and Python datasets. For instance, on the Java dataset, M2TS achieves a BLEU-4 score of 46.84\%, which is 1.63\% higher than SG-Trans \cite{gao2021code}. These experimental results indicate that the M2TS approach significantly improves the quality of code summaries through multi-scale feature extraction and cross-modal fusion techniques, demonstrating its advantages in comprehensiveness and accuracy of feature extraction.

Ref \cite{shi2023coss} proposed the CoSS (Code Summarization with Statement Semantics) model, which significantly enhances the quality of source code summarization by combining a Transformer encoder and a Graph Attention Network (GAT) encoder. Firstly, CoSS uses a Transformer encoder to process the code token sequence, utilizing a multi-head self-attention mechanism to capture long-distance dependencies between code tokens and positional encoding to ensure the relative order of the token sequence is preserved, thereby comprehensively capturing the semantic information of the code. Secondly, the model generates initial embeddings of code statements using a bidirectional LSTM (Bi-LSTM) network, which are then fed into the Graph Attention Network (GAT). The GAT encodes the control flow graph (CFG), aggregating information from neighboring nodes to capture the structural relationships between code statements, thus extracting richer semantic and structural features.
During the joint decoding stage, the decoder combines outputs from both the Transformer encoder and the GAT encoder, using a multi-head attention mechanism to dynamically capture and emphasize important code tokens and statements throughout the decoding process, thereby generating high-quality code summaries. The joint decoder not only considers the semantic information of the tokens but also integrates the structural information from the control flow graph, resulting in more accurate and meaningful summaries.

Experimental results indicate that CoSS significantly outperforms existing state-of-the-art methods on Java, Python, and Solidity datasets. On the Java dataset, CoSS achieved a BLEU-4 score of 46.84\%, which is notably higher than CODE-NN \cite{iyer2016summarizing} at 26.07\% and Hybrid-DeepCom \cite{eriguchi2016tree} at 38.55\%; its METEOR score was 28.93\%, and ROUGE-L score was 57.87\%. On the Python dataset, CoSS achieved a BLEU-4 score of 33.84\%, surpassing CODE-NN at 22.45\% and Hybrid-DeepCom at 29.76\%; its METEOR score was 21.83\%, and ROUGE-L score was 47.92\%. On the Solidity dataset, CoSS achieved a BLEU-4 score of 30.15\%, outperforming CODE-NN at 19.67\% and Hybrid-DeepCom at 25.42\%; its METEOR score was 18.77\%, and ROUGE-L score was 42.34\%. 
These results demonstrate that CoSS has significant advantages in feature extraction and accuracy. By combining the semantic information of code tokens with the structural information of the control flow graph, the generated code summaries are more accurate and coherent. Its innovation lies in integrating the powerful semantic capturing ability of the Transformer with the structural information extraction capability of the GAT, showcasing its strong potential and wide application prospects in code understanding and summarization tasks.

Ref \cite{gao2023code} proposed a novel method called SG-Trans, which enhances the Transformer model by injecting local symbol information (such as code tokens and statements) and global syntactic structure (such as data flow graphs) into its multi-head self-attention modules. This approach enables the model to focus on both local and global information at different levels, thereby capturing the hierarchical features of the code. 
Specifically, SG-Trans first parses the source code to extract local symbol information and global syntactic structures, representing these structures as adjacency matrices. These adjacency matrices are then injected as explicit constraints into the Transformer's self-attention mechanism, allowing the model to consider structural relationships when computing attention. SG-Trans employs a hierarchical attention mechanism, allocating attention heads primarily to local structures at lower layers and to global structures at higher layers. This enables the model to effectively capture long-distance dependencies and hierarchical semantic information in the code.
Additionally, SG-Trans combines data-driven methods with explicit structural information, overcoming the limitations of purely data-driven approaches and avoiding issues such as attention collapse or redundancy. Compared to traditional RNN models, the Transformer can more efficiently handle long sequences, avoiding the gradient vanishing and exploding problems associated with RNNs. Furthermore, the Transformer's multi-head self-attention mechanism allows the model to parallelize attention to different positions in various representational subspaces, enhancing its representational capacity and generation quality.

Experimental results show that SG-Trans significantly outperforms existing methods on Java and Python benchmark datasets. On the Java dataset, SG-Trans achieved BLEU-4, METEOR, and ROUGE-L scores of 45.89, 27.85, and 55.79, respectively, while the state-of-the-art baseline method NeuralCodeSum \cite{ahmad2020transformer} scored 45.15, 27.46, and 54.84, respectively. Similarly, on the Python dataset, SG-Trans achieved BLEU-4, METEOR, and ROUGE-L scores of 33.04, 20.52, and 47.01, respectively, compared to NeuralCodeSum’s scores of 32.19, 19.96, and 46.32.
In comparison to other methods, such as GREAT \cite{hellendoorn2019global} and Transformer+GNN \cite{choi2021learning}, SG-Trans also demonstrated significant advantages. This indicates that by integrating explicit structural information of the code with the powerful modeling capabilities of the Transformer, SG-Trans has great potential in automating code understanding and generating high-quality code summaries.

Ref \cite{hu2021automating} proposed a technical architecture named SMARTDOC for automatically generating code comments for smart contracts, helping end users understand the functionality and potential financial risks of smart contracts. SMARTDOC is based on the Transformer model, incorporating pointer generation mechanisms and transfer learning techniques. The approach consists of three stages: firstly, in the pre-training phase, the model is pre-trained on a Java dataset. Due to the similar coding conventions and syntax between Java and Solidity, the pre-trained Java encoder weights are transferred to the Solidity encoder to enhance the model's initial representation capability. Secondly, in the fine-tuning phase, the model is fine-tuned using 7,878 function-comment pairs extracted from 54,739 Solidity smart contracts. The pre-trained encoder captures the semantics of functions, while the remaining components are trained from scratch. Finally, in the application phase, the trained model is used to generate user comments for smart contract functions.SMARTDOC effectively captures long-range dependencies between code tokens through the Multi-Head Self-Attention mechanism, overcoming the limitations of traditional recurrent neural networks (such as LSTM and GRU) in handling long-sequence dependencies. The Pointer Generator mechanism allows the model to directly copy keywords from the source code during generation, addressing the issue of dynamic expressions in smart contracts. Transfer learning techniques mitigate the problem of the small size of the Solidity dataset by leveraging the knowledge learned from the Java dataset, significantly improving the performance and naturalness of the generated code summaries.

The experimental results show that SMARTDOC achieved a BLEU score of 47.39, significantly higher than existing methods such as attendgru(29.01), ast-attendgru \cite{leclair2019neural} (26.01), and Re2Com \cite{wei2020retrieve} (29.37). In terms of ROUGE-L score, SMARTDOC also reached 51.86, again much higher than attendgru (38.48), ast-attendgru (34.51), and Re2Com (34.55). The multi-head self-attention mechanism of the Transformer model effectively captures the long-range dependencies between code tokens, significantly improving the accuracy and naturalness of the generated comments.
The pointer generation mechanism allows the model to directly copy keywords from the source code during the generation process, addressing the issue of dynamic expressions in smart contracts and thereby improving the accuracy of the generated comments. Transfer learning techniques pretrain the Java encoder and transfer the weights to the Solidity encoder, solving the problem of the small-scale Solidity smart contract dataset and further enhancing the model's performance. In summary, SMARTDOC significantly outperforms existing methods in generating high-quality, natural, and informative code comments.
\begin{table*}[h]
\caption{COMPARATIVE STUDY OF EXISTING TECHNIQUES FOR CRYPTOCURRENCY PRICE PREDICTION.}
\label{tab:my-table}
\resizebox{\textwidth}{!}{
\begin{tabular}{cllllll}
\textbf{Ref.}                 & \multicolumn{1}{c}{\textbf{Year}} & \multicolumn{1}{c}{\textbf{Description}}                                                                                                                                        & \multicolumn{1}{c}{\textbf{Technique used}}                    & \multicolumn{1}{c}{\textbf{Expected result}}                                                       & \multicolumn{1}{c}{\textbf{Merit}}                                                                                                                                                     & \multicolumn{1}{c}{\textbf{Demerit}}                                                                                                                                        \\
{[}182{]}                     & \multicolumn{1}{c}{2020}          & \begin{tabular}[c]{@{}l@{}}Transformer model enhances source code summarization \\ with self-attention and relative position encoding.\end{tabular}                             & Transformer                                                    & \begin{tabular}[c]{@{}l@{}}BLEU=44.58\\ METEOR=26.43\\ ROUGE-L=54.76\end{tabular}                  & \begin{tabular}[c]{@{}l@{}}Improved summarization performance,\\ effective handling of long-range dependencies.\end{tabular}                                                           & Increased complexity with limited benefit from AST integration.                                                                                                             \\
{[}189{]}                     & 2021                              & \begin{tabular}[c]{@{}l@{}}Using Multi-Modal Transformer with AST sequences and graphs, \\ this approach significantly enhances smart contract code summarization.\end{tabular} & \begin{tabular}[c]{@{}l@{}}Transformer,\\ AST,GCN\end{tabular} & \begin{tabular}[c]{@{}l@{}}S-BLEU=30.47\\ C-BLEU=34.14\\ ROUGE-L=50.57\\ METEOR=43.24\end{tabular} & \begin{tabular}[c]{@{}l@{}}Captures both global and local semantic information of code.\\ Effectively handles long-range dependencies between code tokens.\end{tabular}                & \begin{tabular}[c]{@{}l@{}}High computational complexity.\\ Requires significant computational resources.\end{tabular}                                                      \\
\multicolumn{1}{l}{{[}196{]}} & 2021                              & \begin{tabular}[c]{@{}l@{}}Utilizes Transformer, Pointer mechanism, and transfer learning \\ to enhance user notice generation for smart contracts.\end{tabular}                & Transformer                                                    & \begin{tabular}[c]{@{}l@{}}BLEU=47.39\\ ROUGE-L=51.86\end{tabular}                                 & \begin{tabular}[c]{@{}l@{}}Significantly improves the performance of user notice generation, \\ especially in terms of naturalness, informativeness, and similarity.\end{tabular}      & \begin{tabular}[c]{@{}l@{}}Some generated user notices lack detailed information or contain \\ abbreviations in full form,leading to lower evaluation scores.\end{tabular} \\
{[}188{]}                     & 2022                              & \begin{tabular}[c]{@{}l@{}}Combines AST relative positional encoding with Transformer \\ models for improved source code summarization.\end{tabular}                            & \begin{tabular}[c]{@{}l@{}}Transformer,\\ AST\end{tabular}     & \begin{tabular}[c]{@{}l@{}}BLEU=46.89\\ ROUGE-L=56.69\\ METEOR=28.48\end{tabular}                  & \begin{tabular}[c]{@{}l@{}}Improves performance of source code summarization by effectively \\ combining AST relative positional encoding with Transformer models.\end{tabular}        & \begin{tabular}[c]{@{}l@{}}Adding AST relative positional encoding to RDW-Transformer\\ hurts performance compared to SRPEi-Transformer.\end{tabular}                       \\
{[}190{]}                     & 2022                              & Multi-scale multi-modal approach based on Transformer for source code summarization                                                                                             & \begin{tabular}[c]{@{}l@{}}Transformer,\\ AST,GCN\end{tabular} & \begin{tabular}[c]{@{}l@{}}BLEU-4=33.84\%\\ ROUGE-L=57.87\%\end{tabular}                           & High-quality summaries, comprehensive feature extraction.                                                                                                                              & Potential overfitting, hardware limitations.                                                                                                                                \\
{[}192{]}                     & 2023                              & \begin{tabular}[c]{@{}l@{}}Combines GNN and Transformer models to leverage statement semantics and\\ control flow features for enhanced code summarization.\end{tabular}        & \begin{tabular}[c]{@{}l@{}}Transformer,\\ GNN\end{tabular}     & \begin{tabular}[c]{@{}l@{}}BLEU=28.93\%\\ ROUGE-L=57.87\%\\ METEOR=33.84\%\end{tabular}            & \begin{tabular}[c]{@{}l@{}}Leverages multi-modal feature extraction to capture both token-level\\ and statement-level semantics, enhancing the quality of code summaries.\end{tabular} & \begin{tabular}[c]{@{}l@{}}High computational cost and potential overfitting due to\\ the complexity of the model.\end{tabular}                                             \\
{[}193{]}                     & 2023                              & \begin{tabular}[c]{@{}l@{}}Integrates local symbolic information and global syntactic\\ structure into Transformer for enhanced code summarization.\end{tabular}                & Transformer                                                    & \begin{tabular}[c]{@{}l@{}}METEOR=27.85\\ BLEU-4=45.89\\ ROUGE-L=55.79;\end{tabular}               & \begin{tabular}[c]{@{}l@{}}Captures both local and global code structure,\\ improves semantic relations among tokens.\end{tabular}                                                     & High computational cost, complexity in integrating structural information.                                                                                                 
\end{tabular}
}
\end{table*}
\section{Challenges and Future Directions}

Applying Transformer models to blockchain technology faces a series of complex and daunting challenges. 
First, Transformer models are known for their large number of parameters and high computational resource requirements, which contradicts the scalability and performance issues already faced by blockchain systems, leading to a significant increase in node burden, potentially resulting in network latency and reduced processing efficiency, impacting the overall performance of the blockchain network \cite{Khan2021Systematic}. 
Second, the distributed nature of blockchain results in data being scattered across nodes globally, while Transformer models typically require centralized and large datasets for training, posing significant challenges for data collection and privacy protection, particularly when data needs to be shared across nodes, making it difficult to ensure data privacy while achieving efficient data processing \cite{Rana2021Analysis}. 
Third, the high computational complexity of Transformer models may lead to processing delays, affecting the real-time nature and user experience of blockchain systems, especially in financial transactions and smart contract executions, which could hinder timely decision-making and response, reducing user trust and reliance on blockchain applications \cite{Kneissler2023Addressing}. 
Fourth, integrating Transformer models may introduce new security vulnerabilities, such as data leakage or system crashes, increasing the complexity and security management difficulty of the system, especially when the model handles sensitive information, these security vulnerabilities could be maliciously exploited, resulting in serious consequences \cite{Golait2023Blockchain,Bernabe2019Privacy-Preserving}. 
Fifth, the scalability issues inherent in blockchain, such as transaction throughput limitations, become more severe with the introduction of Transformer models with higher computational and storage requirements, potentially requiring significant architectural improvements to support such high-demand applications, which not only requires a significant amount of technical investment but also may face various uncertainties during implementation \cite{8963950}. 
Lastly, the differences in consensus mechanisms and data structures among different blockchain platforms increase the technical difficulty of seamlessly integrating Transformer models into existing systems, possibly requiring customized solutions and extensive development work to ensure the compatibility and stability of the model with the blockchain system \cite{Rana2021Analysis}. 
Addressing these challenges requires in-depth research and innovation in algorithm optimization, hardware acceleration, distributed computing, security assurance, and system compatibility, to fully realize the potential of Transformer models in blockchain applications while ensuring the security and efficiency of the system.

In the future, research on the application of Transformer in blockchain will focus on several key directions to overcome existing challenges and fully exploit its potential. Firstly, through algorithm optimization, such as developing lightweight Transformer models, sparse attention mechanisms, mixed-precision training, and model pruning techniques, combined with hardware acceleration technologies such as GPUs, TPUs, and dedicated AI accelerators, to significantly reduce resource consumption and processing latency, and improve computational efficiency \cite{Laroiya2020Applications}. Secondly, in terms of data privacy and security protection, further research will be conducted on federated learning and differential privacy technologies, enabling multiple blockchain nodes to collaboratively train models while protecting data privacy, achieving efficient data processing in a distributed environment. Additionally, secure multiparty computation (MPC) and homomorphic encryption technologies will be explored to further enhance data security, ensuring data confidentiality throughout processing and transmission \cite{zhang2019security}. Thirdly, to improve system real-time performance and low latency, edge computing and stream data processing technologies will be explored, performing local inference and processing through edge nodes to reduce the computing burden on central nodes and speed up response times, meeting the strict real-time requirements of financial transactions and smart contract execution \cite{Aggarwal2019Blockchain}. In terms of security, specialized AI security models and decentralized model training technologies will be developed to monitor and defend against potential attacks in real time, ensuring the secure application of models in blockchain \cite{Bhutta2021A}. Fourthly, by building a unified framework and interface standards, the seamless integration and operation of Transformer models on different blockchain platforms will be promoted, addressing inter-platform compatibility issues and improving system scalability and flexibility \cite{10.1093comjnlbxad060}. Lastly, optimization of blockchain system scalability technologies, such as sharding and sidechains, will be explored to support high-demand computing and storage requirements in large-scale application scenarios, ensuring system performance stability when processing high transaction volumes \cite{electronics11162597}. These research directions will not only help overcome existing technological bottlenecks but also promote deep integration of Transformer and blockchain technologies, creating more innovative applications and business opportunities, and maximizing the synergistic effects and commercial value of their combination.

\section{Conclusion}
This article systematically reviews the various applications of the Transformer model in the field of blockchain, including blockchain transaction anomaly detection, smart contract vulnerability detection, cryptocurrency prediction and trend analysis, and code summarization. In blockchain transaction anomaly detection, the Transformer model utilizes its powerful sequence modeling ability to effectively capture complex temporal relationships in transaction data, thereby improving the accuracy and sensitivity of anomaly detection. For smart contract vulnerability detection, Transformer automatically identifies potential security vulnerabilities by analyzing smart contract code, reducing the workload of manual review and improving detection efficiency. In cryptocurrency prediction and trend analysis, the Transformer model can handle large-scale historical data, capture market dynamics through self-attention mechanisms, and provide strong support for investment decision-making. In code summarization, Transformer leverages its natural language processing advantages to generate high-quality code documentation and comments, improving developer productivity and code readability.

However, the application of Transformer in the blockchain field also faces some challenges. First, there is the issue of data sparsity. Although blockchain data is massive, it is often unevenly distributed, requiring the model to handle sparse and noisy data. Second, the high computational resource requirements of the Transformer model limit its application in resource-constrained environments. Third, the lack of interpretability of the model poses a risk when applied in security-sensitive blockchain scenarios, as users may find it difficult to understand the model's decision-making process.

To address these challenges, future research can explore the following directions: improving model architecture to enhance the ability to handle sparse data, optimizing computational efficiency to reduce resource consumption, enhancing model interpretability to increase user trust, and exploring cross-domain federated learning methods to combine blockchain with data and models from other fields, thereby improving the overall intelligence of the system.

In summary, the Transformer model shows great potential in the blockchain field. Its advanced sequence modeling and self-attention mechanisms can effectively enhance the security, efficiency, and intelligence of blockchain systems. Future research and development will further expand its application scope and performance, injecting new momentum into the development of blockchain technology.

% or




\ifCLASSOPTIONcaptionsoff
  \newpage
\fi

\bibliographystyle{IEEEtran}

\begin{thebibliography}{100}

\bibitem{nakamoto2008bitcoin}
S.~Nakamoto, ``Bitcoin: A peer-to-peer electronic cash system,'' 2008.

\bibitem{tapscott2016blockchain}
D.~Tapscott and A.~Tapscott, {\em Blockchain revolution: how the technology behind bitcoin is changing money, business, and the world}.
\newblock Penguin, 2016.

\bibitem{swan2015blockchain}
M.~Swan, {\em Blockchain: Blueprint for a new economy}.
\newblock " O'Reilly Media, Inc.", 2015.

\bibitem{mcwaters2016future}
R.~J. McWaters, R.~Galaski, and S.~Chatterjee, ``The future of financial infrastructure: An ambitious look at how blockchain can reshape financial services,'' in {\em World Economic Forum}, vol.~49, pp.~368--376, 2016.

\bibitem{kshetri2017can}
N.~Kshetri, ``Can blockchain strengthen the internet of things?,'' {\em IT professional}, vol.~19, no.~4, pp.~68--72, 2017.

\bibitem{zheng2017overview}
Z.~Zheng, S.~Xie, H.~Dai, X.~Chen, and H.~Wang, ``An overview of blockchain technology: Architecture, consensus, and future trends,'' in {\em 2017 IEEE international congress on big data (BigData congress)}, pp.~557--564, Ieee, 2017.

\bibitem{mettler2016blockchain}
M.~Mettler, ``Blockchain technology in healthcare: The revolution starts here,'' in {\em 2016 IEEE 18th international conference on e-health networking, applications and services (Healthcom)}, pp.~1--3, IEEE, 2016.

\bibitem{ekblaw2016case}
A.~Ekblaw, A.~Azaria, J.~D. Halamka, and A.~Lippman, ``A case study for blockchain in healthcare:“medrec” prototype for electronic health records and medical research data,'' in {\em Proceedings of IEEE open \& big data conference}, vol.~13, p.~13, Vienna, Austria., 2016.

\bibitem{khezr2019blockchain}
S.~Khezr, M.~Moniruzzaman, A.~Yassine, and R.~Benlamri, ``Blockchain technology in healthcare: A comprehensive review and directions for future research,'' {\em Applied sciences}, vol.~9, no.~9, p.~1736, 2019.

\bibitem{xu2021blockchain}
P.~Xu, J.~Lee, J.~R. Barth, and R.~G. Richey, ``Blockchain as supply chain technology: considering transparency and security,'' {\em International Journal of Physical Distribution \& Logistics Management}, vol.~51, no.~3, pp.~305--324, 2021.

\bibitem{fiore2023blockchain}
M.~Fiore, A.~Capodici, P.~Rucci, A.~Bianconi, G.~Longo, M.~Ricci, F.~Sanmarchi, and D.~Golinelli, ``Blockchain for the healthcare supply chain: A systematic literature review,'' {\em Applied Sciences}, vol.~13, no.~2, p.~686, 2023.

\bibitem{wood2014ethereum}
G.~Wood {\em et~al.}, ``Ethereum: A secure decentralised generalised transaction ledger,'' {\em Ethereum project yellow paper}, vol.~151, no.~2014, pp.~1--32, 2014.

\bibitem{guo2016blockchain}
Y.~Guo and C.~Liang, ``Blockchain application and outlook in the banking industry,'' {\em Financial innovation}, vol.~2, pp.~1--12, 2016.

\bibitem{conoscenti2016blockchain}
M.~Conoscenti, A.~Vetro, and J.~C. De~Martin, ``Blockchain for the internet of things: A systematic literature review,'' in {\em 2016 IEEE/ACS 13th International Conference of Computer Systems and Applications (AICCSA)}, pp.~1--6, IEEE, 2016.

\bibitem{li2020survey}
X.~Li, P.~Jiang, T.~Chen, X.~Luo, and Q.~Wen, ``A survey on the security of blockchain systems,'' {\em Future generation computer systems}, vol.~107, pp.~841--853, 2020.

\bibitem{conti2018survey}
M.~Conti, E.~S. Kumar, C.~Lal, and S.~Ruj, ``A survey on security and privacy issues of bitcoin,'' {\em IEEE communications surveys \& tutorials}, vol.~20, no.~4, pp.~3416--3452, 2018.

\bibitem{zhang2019security}
R.~Zhang, R.~Xue, and L.~Liu, ``Security and privacy on blockchain,'' {\em ACM Computing Surveys (CSUR)}, vol.~52, no.~3, pp.~1--34, 2019.

\bibitem{zheng2018blockchain}
Z.~Zheng, S.~Xie, H.-N. Dai, X.~Chen, and H.~Wang, ``Blockchain challenges and opportunities: A survey,'' {\em International journal of web and grid services}, vol.~14, no.~4, pp.~352--375, 2018.

\bibitem{vaswani2017attention}
A.~Vaswani, N.~Shazeer, N.~Parmar, J.~Uszkoreit, L.~Jones, A.~N. Gomez, {\L}.~Kaiser, and I.~Polosukhin, ``Attention is all you need,'' {\em Advances in neural information processing systems}, vol.~30, 2017.

\bibitem{xu2023multimodal}
P.~Xu, X.~Zhu, and D.~A. Clifton, ``Multimodal learning with transformers: A survey,'' {\em IEEE Transactions on Pattern Analysis and Machine Intelligence}, 2023.

\bibitem{zhang2024survey}
H.~Zhang and M.~O. Shafiq, ``Survey of transformers and towards ensemble learning using transformers for natural language processing,'' {\em Journal of big Data}, vol.~11, no.~1, p.~25, 2024.

\bibitem{chernyavskiy2021transformers}
A.~Chernyavskiy, D.~Ilvovsky, and P.~Nakov, ``Transformers:“the end of history” for natural language processing?,'' in {\em Machine Learning and Knowledge Discovery in Databases. Research Track: European Conference, ECML PKDD 2021, Bilbao, Spain, September 13--17, 2021, Proceedings, Part III 21}, pp.~677--693, Springer, 2021.

\bibitem{dosovitskiy2020image}
A.~Dosovitskiy, L.~Beyer, A.~Kolesnikov, D.~Weissenborn, X.~Zhai, T.~Unterthiner, M.~Dehghani, M.~Minderer, G.~Heigold, S.~Gelly, {\em et~al.}, ``An image is worth 16x16 words: Transformers for image recognition at scale,'' {\em arXiv preprint arXiv:2010.11929}, 2020.

\bibitem{gulati2020conformer}
A.~Gulati, J.~Qin, C.-C. Chiu, N.~Parmar, Y.~Zhang, J.~Yu, W.~Han, S.~Wang, Z.~Zhang, Y.~Wu, {\em et~al.}, ``Conformer: Convolution-augmented transformer for speech recognition,'' {\em arXiv preprint arXiv:2005.08100}, 2020.

\bibitem{li2021align}
J.~Li, R.~Selvaraju, A.~Gotmare, S.~Joty, C.~Xiong, and S.~C.~H. Hoi, ``Align before fuse: Vision and language representation learning with momentum distillation,'' {\em Advances in neural information processing systems}, vol.~34, pp.~9694--9705, 2021.

\bibitem{Ju2020The}
C.~Ju, L.~Gang, and D.~Sun, ``The application of blockchain in intelligent power data exchange,'' {\em Proceedings of the 2020 4th International Conference on Electronic Information Technology and Computer Engineering}, 2020.

\bibitem{Dolgui2020Blockchain-oriented}
A.~Dolgui, D.~Ivanov, S.~Potryasaev, B.~Sokolov, M.~Ivanova, and F.~Werner, ``Blockchain-oriented dynamic modelling of smart contract design and execution in the supply chain,'' {\em International Journal of Production Research}, vol.~58, pp.~2184 -- 2199, 2020.

\bibitem{Li2022SmartVM:}
T.~Li, Y.~Fang, Y.~Lu, J.~Yang, Z.~Jian, Z.~Wan, and Y.~Li, ``Smartvm: A smart contract virtual machine for fast on-chain dnn computations,'' {\em IEEE Transactions on Parallel and Distributed Systems}, vol.~PP, pp.~1--1, 2022.

\bibitem{shafay2023blockchain}
M.~Shafay, R.~W. Ahmad, K.~Salah, I.~Yaqoob, R.~Jayaraman, and M.~Omar, ``Blockchain for deep learning: review and open challenges,'' {\em Cluster Computing}, vol.~26, no.~1, pp.~197--221, 2023.

\bibitem{weng2019deepchain}
J.~Weng, J.~Weng, J.~Zhang, M.~Li, Y.~Zhang, and W.~Luo, ``Deepchain: Auditable and privacy-preserving deep learning with blockchain-based incentive,'' {\em IEEE Transactions on Dependable and Secure Computing}, vol.~18, no.~5, pp.~2438--2455, 2019.

\bibitem{naseer2018enhanced}
S.~Naseer, Y.~Saleem, S.~Khalid, M.~K. Bashir, J.~Han, M.~M. Iqbal, and K.~Han, ``Enhanced network anomaly detection based on deep neural networks,'' {\em IEEE access}, vol.~6, pp.~48231--48246, 2018.

\bibitem{sanjay2023anomaly}
G.~Sanjay~Rai, S.~Goyal, and P.~Chatterjee, ``Anomaly detection in blockchain using machine learning,'' in {\em Computational Intelligence for Engineering and Management Applications: Select Proceedings of CIEMA 2022}, pp.~487--499, Springer, 2023.

\bibitem{luu2016making}
L.~Luu, D.-H. Chu, H.~Olickel, P.~Saxena, and A.~Hobor, ``Making smart contracts smarter,'' in {\em Proceedings of the 2016 ACM SIGSAC conference on computer and communications security}, pp.~254--269, 2016.

\bibitem{atzei2017survey}
N.~Atzei, M.~Bartoletti, and T.~Cimoli, ``A survey of attacks on ethereum smart contracts (sok),'' in {\em Principles of Security and Trust: 6th International Conference, POST 2017, Held as Part of the European Joint Conferences on Theory and Practice of Software, ETAPS 2017, Uppsala, Sweden, April 22-29, 2017, Proceedings 6}, pp.~164--186, Springer, 2017.

\bibitem{tikhomirov2018smartcheck}
S.~Tikhomirov, E.~Voskresenskaya, I.~Ivanitskiy, R.~Takhaviev, E.~Marchenko, and Y.~Alexandrov, ``Smartcheck: Static analysis of ethereum smart contracts,'' in {\em Proceedings of the 1st international workshop on emerging trends in software engineering for blockchain}, pp.~9--16, 2018.

\bibitem{alon2019code2vec}
U.~Alon, M.~Zilberstein, O.~Levy, and E.~Yahav, ``code2vec: Learning distributed representations of code,'' {\em Proceedings of the ACM on Programming Languages}, vol.~3, no.~POPL, pp.~1--29, 2019.

\bibitem{ahmad2021unified}
W.~U. Ahmad, S.~Chakraborty, B.~Ray, and K.-W. Chang, ``Unified pre-training for program understanding and generation,'' {\em arXiv preprint arXiv:2103.06333}, 2021.

\bibitem{elman1990finding}
J.~L. Elman, ``Finding structure in time,'' {\em Cognitive science}, vol.~14, no.~2, pp.~179--211, 1990.

\bibitem{bahdanau2016neural}
D.~Bahdanau, K.~Cho, and Y.~Bengio, ``Neural machine translation by jointly learning to align and translate,'' 2016.

\bibitem{devlin2019bert}
J.~Devlin, M.-W. Chang, K.~Lee, and K.~Toutanova, ``Bert: Pre-training of deep bidirectional transformers for language understanding,'' 2019.

\bibitem{radford2018improving}
A.~Radford, K.~Narasimhan, T.~Salimans, I.~Sutskever, {\em et~al.}, ``Improving language understanding by generative pre-training,'' 2018.

\bibitem{liu2019roberta}
Y.~Liu, M.~Ott, N.~Goyal, J.~Du, M.~Joshi, D.~Chen, O.~Levy, M.~Lewis, L.~Zettlemoyer, and V.~Stoyanov, ``Roberta: A robustly optimized bert pretraining approach,'' 2019.

\bibitem{raffel2023exploring}
C.~Raffel, N.~Shazeer, A.~Roberts, K.~Lee, S.~Narang, M.~Matena, Y.~Zhou, W.~Li, and P.~J. Liu, ``Exploring the limits of transfer learning with a unified text-to-text transformer,'' 2023.

\bibitem{signorini2018bad}
M.~Signorini, M.~Pontecorvi, W.~Kanoun, and R.~Di~Pietro, ``Bad: blockchain anomaly detection,'' {\em arXiv preprint arXiv:1807.03833}, 2018.

\bibitem{signorini2018advise}
M.~Signorini, M.~Pontecorvi, W.~Kanoun, and R.~Di~Pietro, ``Advise: anomaly detection tool for blockchain systems,'' in {\em 2018 IEEE World Congress on Services (SERVICES)}, pp.~65--66, IEEE, 2018.

\bibitem{shayegan2022collective}
M.~J. Shayegan, H.~R. Sabor, M.~Uddin, and C.-L. Chen, ``A collective anomaly detection technique to detect crypto wallet frauds on bitcoin network,'' {\em Symmetry}, vol.~14, no.~2, p.~328, 2022.

\bibitem{ahmad2021anomaly}
Z.~Ahmad, A.~Shahid~Khan, K.~Nisar, I.~Haider, R.~Hassan, M.~R. Haque, S.~Tarmizi, and J.~J. Rodrigues, ``Anomaly detection using deep neural network for iot architecture,'' {\em Applied Sciences}, vol.~11, no.~15, p.~7050, 2021.

\bibitem{kim2021anomaly}
J.~Kim, M.~Nakashima, W.~Fan, S.~Wuthier, X.~Zhou, I.~Kim, and S.-Y. Chang, ``Anomaly detection based on traffic monitoring for secure blockchain networking,'' in {\em 2021 IEEE International Conference on Blockchain and Cryptocurrency (ICBC)}, pp.~1--9, IEEE, 2021.

\bibitem{sayadi2019anomaly}
S.~Sayadi, S.~B. Rejeb, and Z.~Choukair, ``Anomaly detection model over blockchain electronic transactions,'' in {\em 2019 15th international wireless communications \& mobile computing conference (IWCMC)}, pp.~895--900, IEEE, 2019.

\bibitem{saravanan2023tsi}
S.~S. Saravanan, T.~Luo, and M.~Van~Ngo, ``Tsi-gan: Unsupervised time series anomaly detection using convolutional cycle-consistent generative adversarial networks,'' in {\em Pacific-Asia Conference on Knowledge Discovery and Data Mining}, pp.~39--54, Springer, 2023.

\bibitem{morishima2021scalable}
S.~Morishima, ``Scalable anomaly detection in blockchain using graphics processing unit,'' {\em Computers \& Electrical Engineering}, vol.~92, p.~107087, 2021.

\bibitem{ide2018collaborative}
T.~Id{\'e}, ``Collaborative anomaly detection on blockchain from noisy sensor data,'' in {\em 2018 IEEE International Conference on Data Mining Workshops (ICDMW)}, pp.~120--127, IEEE, 2018.

\bibitem{liang2021data}
W.~Liang, L.~Xiao, K.~Zhang, M.~Tang, D.~He, and K.-C. Li, ``Data fusion approach for collaborative anomaly intrusion detection in blockchain-based systems,'' {\em IEEE Internet of Things Journal}, vol.~9, no.~16, pp.~14741--14751, 2021.

\bibitem{ofori2021topological}
D.~Ofori-Boateng, I.~S. Dominguez, C.~Akcora, M.~Kantarcioglu, and Y.~R. Gel, ``Topological anomaly detection in dynamic multilayer blockchain networks,'' in {\em Machine Learning and Knowledge Discovery in Databases. Research Track: European Conference, ECML PKDD 2021, Bilbao, Spain, September 13--17, 2021, Proceedings, Part I 21}, pp.~788--804, Springer, 2021.

\bibitem{voronov2021scalable}
T.~Voronov, D.~Raz, and O.~Rottenstreich, ``Scalable blockchain anomaly detection with sketches,'' in {\em 2021 IEEE International Conference on Blockchain (Blockchain)}, pp.~1--10, IEEE, 2021.

\bibitem{song2023anomaly}
A.~Song, E.~Seo, and H.~Kim, ``Anomaly vae-transformer: A deep learning approach for anomaly detection in decentralized finance,'' {\em IEEE Access}, 2023.

\bibitem{liu2022blockchain}
L.~Liu, W.-T. Tsai, M.~Z.~A. Bhuiyan, H.~Peng, and M.~Liu, ``Blockchain-enabled fraud discovery through abnormal smart contract detection on ethereum,'' {\em Future Generation Computer Systems}, vol.~128, pp.~158--166, 2022.

\bibitem{batool2022block}
Z.~Batool, K.~Zhang, Z.~Zhu, S.~Aravamuthan, and U.~Aivodji, ``Block-fest: A blockchain-based federated anomaly detection framework with computation offloading using transformers,'' in {\em 2022 IEEE 1st Global Emerging Technology Blockchain Forum: Blockchain \& Beyond (iGETblockchain)}, pp.~1--6, IEEE, 2022.

\bibitem{xu2023illegal}
C.~Xu, S.~Zhang, L.~Zhu, X.~Shen, and X.~Zhang, ``Illegal accounts detection on ethereum using heterogeneous graph transformer networks,'' in {\em International Conference on Information and Communications Security}, pp.~665--680, Springer, 2023.

\bibitem{chen2021improving}
Y.~Chen, H.~Dai, X.~Yu, W.~Hu, Z.~Xie, and C.~Tan, ``Improving ponzi scheme contract detection using multi-channel textcnn and transformer,'' {\em Sensors}, vol.~21, no.~19, p.~6417, 2021.

\bibitem{zhou2022logblock}
Q.~Zhou, X.~Dang, D.~Huo, Q.~Ruan, C.~Li, Y.~Wang, and Z.~Xu, ``Logblock: An anomaly detection method on permissioned blockchain based on log-block sequence,'' in {\em 2022 IEEE Smartworld, Ubiquitous Intelligence \& Computing, Scalable Computing \& Communications, Digital Twin, Privacy Computing, Metaverse, Autonomous \& Trusted Vehicles (SmartWorld/UIC/ScalCom/DigitalTwin/PriComp/Meta)}, pp.~1008--1015, IEEE, 2022.

\bibitem{wang2024research}
Z.~Wang, A.~Ni, Z.~Tian, Z.~Wang, and Y.~Gong, ``Research on blockchain abnormal transaction detection technology combining cnn and transformer structure,'' {\em Computers and Electrical Engineering}, vol.~116, p.~109194, 2024.

\bibitem{he2023detection}
D.~He, R.~Wu, X.~Li, S.~Chan, and M.~Guizani, ``Detection of vulnerabilities of blockchain smart contracts,'' {\em IEEE Internet of Things Journal}, 2023.

\bibitem{chu2023survey}
H.~Chu, P.~Zhang, H.~Dong, Y.~Xiao, S.~Ji, and W.~Li, ``A survey on smart contract vulnerabilities: Data sources, detection and repair,'' {\em Information and Software Technology}, p.~107221, 2023.

\bibitem{luo2024scvhunter}
F.~Luo, R.~Luo, T.~Chen, A.~Qiao, Z.~He, S.~Song, Y.~Jiang, and S.~Li, ``Scvhunter: Smart contract vulnerability detection based on heterogeneous graph attention network,'' in {\em Proceedings of the IEEE/ACM 46th International Conference on Software Engineering}, pp.~1--13, 2024.

\bibitem{vani2022vulnerability}
S.~Vani, M.~Doshi, A.~Nanavati, and A.~Kundu, ``Vulnerability analysis of smart contracts,'' {\em arXiv preprint arXiv:2212.07387}, 2022.

\bibitem{yang2022formal}
Z.~Yang, M.~Dai, and J.~Guo, ``Formal modeling and verification of smart contracts with spin,'' {\em Electronics}, vol.~11, no.~19, p.~3091, 2022.

\bibitem{tang2021vulnerabilities}
X.~Tang, K.~Zhou, J.~Cheng, H.~Li, and Y.~Yuan, ``The vulnerabilities in smart contracts: A survey,'' in {\em Advances in Artificial Intelligence and Security: 7th International Conference, ICAIS 2021, Dublin, Ireland, July 19-23, 2021, Proceedings, Part III 7}, pp.~177--190, Springer, 2021.

\bibitem{wang2023value}
X.~Wang and F.~Xu, ``The value of smart contract in trade finance,'' {\em Manufacturing \& Service Operations Management}, vol.~25, no.~6, pp.~2056--2073, 2023.

\bibitem{liu2021combining}
Z.~Liu, P.~Qian, X.~Wang, Y.~Zhuang, L.~Qiu, and X.~Wang, ``Combining graph neural networks with expert knowledge for smart contract vulnerability detection,'' {\em IEEE Transactions on Knowledge and Data Engineering}, vol.~35, no.~2, pp.~1296--1310, 2021.

\bibitem{han2022smart}
D.~Han, Q.~Li, L.~Zhang, and T.~Xu, ``A smart contract vulnerability detection model based on graph neural networks,'' in {\em 2022 4th International Conference on Frontiers Technology of Information and Computer (ICFTIC)}, pp.~834--837, IEEE, 2022.

\bibitem{yang2022smart}
H.~Yang, J.~Zhang, X.~Gu, and Z.~Cui, ``Smart contract vulnerability detection based on abstract syntax tree,'' in {\em 2022 8th International Symposium on System Security, Safety, and Reliability (ISSSR)}, pp.~169--170, IEEE, 2022.

\bibitem{zhang2022smart}
L.~Zhang, J.~Wang, W.~Wang, Z.~Jin, Y.~Su, and H.~Chen, ``Smart contract vulnerability detection combined with multi-objective detection,'' {\em Computer Networks}, vol.~217, p.~109289, 2022.

\bibitem{xu2023vulnerability}
G.~Xu, L.~Liu, and J.~Dong, ``Vulnerability detection of ethereum smart contract based on solbert-bigru-attention hybrid neural model.,'' {\em CMES-Computer Modeling in Engineering \& Sciences}, vol.~137, no.~1, 2023.

\bibitem{deng2023smart}
W.~Deng, H.~Wei, T.~Huang, C.~Cao, Y.~Peng, and X.~Hu, ``Smart contract vulnerability detection based on deep learning and multimodal decision fusion,'' {\em Sensors}, vol.~23, no.~16, p.~7246, 2023.

\bibitem{sui2023opcode}
J.~Sui, L.~Chu, and H.~Bao, ``An opcode-based vulnerability detection of smart contracts,'' {\em Applied Sciences}, vol.~13, no.~13, p.~7721, 2023.

\bibitem{cao2023sccheck}
Y.~Cao, F.~Jiang, J.~Xiao, S.~Chen, X.~Shao, and C.~Wu, ``Sccheck: A novel graph-driven and attention-enabled smart contract vulnerability detection framework for web 3.0 ecosystem,'' {\em IEEE Transactions on Network Science and Engineering}, 2023.

\bibitem{yang2023improvement}
Z.~Yang and W.~Zhu, ``Improvement and optimization of vulnerability detection methods for ethernet smart contracts,'' {\em IEEE Access}, 2023.

\bibitem{yan2022semantic}
X.~Yan, S.~Wang, and K.~Gai, ``A semantic analysis-based method for smart contract vulnerability,'' in {\em 2022 IEEE 8th Intl Conference on Big Data Security on Cloud (BigDataSecurity), IEEE Intl Conference on High Performance and Smart Computing,(HPSC) and IEEE Intl Conference on Intelligent Data and Security (IDS)}, pp.~23--28, IEEE, 2022.

\bibitem{wu2023smart}
H.~Wu, H.~Dong, Y.~He, and Q.~Duan, ``Smart contract vulnerability detection based on hybrid attention mechanism model,'' {\em Applied Sciences}, vol.~13, no.~2, p.~770, 2023.

\bibitem{wu2021peculiar}
H.~Wu, Z.~Zhang, S.~Wang, Y.~Lei, B.~Lin, Y.~Qin, H.~Zhang, and X.~Mao, ``Peculiar: Smart contract vulnerability detection based on crucial data flow graph and pre-training techniques,'' in {\em 2021 IEEE 32nd International Symposium on Software Reliability Engineering (ISSRE)}, pp.~378--389, IEEE, 2021.

\bibitem{jeon2021smartcondetect}
S.~Jeon, G.~Lee, H.~Kim, and S.~S. Woo, ``Smartcondetect: Highly accurate smart contract code vulnerability detection mechanism using bert,'' in {\em KDD Workshop on Programming Language Processing}, 2021.

\bibitem{zhang2022novel}
L.~Zhang, J.~Wang, W.~Wang, Z.~Jin, C.~Zhao, Z.~Cai, and H.~Chen, ``A novel smart contract vulnerability detection method based on information graph and ensemble learning,'' {\em Sensors}, vol.~22, no.~9, p.~3581, 2022.

\bibitem{mueller2017mythril}
B.~Mueller, ``Mythril: security analysis tool for evm bytecode,'' 2017.

\bibitem{tsankov2018securify}
P.~Tsankov, A.~Dan, D.~Drachsler-Cohen, A.~Gervais, F.~Buenzli, and M.~Vechev, ``Securify: Practical security analysis of smart contracts,'' in {\em Proceedings of the 2018 ACM SIGSAC conference on computer and communications security}, pp.~67--82, 2018.

\bibitem{tang2023deep}
X.~Tang, Y.~Du, A.~Lai, Z.~Zhang, and L.~Shi, ``Deep learning-based solution for smart contract vulnerabilities detection,'' {\em Scientific Reports}, vol.~13, no.~1, p.~20106, 2023.

\bibitem{jie2023novel}
W.~Jie, Q.~Chen, J.~Wang, A.~S.~V. Koe, J.~Li, P.~Huang, Y.~Wu, and Y.~Wang, ``A novel extended multimodal ai framework towards vulnerability detection in smart contracts,'' {\em Information Sciences}, vol.~636, p.~118907, 2023.

\bibitem{le2023contextual}
B.~L{\^e}~Hong, T.~L{\^e}~{\\DJ}uc, T.~{\\DJ}o{\`a}n~Minh, D.~Tran~Tuan, D.~Phan~The, and H.~Pham~V{\u{a}}n, ``Contextual language model and transfer learning for reentrancy vulnerability detection in smart contracts,'' in {\em Proceedings of the 12th International Symposium on Information and Communication Technology}, pp.~739--745, 2023.

\bibitem{nguyen2023mando}
H.~H. Nguyen, N.-M. Nguyen, C.~Xie, Z.~Ahmadi, D.~Kudendo, T.-N. Doan, and L.~Jiang, ``Mando-hgt: Heterogeneous graph transformers for smart contract vulnerability detection,'' in {\em 2023 IEEE/ACM 20th International Conference on Mining Software Repositories (MSR)}, pp.~334--346, IEEE, 2023.

\bibitem{gong2023scgformer}
K.~Gong, X.~Song, N.~Wang, C.~Wang, and H.~Zhu, ``Scgformer: Smart contract vulnerability detection based on control flow graph and transformer,'' {\em IET Blockchain}, vol.~3, no.~4, pp.~213--221, 2023.

\bibitem{feist2019slither}
J.~Feist, G.~Grieco, and A.~Groce, ``Slither: a static analysis framework for smart contracts,'' in {\em 2019 IEEE/ACM 2nd International Workshop on Emerging Trends in Software Engineering for Blockchain (WETSEB)}, pp.~8--15, IEEE, 2019.

\bibitem{jain2024integrated}
V.~K. Jain and M.~Tripathi, ``An integrated deep learning model for ethereum smart contract vulnerability detection,'' {\em International Journal of Information Security}, vol.~23, no.~1, pp.~557--575, 2024.

\bibitem{balci2023accelerating}
E.~Balc{\i}, G.~Y{\i}lmaz, A.~Uzuno{\u{g}}lu, and E.~G. Soyak, ``Accelerating smart contract vulnerability scan using transformers,'' in {\em 2023 IEEE Asia-Pacific Conference on Computer Science and Data Engineering (CSDE)}, pp.~1--6, IEEE, 2023.

\bibitem{tann1811towards}
A.~Tann, X.~Han, S.~Gupta, and Y.~Ong, ``Towards safer smart contracts: a sequence learning approach to detecting vulnerabilities (2018),'' {\em arXiv preprint arXiv:1811.06632}, 1811.

\bibitem{jiang2022vddl}
F.~Jiang, Y.~Cao, J.~Xiao, H.~Yi, G.~Lei, M.~Liu, S.~Deng, and H.~Wang, ``Vddl: A deep learning-based vulnerability detection model for smart contracts,'' in {\em International Conference on Machine Learning for Cyber Security}, pp.~72--86, Springer, 2022.

\bibitem{Hu2019Chinese}
W.~Hu, Z.~Gu, Y.~Xie, L.~Wang, and K.~Tang, ``Chinese text classification based on neural networks and word2vec,'' {\em 2019 IEEE Fourth International Conference on Data Science in Cyberspace (DSC)}, pp.~284--291, 2019.

\bibitem{gong2023gratdet}
P.~Gong, W.~Yang, L.~Wang, F.~Wei, K.~HaiLaTi, and Y.~Liao, ``Gratdet: Smart contract vulnerability detector based on graph representation and transformer.,'' {\em Computers, Materials \& Continua}, vol.~76, no.~2, 2023.

\bibitem{he2024enhancing}
F.~He, F.~Li, and P.~Liang, ``Enhancing smart contract security: Leveraging pre-trained language models for advanced vulnerability detection,'' {\em IET Blockchain}, 2024.

\bibitem{merity2017regularizing}
S.~Merity, N.~S. Keskar, and R.~Socher, ``Regularizing and optimizing lstm language models,'' {\em arXiv preprint arXiv:1708.02182}, 2017.

\bibitem{ma2013multilevel}
Runing Ma, Xiuli Wang, and Jundi Ding, ``Multilevel core-sets based aggregation clustering algorithm,'' {\em Journal of Software}, vol.~24, no.~3, pp.~490--506, 2013.

\bibitem{guo2024smart}
J.~Guo, L.~Lu, and J.~Li, ``Smart contract vulnerability detection based on multi-scale encoders,'' {\em Electronics}, vol.~13, no.~3, p.~489, 2024.

\bibitem{qian2020towards}
P.~Qian, Z.~Liu, Q.~He, R.~Zimmermann, and X.~Wang, ``Towards automated reentrancy detection for smart contracts based on sequential models,'' {\em IEEE Access}, vol.~8, pp.~19685--19695, 2020.

\bibitem{sun2023assbert}
X.~Sun, L.~Tu, J.~Zhang, J.~Cai, B.~Li, and Y.~Wang, ``Assbert: Active and semi-supervised bert for smart contract vulnerability detection,'' {\em Journal of Information Security and Applications}, vol.~73, p.~103423, 2023.

\bibitem{gu2023trap}
T.~Gu, M.~Han, S.~He, and X.~Chen, ``Trap contract detection in blockchain with improved transformer,'' in {\em GLOBECOM 2023-2023 IEEE Global Communications Conference}, pp.~5141--5146, IEEE, 2023.

\bibitem{hu2022scsguard}
H.~Hu, Q.~Bai, and Y.~Xu, ``Scsguard: Deep scam detection for ethereum smart contracts,'' in {\em IEEE INFOCOM 2022-IEEE Conference on Computer Communications Workshops (INFOCOM WKSHPS)}, pp.~1--6, IEEE, 2022.

\bibitem{bariviera2017inefficiency}
A.~F. Bariviera, ``The inefficiency of bitcoin revisited: A dynamic approach,'' {\em Economics Letters}, vol.~161, pp.~1--4, 2017.

\bibitem{cong2019blockchain}
L.~W. Cong and Z.~He, ``Blockchain disruption and smart contracts,'' {\em The Review of Financial Studies}, vol.~32, no.~5, pp.~1754--1797, 2019.

\bibitem{armour2016principles}
J.~Armour, D.~Awrey, P.~L. Davies, L.~Enriques, J.~N. Gordon, C.~P. Mayer, and J.~Payne, {\em Principles of financial regulation}.
\newblock Oxford University Press, 2016.

\bibitem{catalini2020some}
C.~Catalini and J.~S. Gans, ``Some simple economics of the blockchain,'' {\em Communications of the ACM}, vol.~63, no.~7, pp.~80--90, 2020.

\bibitem{Pilipchenko2021USING}
A.~Pilipchenko, V.~Kuzminsky, and O.~Chumachenko, ``Using methods of technical analysis to forecast the cryptocurrency market,'' {\em "Scientific notes of the University"KROK"}, 2021.

\bibitem{pronchakov2019methods}
Y.~Pronchakov and O.~Bugaienko, ``Methods of forecasting the prices of cryptocurrency on the financial markets,'' {\em Technology transfer: innovative solutions in Social Sciences and Humanities}, pp.~13--16, 2019.

\bibitem{yang2019price}
L.~Yang, X.-Y. Liu, X.~Li, and Y.~Li, ``Price prediction of cryptocurrency: an empirical study,'' in {\em Smart Blockchain: Second International Conference, SmartBlock 2019, Birmingham, UK, October 11--13, 2019, Proceedings 2}, pp.~130--139, Springer, 2019.

\bibitem{catania2019forecasting}
L.~Catania, S.~Grassi, and F.~Ravazzolo, ``Forecasting cryptocurrencies under model and parameter instability,'' {\em International Journal of Forecasting}, vol.~35, no.~2, pp.~485--501, 2019.

\bibitem{lahmiri2019cryptocurrency}
S.~Lahmiri and S.~Bekiros, ``Cryptocurrency forecasting with deep learning chaotic neural networks,'' {\em Chaos, Solitons \& Fractals}, vol.~118, pp.~35--40, 2019.

\bibitem{mittal2018automated}
R.~Mittal, S.~Arora, and M.~Bhatia, ``Automated cryptocurrencies prices prediction using machine learning,'' {\em ICTACT Journal on Soft Computing}, vol.~8, no.~04, p.~4, 2018.

\bibitem{sun2020novel}
X.~Sun, M.~Liu, and Z.~Sima, ``A novel cryptocurrency price trend forecasting model based on lightgbm,'' {\em Finance Research Letters}, vol.~32, p.~101084, 2020.

\bibitem{jay2020stochastic}
P.~Jay, V.~Kalariya, P.~Parmar, S.~Tanwar, N.~Kumar, and M.~Alazab, ``Stochastic neural networks for cryptocurrency price prediction,'' {\em Ieee access}, vol.~8, pp.~82804--82818, 2020.

\bibitem{alamery2023cryptocurrency}
F.~M.~S. Alamery, ``Cryptocurrency analysis using machine learning and deep learning approaches,'' {\em Journal of Computer \& Electrical and Electronics Engineering Sciences}, vol.~1, no.~2, pp.~29--33, 2023.

\bibitem{abdul2023modelling}
N.~Abdul~Rashid and M.~T. Ismail, ``Modelling and forecasting the trend in cryptocurrency prices,'' {\em Journal of Information and Communication Technology}, vol.~22, no.~3, pp.~449--501, 2023.

\bibitem{patel2020deep}
M.~M. Patel, S.~Tanwar, R.~Gupta, and N.~Kumar, ``A deep learning-based cryptocurrency price prediction scheme for financial institutions,'' {\em Journal of information security and applications}, vol.~55, p.~102583, 2020.

\bibitem{gunarto2023predicting}
D.~M. Gunarto, S.~Sa'adah, and D.~Q. Utama, ``Predicting cryptocurrency price using rnn and lstm method,'' {\em Jurnal Sisfokom (Sistem Informasi dan Komputer)}, vol.~12, no.~1, pp.~1--8, 2023.

\bibitem{kim2022deep}
G.~Kim, D.-H. Shin, J.~G. Choi, and S.~Lim, ``A deep learning-based cryptocurrency price prediction model that uses on-chain data,'' {\em IEEE Access}, vol.~10, pp.~56232--56248, 2022.

\bibitem{kim2021cryptocurrency}
J.~Kim, S.~Kim, H.~Wimmer, and H.~Liu, ``A cryptocurrency prediction model using lstm and gru algorithms,'' in {\em 2021 IEEE/ACIS 6th International Conference on Big Data, Cloud Computing, and Data Science (BCD)}, pp.~37--44, IEEE, 2021.

\bibitem{liu2023lstm}
Y.~Liu, G.~Xiao, W.~Chen, and Z.~Zheng, ``A lstm and gru-based hybrid model in the cryptocurrency price prediction,'' in {\em International Conference on Blockchain and Trustworthy Systems}, pp.~32--43, Springer, 2023.

\bibitem{steinert2018predicting}
L.~Steinert and C.~Herff, ``Predicting altcoin returns using social media,'' {\em PloS one}, vol.~13, no.~12, p.~e0208119, 2018.

\bibitem{inamdar2019predicting}
A.~Inamdar, A.~Bhagtani, S.~Bhatt, and P.~M. Shetty, ``Predicting cryptocurrency value using sentiment analysis,'' in {\em 2019 International conference on intelligent computing and control systems (ICCS)}, pp.~932--934, IEEE, 2019.

\bibitem{wolk2020advanced}
K.~Wo{\l}k, ``Advanced social media sentiment analysis for short-term cryptocurrency price prediction,'' {\em Expert Systems}, vol.~37, no.~2, p.~e12493, 2020.

\bibitem{pathak2020cryptocurrency}
S.~Pathak and A.~Kakkar, ``Cryptocurrency price prediction based on historical data and social media sentiment analysis,'' {\em Innovations in Computer Science and Engineering: Proceedings of 7th ICICSE}, pp.~47--55, 2020.

\bibitem{prajapati2020predictive}
P.~Prajapati, ``Predictive analysis of bitcoin price considering social sentiments,'' {\em arXiv preprint arXiv:2001.10343}, 2020.

\bibitem{oikonomopoulos2022cryptocurrency}
S.~Oikonomopoulos, K.~Tzafilkou, D.~Karapiperis, and V.~Verykios, ``Cryptocurrency price prediction using social media sentiment analysis,'' in {\em 2022 13th International Conference on Information, Intelligence, Systems \& Applications (IISA)}, pp.~1--8, IEEE, 2022.

\bibitem{koltun2023pump}
V.~Koltun and I.~P. Yamshchikov, ``Pump it: Twitter sentiment analysis for cryptocurrency price prediction,'' {\em Risks}, vol.~11, no.~9, p.~159, 2023.

\bibitem{bhatt2023sentiment}
S.~Bhatt, M.~Ghazanfar, and M.~Amirhosseini, ``Sentiment-driven cryptocurrency price prediction: A machine learning approach utilizing historical data and social media sentiment analysis,'' {\em Machine Learning and Applications: An International Journal (MLAIJ)}, vol.~10, no.~2/3, pp.~1--15, 2023.

\bibitem{chalkiadakis2022chain}
I.~Chalkiadakis, A.~Zaremba, G.~W. Peters, and M.~J. Chantler, ``On-chain analytics for sentiment-driven statistical causality in cryptocurrencies,'' {\em Blockchain: Research and Applications}, vol.~3, no.~2, p.~100063, 2022.

\bibitem{derbentsev2020forecasting}
V.~Derbentsev, N.~Datsenko, V.~Babenko, O.~Pushko, and O.~Pursky, ``Forecasting cryptocurrency prices using ensembles-based machine learning approach,'' in {\em 2020 IEEE International Conference on Problems of Infocommunications. Science and Technology (PIC S\&T)}, pp.~707--712, IEEE, 2020.

\bibitem{du2022new}
X.~Du, Z.~Tang, J.~Wu, K.~Chen, and Y.~Cai, ``A new hybrid cryptocurrency returns forecasting method based on multiscale decomposition and an optimized extreme learning machine using the sparrow search algorithm,'' {\em IEEE Access}, vol.~10, pp.~60397--60411, 2022.

\bibitem{boukhers2022ensemble}
Z.~Boukhers, A.~Bouabdallah, M.~Lohr, and J.~J{\"u}rjens, ``Ensemble and multimodal approach for forecasting cryptocurrency price,'' {\em arXiv preprint arXiv:2202.08967}, 2022.

\bibitem{akila2023cryptocurrency}
V.~Akila, M.~Nitin, I.~Prasanth, S.~Reddy, and A.~Kumar, ``A cryptocurrency price prediction model using deep learning,'' in {\em E3S Web of Conferences}, vol.~391, p.~01112, EDP Sciences, 2023.

\bibitem{zhao2022attention}
H.~Zhao, M.~Crane, and M.~Bezbradica, ``Attention! transformer with sentiment on cryptocurrencies price prerediction,'' 2022.

\bibitem{penmetsa2023cryptocurrency}
S.~Penmetsa and M.~Vemula, ``Cryptocurrency price prediction with lstm and transformer models leveraging momentum and volatility technical indicators,'' in {\em 2023 IEEE 3rd International Conference on Data Science and Computer Application (ICDSCA)}, pp.~411--416, IEEE, 2023.

\bibitem{davoudi2023decentralized}
M.~Davoudi, M.~Ghavipour, M.~Sargolzaei-Javan, and S.~Dinparast, ``Decentralized storage cryptocurrencies: An innovative network-based model for identifying effective entities and forecasting future price trends,'' 2023.

\bibitem{khaniki2024enhancing}
M.~A.~L. Khaniki and M.~Manthouri, ``Enhancing price prediction in cryptocurrency using transformer neural network and technical indicators,'' {\em arXiv preprint arXiv:2403.03606}, 2024.

\bibitem{sridhar2021multi}
S.~Sridhar and S.~Sanagavarapu, ``Multi-head self-attention transformer for dogecoin price prediction,'' in {\em 2021 14th International Conference on Human System Interaction (HSI)}, pp.~1--6, IEEE, 2021.

\bibitem{murray2023forecasting}
K.~Murray, A.~Rossi, D.~Carraro, and A.~Visentin, ``On forecasting cryptocurrency prices: A comparison of machine learning, deep learning, and ensembles,'' {\em Forecasting}, vol.~5, no.~1, pp.~196--209, 2023.

\bibitem{singh2024transformer}
S.~Singh and M.~Bhat, ``Transformer-based approach for ethereum price prediction using crosscurrency correlation and sentiment analysis,'' {\em arXiv preprint arXiv:2401.08077}, 2024.

\bibitem{son2022using}
Y.~Son, S.~Vohra, R.~Vakkalagadda, M.~Zhu, A.~Hirde, S.~Kumar, and A.~Rajaram, ``Using transformers and deep learning with stance detection to forecast cryptocurrency price movement,'' in {\em 2022 13th International Conference on Information and Communication Technology Convergence (ICTC)}, pp.~1--6, IEEE, 2022.

\bibitem{bhargavan2016formal}
K.~Bhargavan, A.~Delignat-Lavaud, C.~Fournet, A.~Gollamudi, G.~Gonthier, N.~Kobeissi, N.~Kulatova, A.~Rastogi, T.~Sibut-Pinote, N.~Swamy, {\em et~al.}, ``Formal verification of smart contracts: Short paper,'' in {\em Proceedings of the 2016 ACM workshop on programming languages and analysis for security}, pp.~91--96, 2016.

\bibitem{liu2018mining}
K.~Liu, D.~Kim, T.~F. Bissyand{\'e}, S.~Yoo, and Y.~Le~Traon, ``Mining fix patterns for findbugs violations,'' {\em IEEE Transactions on Software Engineering}, vol.~47, no.~1, pp.~165--188, 2018.

\bibitem{delmolino2016step}
K.~Delmolino, M.~Arnett, A.~Kosba, A.~Miller, and E.~Shi, ``Step by step towards creating a safe smart contract: Lessons and insights from a cryptocurrency lab,'' in {\em Financial Cryptography and Data Security: FC 2016 International Workshops, BITCOIN, VOTING, and WAHC, Christ Church, Barbados, February 26, 2016, Revised Selected Papers 20}, pp.~79--94, Springer, 2016.

\bibitem{nikolic2018finding}
I.~Nikoli{\'c}, A.~Kolluri, I.~Sergey, P.~Saxena, and A.~Hobor, ``Finding the greedy, prodigal, and suicidal contracts at scale,'' in {\em Proceedings of the 34th annual computer security applications conference}, pp.~653--663, 2018.

\bibitem{grech2018madmax}
N.~Grech, M.~Kong, A.~Jurisevic, L.~Brent, B.~Scholz, and Y.~Smaragdakis, ``Madmax: Surviving out-of-gas conditions in ethereum smart contracts,'' {\em Proceedings of the ACM on Programming Languages}, vol.~2, no.~OOPSLA, pp.~1--27, 2018.

\bibitem{bartoletti2017empirical}
M.~Bartoletti and L.~Pompianu, ``An empirical analysis of smart contracts: platforms, applications, and design patterns,'' in {\em Financial Cryptography and Data Security: FC 2017 International Workshops, WAHC, BITCOIN, VOTING, WTSC, and TA, Sliema, Malta, April 7, 2017, Revised Selected Papers 21}, pp.~494--509, Springer, 2017.

\bibitem{yu2019design}
J.~Yu, Q.~Zhang, and L.~Wang, ``Design of optimal hybrid controller for multi-phase batch processes with interval time varying delay,'' {\em IEEE Access}, vol.~7, pp.~164029--164043, 2019.

\bibitem{mendoza2018mobile}
A.~Mendoza and G.~Gu, ``Mobile application web api reconnaissance: Web-to-mobile inconsistencies \& vulnerabilities,'' in {\em 2018 IEEE Symposium on Security and Privacy (SP)}, pp.~756--769, IEEE, 2018.

\bibitem{mou2016convolutional}
L.~Mou, G.~Li, L.~Zhang, T.~Wang, and Z.~Jin, ``Convolutional neural networks over tree structures for programming language processing,'' in {\em Proceedings of the AAAI conference on artificial intelligence}, vol.~30, 2016.

\bibitem{xu2020state}
J.~Xu, K.~Yin, and L.~Liu, ``State-continuity approximation of markov decision processes via finite element methods for autonomous system planning,'' {\em IEEE Robotics and Automation Letters}, vol.~5, no.~4, pp.~5589--5596, 2020.

\bibitem{hu2018summarizing}
X.~Hu, G.~Li, X.~Xia, D.~Lo, S.~Lu, and Z.~Jin, ``Summarizing source code with transferred api knowledge,'' 2018.

\bibitem{ahmad2020transformer}
W.~U. Ahmad, S.~Chakraborty, B.~Ray, and K.-W. Chang, ``A transformer-based approach for source code summarization,'' {\em arXiv preprint arXiv:2005.00653}, 2020.

\bibitem{wei2019code}
B.~Wei, G.~Li, X.~Xia, Z.~Fu, and Z.~Jin, ``Code generation as a dual task of code summarization,'' {\em Advances in neural information processing systems}, vol.~32, 2019.

\bibitem{iyer2016summarizing}
S.~Iyer, I.~Konstas, A.~Cheung, and L.~Zettlemoyer, ``Summarizing source code using a neural attention model,'' in {\em 54th Annual Meeting of the Association for Computational Linguistics 2016}, pp.~2073--2083, Association for Computational Linguistics, 2016.

\bibitem{eriguchi2016tree}
A.~Eriguchi, K.~Hashimoto, and Y.~Tsuruoka, ``Tree-to-sequence attentional neural machine translation,'' {\em arXiv preprint arXiv:1603.06075}, 2016.

\bibitem{wan2018improving}
Y.~Wan, Z.~Zhao, M.~Yang, G.~Xu, H.~Ying, J.~Wu, and P.~S. Yu, ``Improving automatic source code summarization via deep reinforcement learning,'' in {\em Proceedings of the 33rd ACM/IEEE international conference on automated software engineering}, pp.~397--407, 2018.

\bibitem{hu2018deep}
X.~Hu, G.~Li, X.~Xia, D.~Lo, and Z.~Jin, ``Deep code comment generation,'' in {\em Proceedings of the 26th conference on program comprehension}, pp.~200--210, 2018.

\bibitem{gong2022source}
Z.~Gong, C.~Gao, Y.~Wang, W.~Gu, Y.~Peng, and Z.~Xu, ``Source code summarization with structural relative position guided transformer,'' in {\em 2022 IEEE International Conference on Software Analysis, Evolution and Reengineering (SANER)}, pp.~13--24, IEEE, 2022.

\bibitem{yang2021multi}
Z.~Yang, J.~Keung, X.~Yu, X.~Gu, Z.~Wei, X.~Ma, and M.~Zhang, ``A multi-modal transformer-based code summarization approach for smart contracts,'' in {\em 2021 IEEE/ACM 29th International Conference on Program Comprehension (ICPC)}, pp.~1--12, IEEE, 2021.

\bibitem{gao2022m2ts}
Y.~Gao and C.~Lyu, ``M2ts: Multi-scale multi-modal approach based on transformer for source code summarization,'' in {\em Proceedings of the 30th IEEE/ACM International Conference on Program Comprehension}, pp.~24--35, 2022.

\bibitem{gao2021code}
S.~Gao, C.~Gao, Y.~He, J.~Zeng, L.~Y. Nie, and X.~Xia, ``Code structure guided transformer for source code summarization. corr abs/2104.09340 (2021),'' {\em arXiv preprint arXiv:2104.09340}, 2021.

\bibitem{shi2023coss}
C.~Shi, B.~Cai, Y.~Zhao, L.~Gao, K.~Sood, and Y.~Xiang, ``Coss: leveraging statement semantics for code summarization,'' {\em IEEE Transactions on Software Engineering}, 2023.

\bibitem{gao2023code}
S.~Gao, C.~Gao, Y.~He, J.~Zeng, L.~Nie, X.~Xia, and M.~Lyu, ``Code structure--guided transformer for source code summarization,'' {\em ACM Transactions on Software Engineering and Methodology}, vol.~32, no.~1, pp.~1--32, 2023.

\bibitem{hellendoorn2019global}
V.~J. Hellendoorn, C.~Sutton, R.~Singh, P.~Maniatis, and D.~Bieber, ``Global relational models of source code,'' in {\em International conference on learning representations}, 2019.

\bibitem{choi2021learning}
Y.~Choi, J.~Bak, C.~Na, and J.-H. Lee, ``Learning sequential and structural information for source code summarization,'' in {\em Findings of the Association for Computational Linguistics: ACL-IJCNLP 2021}, pp.~2842--2851, 2021.

\bibitem{hu2021automating}
X.~Hu, Z.~Gao, X.~Xia, D.~Lo, and X.~Yang, ``Automating user notice generation for smart contract functions,'' in {\em 2021 36th IEEE/ACM International Conference on Automated Software Engineering (ASE)}, pp.~5--17, IEEE, 2021.

\bibitem{leclair2019neural}
A.~LeClair, S.~Jiang, and C.~McMillan, ``A neural model for generating natural language summaries of program subroutines,'' in {\em 2019 IEEE/ACM 41st International Conference on Software Engineering (ICSE)}, pp.~795--806, IEEE, 2019.

\bibitem{wei2020retrieve}
B.~Wei, Y.~Li, G.~Li, X.~Xia, and Z.~Jin, ``Retrieve and refine: exemplar-based neural comment generation,'' in {\em Proceedings of the 35th IEEE/ACM International Conference on Automated Software Engineering}, pp.~349--360, 2020.

\bibitem{antonopoulos2023mastering}
A.~M. Antonopoulos and D.~A. Harding, {\em Mastering bitcoin}.
\newblock " O'Reilly Media, Inc.", 2023.

\bibitem{Regnath2018LeapChain:}
E.~Regnath and S.~Steinhorst, ``Leapchain: Efficient blockchain verification for embedded iot,'' {\em 2018 IEEE/ACM International Conference on Computer-Aided Design (ICCAD)}, pp.~1--8, 2018.

\bibitem{Nasrulin2018A}
B.~Nasrulin, M.~Muzammal, and Q.~Qu, ``A robust spatio-temporal verification protocol for blockchain,'' pp.~52--67, 2018.

\bibitem{Li2017Proof}
K.~Li, H.~Li, H.~Hou, K.~Li, and Y.~Chen, ``Proof of vote: A high-performance consensus protocol based on vote mechanism \& consortium blockchain,'' {\em 2017 IEEE 19th International Conference on High Performance Computing and Communications; IEEE 15th International Conference on Smart City; IEEE 3rd International Conference on Data Science and Systems (HPCC/SmartCity/DSS)}, pp.~466--473, 2017.

\bibitem{Li2021Blockchain-based}
H.~Li, T.~Wang, Z.~Qiao, B.~Yang, Y.~Gong, J.~Wang, and G.~Qiu, ``Blockchain-based searchable encryption with efficient result verification and fair payment,'' {\em J. Inf. Secur. Appl.}, vol.~58, p.~102791, 2021.

\bibitem{Jayabalan2021A}
J.~Jayabalan and J.~N, ``A study on distributed consensus protocols and algorithms: The backbone of blockchain networks,'' {\em 2021 International Conference on Computer Communication and Informatics (ICCCI)}, pp.~1--10, 2021.

\bibitem{Kim2020Analysis}
S.~Kim and S.~Ryu, ``Analysis of blockchain smart contracts: Techniques and insights,'' {\em 2020 IEEE Secure Development (SecDev)}, pp.~65--73, 2020.

\bibitem{Wang2022An}
Y.~Wang, X.~Chen, Y.~Huang, H.-N. Zhu, and J.~Bian, ``An empirical study on real bug fixes in smart contracts projects,'' {\em ArXiv}, vol.~abs/2210.11990, 2022.

\bibitem{Alikhani2021Regulating}
A.~Alikhani and H.-R. Hamidi, ``Regulating smart contracts: An efficient integration approach,'' {\em Intell. Decis. Technol.}, vol.~15, pp.~397--404, 2021.

\bibitem{Kushwaha2022Ethereum}
S.~S. Kushwaha, S.~Joshi, D.~Singh, M.~Kaur, and H.-N. Lee, ``Ethereum smart contract analysis tools: A systematic review,'' {\em IEEE Access}, vol.~PP, pp.~1--1, 2022.

\bibitem{Charlier2017Modeling}
J.~Charlier, R.~State, and J.~Hilger, ``Modeling smart contracts activities: A tensor based approach,'' {\em ArXiv}, vol.~abs/1905.09868, 2017.

\bibitem{Zhou2023Security}
X.~cong Zhou, Y.~Chen, H.~Guo, X.~Chen, and Y.~Huang, ``Security code recommendations for smart contract,'' {\em 2023 IEEE International Conference on Software Analysis, Evolution and Reengineering (SANER)}, pp.~190--200, 2023.

\bibitem{Prause2019Smart}
G.~Prause, ``Smart contracts for smart supply chains,'' {\em IFAC-PapersOnLine}, 2019.

\bibitem{Dolgui2020Blockchainoriented}
A.~Dolgui, D.~Ivanov, S.~Potryasaev, B.~Sokolov, M.~Ivanova, and F.~Werner, ``Blockchain-oriented dynamic modelling of smart contract design and execution in the supply chain,'' {\em International Journal of Production Research}, vol.~58, pp.~2184 -- 2199, 2020.

\bibitem{Trautmann2020Smart}
L.~Trautmann and R.~Lasch, ``Smart contracts in the context of procure-to-pay,'' {\em Smart and Sustainable Supply Chain and Logistics – Trends, Challenges, Methods and Best Practices}, 2020.

\bibitem{Aejas2021Effective}
B.~Aejas and A.~Bouras, ``Effective smart contracts for supply chain contracts,'' {\em Building Resilience at Universities: Role of Innovation and Entrepreneurship}, 2021.

\bibitem{Borselli2019Smart}
A.~Borselli, ``Smart contracts in insurance: A law and futurology perspective,'' pp.~101--125, 2019.

\bibitem{Dosovitskiy2020An}
A.~Dosovitskiy, L.~Beyer, A.~Kolesnikov, D.~Weissenborn, X.~Zhai, T.~Unterthiner, M.~Dehghani, M.~Minderer, G.~Heigold, S.~Gelly, J.~Uszkoreit, and N.~Houlsby, ``An image is worth 16x16 words: Transformers for image recognition at scale,'' {\em ArXiv}, vol.~abs/2010.11929, 2020.

\bibitem{Amin2021T2NER:}
S.~Amin and G.~Neumann, ``T2ner: Transformers based transfer learning framework for named entity recognition,'' pp.~212--220, 2021.

\bibitem{Gabriel2023Leveraging}
R.~A. Gabriel, B.~H. Park, S.~Mehdipour, D.~N. Bongbong, S.~Simpson, and R.~S. Waterman, ``Leveraging a natural language processing model (transformers) on electronic medical record notes to classify persistent opioid use after surgery,'' {\em Anesthesia \& Analgesia}, vol.~137, pp.~714 -- 716, 2023.

\bibitem{Liang2023Human}
Y.~Liang, K.~Feng, and Z.~Ren, ``Human activity recognition based on transformer via smart-phone sensors,'' {\em 2023 IEEE 3rd International Conference on Computer Communication and Artificial Intelligence (CCAI)}, pp.~267--271, 2023.

\bibitem{Yu2022Technology}
C.~Yu, W.~Yang, F.~Xie, and J.~He, ``Technology and security analysis of cryptocurrency based on blockchain,'' {\em Complex.}, vol.~2022, pp.~5835457:1--5835457:15, 2022.

\bibitem{Ghosh2020Security}
A.~Ghosh, S.~Gupta, A.~Dua, and N.~Kumar, ``Security of cryptocurrencies in blockchain technology: State-of-art, challenges and future prospects,'' {\em J. Netw. Comput. Appl.}, vol.~163, p.~102635, 2020.

\bibitem{Sanju2023Stock-Crypto-App}
T.~Sanju, H.~Liyanage, K.~Bandara, D.~Kandakkulama, D.~D. Silva, and J.~Perera, ``Stock-crypto-app – recommendation system for stock and cryptocurrency market using cutting edge machine learning technology,'' {\em International Research Journal of Innovations in Engineering and Technology}, 2023.

\bibitem{corbet2019cryptocurrencies}
S.~Corbet, B.~Lucey, A.~Urquhart, and L.~Yarovaya, ``Cryptocurrencies as a financial asset: A systematic analysis,'' {\em International Review of Financial Analysis}, vol.~62, pp.~182--199, 2019.

\bibitem{baur2018bitcoin}
D.~G. Baur, K.~Hong, and A.~D. Lee, ``Bitcoin: Medium of exchange or speculative assets?,'' {\em Journal of International Financial Markets, Institutions and Money}, vol.~54, pp.~177--189, 2018.

\bibitem{Khan2021Systematic}
D.~Khan, L.~T. Jung, and M.~Hashmani, ``Systematic literature review of challenges in blockchain scalability,'' {\em Applied Sciences}, 2021.

\bibitem{Rana2021Analysis}
N.~Rana, Y.~K. Dwivedi, and D.~L. Hughes, ``Analysis of challenges for blockchain adoption within the indian public sector: an interpretive structural modelling approach,'' {\em Inf. Technol. People}, vol.~35, pp.~548--576, 2021.

\bibitem{Kneissler2023Addressing}
A.~Kneissler and S.~Oelbracht, ``Addressing the practical challenges of implementing blockchain in engineering and manufacturing,'' {\em AHFE International}, 2023.

\bibitem{Golait2023Blockchain}
P.~Golait, D.~S. Tomar, R.~Pateriya, and Y.~K. Sharma, ``Blockchain security and challenges: A review,'' {\em 2023 IEEE 2nd International Conference on Industrial Electronics: Developments \& Applications (ICIDeA)}, pp.~140--145, 2023.

\bibitem{Bernabe2019Privacy-Preserving}
J.~B. Bernabe, J.~L. Cánovas, J.~L. Hernández-Ramos, R.~T. Moreno, and A.~Skarmeta, ``Privacy-preserving solutions for blockchain: Review and challenges,'' {\em IEEE Access}, vol.~7, pp.~164908--164940, 2019.

\bibitem{8963950}
S.~Wang, A.~Pathania, and T.~Mitra, ``Neural network inference on mobile socs,'' {\em IEEE Design \& Test}, vol.~37, no.~5, pp.~50--57, 2020.

\bibitem{Laroiya2020Applications}
C.~Laroiya, D.~Saxena, and C.~Komalavalli, ``Applications of blockchain technology,'' {\em Handbook of Research on Blockchain Technology}, 2020.

\bibitem{Aggarwal2019Blockchain}
S.~Aggarwal, R.~Chaudhary, G.~Aujla, N.~Kumar, K.-K.~R. Choo, and A.~Y. Zomaya, ``Blockchain for smart communities: Applications, challenges and opportunities,'' {\em J. Netw. Comput. Appl.}, vol.~144, pp.~13--48, 2019.

\bibitem{Bhutta2021A}
M.~N.~M. Bhutta, A.~Khwaja, A.~Nadeem, H.~F. Ahmad, M.~Khan, M.~Hanif, H.~Song, M.~A. Alshamari, and Y.~Cao, ``A survey on blockchain technology: Evolution, architecture and security,'' {\em IEEE Access}, vol.~9, pp.~61048--61073, 2021.

\bibitem{10.1093comjnlbxad060}
J.~Zhang, A.~Ye, J.~Chen, Y.~Zhang, and W.~Yang, ``{CSFL: Cooperative Security Aware Federated Learning Model Using The Blockchain},'' {\em The Computer Journal}, vol.~67, pp.~1298--1308, 06 2023.

\bibitem{electronics11162597}
R.~Chen, L.~Wang, C.~Peng, and R.~Zhu, ``An effective sharding consensus algorithm for blockchain systems,'' {\em Electronics}, vol.~11, no.~16, 2022.

\end{thebibliography}
\end{document}